\documentclass{article} 
\usepackage{iclr2027_conference,times}

\usepackage{amsmath,amsfonts,bm}

\def\eqref#1{equation~\ref{#1}}

\def\1{\bm{1}}

\DeclareMathAlphabet{\mathsfit}{\encodingdefault}{\sfdefault}{m}{sl}
\SetMathAlphabet{\mathsfit}{bold}{\encodingdefault}{\sfdefault}{bx}{n}

\usepackage{amsmath}
\usepackage{amssymb}
\usepackage{algorithm}
\usepackage{algpseudocode}
\usepackage{graphicx}
\usepackage{url}
\usepackage{titletoc}
\usepackage[export]{adjustbox}
\definecolor{citecolor}{HTML}{0071bc}
\usepackage[pagebackref=false,breaklinks=true,letterpaper=true,colorlinks,citecolor=citecolor,bookmarks=false]{hyperref}
\usepackage{adjustbox}
\usepackage{subcaption}
\usepackage{booktabs}
\usepackage{multirow}
\usepackage{array}
\usepackage{wrapfig}

\usepackage{xcolor}
\definecolor{bestblue}{HTML}{14479E}
\definecolor{secondorange}{HTML}{BF5B04}
\newcommand{\best}[1]{\textcolor{bestblue}{\textbf{#1}}}
\newcommand{\second}[1]{\textcolor{secondorange}{#1}}
\newcommand{\rankkey}{\best{best} and \second{second-best}}

\definecolor{rebuttalteal}{HTML}{008C95}

\newcommand{\lorekvE}{LORE-KV\,/E}
\newcommand{\lorekvD}{LORE-KV\,/D}

\algrenewcommand{\algorithmiccomment}[1]{\hfill\(\triangleright\)\,\footnotesize\textit{#1}}
\newcommand{\algstage}[2]{\Statex\vspace{1.5pt}\textbf{#1}\hfill\textit{(#2)}\vspace{1pt}}
\title{Monte Carlo Estimation for KV Cache Eviction}

\author{%
Ahsan Bilal\textsuperscript{1},\;
Muhammad Ahmed Mohsin\textsuperscript{2},\;
Muhammad Umer\textsuperscript{2},\;
Wajih Hassan Raza\textsuperscript{3}, \\
\textbf{Atta Ul Asad\textsuperscript{4},\;
Young D.~Kwon\textsuperscript{5},\;
Michal Valko\textsuperscript{6},\;
Dean F.~Hougen\textsuperscript{1}} \\[4pt]
\textsuperscript{1}University of Oklahoma \quad
\textsuperscript{2}Stanford University \quad
\textsuperscript{3}University of Houston \\
\textsuperscript{4}Lahore University of Management Sciences \quad
\textsuperscript{5}University of Cambridge \quad
\textsuperscript{6}Isara Labs
}

 \usepackage[textsize=tiny]{todonotes}

\iclrfinalcopy  

\begin{document}

\maketitle
\lhead{Preprint}  

\begin{abstract}
Most KV-cache eviction methods ask, in effect, \emph{which memory appeared important while reading the prompt?} We instead ask, \emph{which memory will matter while answering?} Since decoding queries are unavailable at eviction time, prior future-aware methods rely on pseudo-responses or synthetic future-query estimates. We cast fixed-budget future-aware eviction as distributional estimation over plausible model-conditional query trajectories and introduce \textbf{LORE-KV} (\emph{\textbf{\underline{L}}ookahead \textbf{\underline{O}}utput-perturbation with \textbf{\underline{R}}eliability-weighted \textbf{\underline{E}}nsembles for \textbf{\underline{K}}ey-\textbf{\underline{V}}alue caches}), a training-free method that samples short autoregressive continuations from the frozen target model and uses their response-side query states to estimate prompt-token utility. Tokens are scored by projected leave-one-out attention-output deletion cost and aggregated across sampled futures with optional trajectory weighting. The temporary continuations are discarded before final decoding, requiring no auxiliary model or training. Ablations isolate the mechanism: at $B{=}128$, a single response-side continuation recovers about $89\%$ of the gain over the prompt-window control, while additional futures provide smaller improvements. At $B{=}128$, LORE-KV raises the LongBench average on Qwen2.5-14B from $45.49$ to $48.24$ ($+2.75$) and the 16K RULER average on Mistral-7B from $45.20$ to $51.05$ ($+5.85$); on Llama-3.1-8B, the RULER margin over AnDPro widens to $+8.13$ and $+11.65$ points at 64K and 128K contexts. Gains diminish at larger cache budgets and coexist with task-level regressions. LORE-KV incurs $1.46$-$2.77\times$ AnDPro's per-sample wall-clock time as a one-time compression overhead across six dense and hybrid-attention backbones.
\textbf{Project webpage:} \url{https://ahsanbilal7.github.io/train_free_kv/}
\end{abstract}

\section{Introduction} \label{sec:introduction}
Autoregressive large language models (LLMs) rely on key-value (KV) caching to reuse previously computed hidden states during decoding. Although KV caching accelerates generation, its memory and bandwidth costs grow with sequence length and batch size, making long-context inference increasingly expensive \citep{behnam2025rocketkv,li2025kvtuner,lancucki2026inference}. KV-cache eviction reduces this cost by retaining only a subset of KV pairs, with existing methods estimating importance using recency or attention sinks \citep{xiao2023streamingllm}, accumulated or prompt-end attention \citep{zhang2023h2o,liu2023scissorhands,li2024snapkv}, or key-space geometry \citep{devoto2024simple,park2026keydiff}.

A key challenge is that KV eviction happens before future decoding queries are known, so prompt-based scores may keep tokens that matter during prefill but not during generation. Future-aware methods reduce this mismatch using pseudo or synthetic queries \citep{wang2025lookahead}, but the benefit and cost of using multiple plausible futures remain unclear. Learned predictors can estimate future utility, but require model-specific training \citep{ahn2026lookaheadkv,moschella2026learning,bui2026make,yang2026indexmem}. Moreover, attention alone is incomplete because deletion impact also depends on the value vectors \citep{guo2024criticalkv,geng2025accurate,mai2026reformulating}. These gaps motivate a training-free method that combines future-query estimation with output-aware token scoring.

To address these limitations, we propose \textbf{LORE-KV} (\emph{\textbf{\underline{L}}ookahead \textbf{\underline{O}}utput-perturbation with \textbf{\underline{R}}eliability-weighted \textbf{\underline{E}}nsembles for \textbf{\underline{K}}ey-\textbf{\underline{V}}alue caches}), a training-free method that estimates token importance using the queries the model is likely to encounter during generation. Prior future-aware methods estimate decoding demand from either a pseudo-response trajectory, as in LAQ \citep{wang2025lookahead}, or synthetic queries from a fitted hidden-state distribution. LORE-KV instead samples multiple model-conditional autoregressive trajectories from the frozen target model and estimates fixed-budget token utility via reliability-weighted expected projected deletion cost over their query states. Our focus is the query reference: what multiple sampled futures add beyond one, and the explicit trade-off of additional compression-time compute for retained-cache quality. LORE-KV is training-free to deploy but incurs lookahead overhead on every request.

\textbf{Contributions.}
Our contributions are threefold. \textbf{First}, we quantify and mechanistically characterize the \emph{query-reference mismatch} between prompt-side and future decoding queries, and show that the benefit of future-aware eviction increases with this mismatch. \textbf{Second}, we use multiple short future trajectories to estimate which cached tokens are most important for upcoming generation, and softly weight more reliable trajectories. \textbf{Third}, we validate this mechanism through query controls and matched lookahead-token budget controls, future-query oracle analyses, and evaluations across six dense and hybrid-attention backbones, cache budgets, context lengths up to 128K tokens, and benchmarks.

\section{Related Work}
\label{sec:related_work}
\subsection{Prompt-side KV eviction} \label{subsec:rw_eviction}
Most KV-cache eviction methods estimate token importance from information available during prefill. StreamingLLM preserves attention sinks and recent tokens \citep{xiao2023streamingllm}, H2O accumulates historical attention \citep{zhang2023h2o}, and SnapKV scores tokens using a prompt-end observation window \citep{li2024snapkv}. Other methods improve how a fixed budget is distributed across layers or heads, including PyramidKV and Ada-KV \citep{cai2024pyramidkv,feng2024adakv}, while DefensiveKV improves robustness to noisy query estimates \citep{feng2025taming}. These methods refine retention or budget allocation but remain prompt-side; LORE-KV instead estimates token utility from plausible future decoding queries.
\subsection{Future-query estimation and output-aware scoring} \label{subsec:rw_future}
A more recent line of work approximates the queries encountered during decoding. Expected Attention analytically estimates attention under a future-query distribution \citep{devoto2025expected}, while DapQ constructs position-aware pseudo-queries \citep{tian2026dapq}. LAQ obtains lookahead queries from a short pseudo-response, LAQ++ combines them with prompt-local queries \citep{wang2025lookahead}, and GVote samples synthetic queries from a fitted hidden-state distribution to determine an adaptive cache budget \citep{tang2025adaptive}. Learned approaches such as LookaheadKV and IndexMem predict future utility or learn compact memory mechanisms \citep{ahn2026lookaheadkv,yang2026indexmem}. Complementary methods improve the eviction objective: OBCache measures attention-output perturbation \citep{gu2025obcache}, AnDPro uses projected attention-output information \citep{geng2025accurate}, and LaProx incorporates projected values and inter-head interactions \citep{mai2026reformulating}. LORE-KV samples multiple model-conditional autoregressive trajectories and, under a fixed cache budget, estimates token utility via reliability-weighted expected projected deletion cost over their query states.
\subsection{Orthogonal KV-cache compression} \label{subsec:rw_orthogonal}
Other work reduces KV-cache cost along dimensions orthogonal to prompt-time token selection. Quantization methods such as KIVI and KVQuant reduce cache precision \citep{liu2024kivi,hooper2024kvquant}; representation-oriented methods exploit low-rank, cross-layer, or mergeable structure \citep{kang2024gear,liu2024minicache}; and serving or decode-time methods control cache growth during long generation \citep{liu2026kvserve,lancucki2026inference,su2026longflow}. These techniques are complementary to LORE-KV because they modify cache representation, precision, serving, or decode-time maintenance rather than the prompt-time criterion used to determine which KV entries should survive eviction. Aggressive eviction can additionally remove instructions or answer-bearing evidence \citep{yang2024notokenleft,chen2025pitfalls, garcia2026protection}. Additional related work is discussed in Appendix~\ref{app:extended_related_work}.

\section{Methodology} \label{sec:methodology}
LORE-KV consists of three stages: (i) Monte Carlo future-query estimation from $M$ short continuations (Section~\ref{subsec:lookahead}), (ii) reliability weighting using likelihood, entropy, or disagreement (Section~\ref{subsec:confidence}), and (iii) projected counterfactual eviction scoring (Sections~\ref{subsec:deletion}-\ref{subsec:core_score}). Algorithm~\ref{alg:lorekv} summarizes the method, Table~\ref{tab:hyperparams} lists the hyperparameters, and Appendices~\ref{app:method_details} and~\ref{sec:ablation_components} provide implementation details and ablations.

\subsection{Problem Setup} \label{subsec:problem_setup}
Let $f_\theta$ be a frozen pretrained transformer. Given a prompt $X=(x_1,\ldots,x_n)$, a prefill pass produces cached keys and values $\mathbf{K}^{l,h}$ and $\mathbf{V}^{l,h}$ for each layer $l$ and KV head $h$, along with prompt-window queries $\mathcal{Q}^{l,h}_{\mathcal{W}}$. We define the prompt-end window as $\mathcal{W}=\{n-w+1,\ldots,n\}$, where $w$ is the number of recent prompt tokens included in $\mathcal{W}$, set to $32$ by default. We optionally protect structural tokens $\mathcal{B}$. Together with the first-token attention sink \citep{xiao2023streamingllm}, the protected set is $\mathcal{P}=\{1\}\cup\mathcal{W}\cup\mathcal{B}$, leaving eviction candidates $\mathcal{C}=\{1,\ldots,n\}\setminus\mathcal{P}$. Further setup details are provided in Appendix~\ref{app:method_details}.

\subsection{Test-Time Lookahead Query Construction} \label{subsec:lookahead}
Because prompt-side queries can poorly represent those used during generation, LORE-KV uses a small test-time budget to sample $M$ short continuations from the frozen model, providing multiple plausible response-side query trajectories for estimating future token utility:
\begin{equation}
Y^{(m)}\sim f_\theta(\cdot\mid X),\qquad
Y^{(m)}=\{y^{(m)}_1,\ldots,y^{(m)}_T\},\qquad
m=1,\ldots,M,
\label{eq:lookahead_cont}
\end{equation}
where $T$ is small, and each continuation serves as a temporary probe of future decoding behavior. LORE-KV samples multiple autoregressive future trajectories, so the resulting query states reflect plausible response-conditioned futures before fixed-budget eviction. For trajectory $m$, we store the corresponding lookahead queries $\mathcal{Q}^{(m),l,h}_{\mathrm{look}}=\{\mathbf{q}^{(m),l,h}_{1},\ldots,\mathbf{q}^{(m),l,h}_{T}\}$ and form the scoring set $\mathcal{Q}^{l,h}_{\mathrm{score}}=\mathcal{Q}^{l,h}_{\mathcal{W}}\cup\bigcup_{m=1}^{M}\mathcal{Q}^{(m),l,h}_{\mathrm{look}}$. Each trajectory is evaluated under the same distribution from which it is sampled so that the reliability score remains correctly centered. Together, these trajectories provide a Monte Carlo approximation of future query states, improving coverage over prompt-only queries (Figure~\ref{fig:montecarlo_queries}; Appendix~\ref{app:query_coverage}). The single-surrogate case is detailed in Appendix~\ref{app:lookahead_details}, while Tables~\ref{tab:ablation_geometry} and~\ref{tab:compute_pareto} show gains with increasing lookahead compute that eventually saturate.
\begin{figure}
\centering
\includegraphics[width=\textwidth]{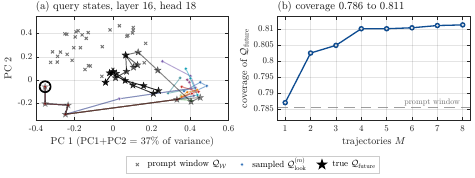}
\caption{Monte Carlo future-query estimation: lookahead queries better match future decoding queries, with coverage saturating near $M=4$.}
\label{fig:montecarlo_queries}
\end{figure}
\subsection{Particle Reliability Weighting: Entropy and Trajectory Disagreement} \label{subsec:confidence}
Because lookahead trajectories vary in reliability, LORE-KV assigns each a weight to reduce the influence of uncertain or atypical branches. Let $u\in\{1,\ldots,T\}$ index lookahead steps, $h_u^{(m)}=(X,y_{<u}^{(m)})$ denote the context of trajectory $m$ at step $u$, and $q(\cdot\mid h)$ the sampling distribution. Since raw log-probability conflates token atypicality with contextual uncertainty, we separate these effects using the centered score
\begin{equation}
  c^{(m)}_u
  =\log q\!\left(y^{(m)}_u\mid h^{(m)}_u\right)
  +H\!\left(q(\cdot\mid h^{(m)}_u)\right),
  \label{eq:centered_score}
\end{equation}
for which $\mathbb{E}[c^{(m)}_u\mid h^{(m)}_u]=0$. With $\bar c_m=\tfrac1T\sum_u c^{(m)}_u$ and $\bar H_m=\tfrac1T\sum_u H(q(\cdot\mid h^{(m)}_u))$, trajectory reliability is
\begin{equation}
  R_m=\beta\,\bar c_m-\gamma\,\bar H_m-\lambda_D\,\hat D_m,
  \label{eq:confidence}
\end{equation}
where $\bar c_m$ measures trajectory atypicality relative to local uncertainty, $\bar H_m$ captures uncertainty, and $\hat D_m$ measures disagreement with other trajectories. These signals are combined into a reliability score $R_m$, which is converted into soft trajectory weights,
\begin{equation}
\omega_m
= \frac{\exp(R_m)}
{\sum_{r=1}^{M}\exp(R_r)},
\label{eq:particle_weight}
\end{equation}
so that more reliable trajectories receive larger weights. We use soft weighting because reliability is only approximate. \emph{\lorekvE} uses entropy-based uncertainty, while \emph{\lorekvD} additionally uses query-space disagreement; both share the same trajectories and differ only in their reliability coefficients. Details and settings are provided in Appendix~\ref{app:reliability_details}, Table~\ref{tab:hyperparams}, and Table~\ref{tab:ablation_summary}.

\subsection{Closed-Form Counterfactual Deletion Effect} \label{subsec:deletion}
With the scoring queries in hand, we now specify \emph{what} is measured for each candidate token. For a scoring query $\mathbf{q}^{l,h}$ and candidate token $i\in\mathcal{C}$, the full-cache attention weight is
$a^{l,h}_{q,i}=\operatorname{Softmax}_i\!\left((\mathbf{q}^{l,h})^\top\mathbf{k}^{l,h}_{i}/\sqrt{d}\right)$,
where $d$ is the head dimension, and the corresponding full-cache attention output is
$\mathbf{o}^{l,h}_{q}=\sum_{j=1}^{n}a^{l,h}_{q,j}\mathbf{v}^{l,h}_{j}$.
We write
$(a^{l,h}_{q,i},\mathbf{o}^{l,h}_{q})
=\operatorname{Attn}(\mathbf{q}^{l,h},\mathbf{K}^{l,h},\mathbf{V}^{l,h})$
for this full-cache attention weight--output pair. If token $i$ is removed and the remaining attention weights are renormalized, the post-deletion output becomes
$\mathbf{o}^{l,h}_{q,-i}
=\sum_{j\neq i}\frac{a^{l,h}_{q,j}}{1-a^{l,h}_{q,i}+\epsilon}\mathbf{v}^{l,h}_{j}
=\frac{\mathbf{o}^{l,h}_{q}-a^{l,h}_{q,i}\mathbf{v}^{l,h}_{i}}
{1-a^{l,h}_{q,i}+\epsilon}$,
where $\epsilon$ prevents numerical instability. The corresponding leave-one-out deletion effect is therefore
\begin{equation}
\Delta\mathbf{o}^{l,h}_{q,i}
=\mathbf{o}^{l,h}_{q}-\mathbf{o}^{l,h}_{q,-i}
\approx
\frac{a^{l,h}_{q,i}}{1-a^{l,h}_{q,i}+\epsilon}
\left(\mathbf{v}^{l,h}_{i}-\mathbf{o}^{l,h}_{q}\right).
\label{eq:deletion_effect}
\end{equation}
Eq.~\ref{eq:deletion_effect} shows why attention alone is insufficient: a token is important only when it receives non-negligible attention and its value meaningfully differs from the current attention output. Let $\mathbf{W}^{O}_{l,h}$ denote the output-projection block for head $h$ in layer $l$. LORE-KV scores the residual-stream impact of deleting token $i$ as
\begin{equation}
e^{l,h}_{q,i}
=
\left\|
\mathbf{W}^{O}_{l,h}\Delta\mathbf{o}^{l,h}_{q,i}
\right\|_2^2 .
\label{eq:projected_score}
\end{equation}
Unlike raw attention, this score measures the deletion effect after projection into the model stream. Further details are given in Appendix~\ref{app:lookahead_details}. Figure~\ref{fig:cache_distribution} evaluates the resulting cache fidelity, while Figure~\ref{fig:layer_head_analysis} shows how the perturbation is distributed across layers and heads.

\subsection{Core Score, Selection, and Compressed Cache} \label{subsec:core_score}
For each token $i$, we group the per-query perturbation scores into $\mathcal{E}^{l,h}_{i,\mathcal{W}}=\{e^{l,h}_{q,i}:q\in\mathcal{Q}^{l,h}_{\mathcal{W}}\}$ for the prompt window and $\mathcal{E}^{l,h}_{i,m}=\{e^{l,h}_{q,i}:q\in\mathcal{Q}^{(m),l,h}_{\mathrm{look}}\}$ for trajectory $m$, then aggregate them using $\psi(\cdot)$, which reduces to the mean in the core configuration, to obtain the final token score,
\begin{equation}
S^{l,h}_{i}
=\psi(\mathcal{E}^{l,h}_{i,\mathcal{W}})
+\lambda_L\sum_{m=1}^{M}\omega_m\,\psi(\mathcal{E}^{l,h}_{i,m}).
\label{eq:final_score}
\end{equation}
The lookahead term changes which tokens are retained: at $B=128$, about 20\% differ from the prompt-only rule (Figure~\ref{fig:score_comparison}). With uniform weights, Eq.~\ref{eq:final_score} is a Monte Carlo estimate of expected projected deletion cost, while entropy/disagreement weighting provides a soft finite-sample robustness correction. Table~\ref{tab:ablation_summary} isolates this future-aware contribution. The resulting scores are then converted into the compressed cache through two core operators:
\begin{equation}
\{B_{l,h}\}_{h}=\operatorname{Alloc}\!\left(B_l,\{E_{l,h}\}_{h};\tau\right),
\qquad
\mathcal{R}^{l,h}=\operatorname{Select}\!\left(\mathbf{S}^{l,h},B_{l,h}\right).
\label{eq:alloc_select}
\end{equation}
For the core method, $\operatorname{Alloc}$ uniformly distributes each layer's budget across heads, and $\operatorname{Select}$ keeps the top-scoring tokens per head. The final retained set is $\mathcal{R}^{l,h}_{\mathrm{final}}=\mathcal{P}\cup\mathcal{R}^{l,h}$, with $(\mathbf{K}^{l,h}_{\star},\mathbf{V}^{l,h}_{\star})=\operatorname{Gather}(\mathbf{K}^{l,h},\mathbf{V}^{l,h},\mathcal{R}^{l,h}_{\mathrm{final}})$. The lookahead continuations are then discarded and decoding uses only the compressed cache. Adaptive allocation and diversity-aware selection are ablated separately in Appendix~\ref{sec:ablation_components}.

\section{Experiments} \label{sec:setup}
We evaluate LongBench \citep{bai2024longbench}, RULER \citep{hsieh2024ruler}, and Needle-in-a-Haystack (NIAH) for long-context understanding, controlled context-length stress testing from 4K to 128K tokens, and long-range retrieval, respectively. Mistral-7B-Instruct-v0.2 \citep{jiang2023mistral} and Llama-3.1-8B-Instruct \citep{dubey2024llama} support comparison with prior KV-cache methods, while Qwen2.5-14B-Instruct \citep{qwen2025qwen25technicalreport} and Ministral-3-8B-Instruct-2512 \citep{liu2026ministral3} test generalization across newer model families and scales. Qwen3.5-9B and Gemma-4-12B-it further test \emph{hybrid-attention} backbones, in which only a minority of layers keep a growing KV cache as discussed in Appendix~\ref{app:hybrid}. Shared settings, hyperparameters, and dataset details are provided in Tables~\ref{tab:hyperparams} and~\ref{tab:longbench_datasets}; main results appear in Sections~\ref{subsec:longbench}, \ref{subsec:niah}, and~\ref{subsec:ruler}, efficiency and mechanism analyses in Sections~\ref{subsec:efficiency} and~\ref{subsec:mechanism}, ablations in Table~\ref{tab:ablation_summary} and Appendix~\ref{sec:ablation_components}, and implementation details in Appendix~\ref{app:experimental_details}. We report \lorekvE{} and \lorekvD{} under the same pipeline, with further details in Appendices~\ref{app:reparam_grid} and~\ref{app:divergence}, and evaluate all methods following the protocol of \citet{geng2025accurate} with matched cache budgets, backbones, and observation windows. We highlight the \rankkey{} results at each cache budget. 

\begin{table}[!t]
\caption{LongBench results on Mistral-7B-Instruct-v0.2. Dataset abbreviations follow Table~\ref{tab:longbench_datasets}.}
\label{tab:longbench_mistral}
\centering
\setlength{\tabcolsep}{2pt}
\begin{adjustbox}{max width=\textwidth}
\begin{tabular}{@{}l*{17}{c}@{}}
\toprule
& \multicolumn{3}{c}{Single-Doc QA} & \multicolumn{3}{c}{Multi-Doc QA}
& \multicolumn{3}{c}{Summ.} & \multicolumn{3}{c}{Few-shot}
& \multicolumn{2}{c}{Synth.} & \multicolumn{2}{c}{Code} & \\
\cmidrule(lr){2-4}\cmidrule(lr){5-7}\cmidrule(lr){8-10}
\cmidrule(lr){11-13}\cmidrule(lr){14-15}\cmidrule(lr){16-17}
Method & NQA & Qasp & MF & HQA & 2WQA & Musq
& GovR & QMS & MN & TREC & TQA & SAMS
& PC & PR & LCC & RB-P & Avg. \\
\midrule

Full Cache
& 26.63 & 32.99 & 49.34
& 42.77 & 27.35 & 18.77
& 32.87 & 24.24 & 27.10
& 71.00 & 86.23 & 42.96
& 2.75 & 86.98
& 55.33 & 52.87
& 42.51 \\
\midrule
\multicolumn{18}{c}{\textit{Budget} $B=128$} \\
\midrule
H2O
& 21.19 & 21.66 & 38.60
& 30.63 & 20.65 & 12.19
& 20.65 & 22.42 & 21.81
& 39.00 & 82.52 & 40.68
& 2.98 & 79.56
& 49.13 & 46.76
& 34.40 \\

StreamingLLM
& 16.61 & 14.74 & 31.40
& 28.05 & 21.36 & 12.08
& 18.44 & 18.91 & 19.26
& 43.50 & 74.22 & 29.00
& 2.75 & 31.65
& 41.27 & 38.84
& 27.63 \\

SnapKV
& 19.17 & 21.40 & 42.93
& 36.76 & 22.44 & 15.86
& 19.16 & 21.84 & 21.55
& 47.50 & 84.15 & 40.24
& 2.30 & 68.26
& 50.69 & 47.13
& 35.09 \\

PyramidKV
& 20.16 & 21.77 & 43.55
& 36.78 & 23.12 & 14.39
& 19.53 & 22.03 & 21.47
& 51.00 & 84.62 & 40.24
& 2.79 & 70.77
& 50.57 & 46.53
& 35.58 \\

Ada-KV
& 21.79 & 23.03 & 47.07
& 38.70 & 22.85 & 15.92
& 19.94 & 23.05 & 21.84
& 63.00 & 85.36 & 40.01
& 3.20 & 72.13
& 51.36 & 49.29
& 37.41 \\

CriticalKV
& 21.55 & 23.27 & 46.67
& 38.95 & 24.95 & 16.85
& 20.83 & \second{23.14} & 22.16
& 65.00 & 85.53 & 40.02
& 3.48 & 77.81
& 53.58 & 50.98
& 38.42 \\

AnDPro
& \second{24.84} & 25.14 & \second{47.33}
& 39.79 & 23.99 & 17.34
& 20.40 & \best{23.45} & 21.94
& 67.50 & 85.59 & 40.62
& 3.19 & 76.94
& 53.77 & \second{51.55}
& 38.96 \\

LAQ++(8)
& \best{25.62} & \second{27.21} & 46.16
& 40.60 & 25.93 & 18.44
& \best{21.60} & 23.07 & \second{22.42}
& \best{70.00} & 86.18 & \best{42.03}
& \second{3.51} & 74.81
& \second{54.68} & 50.92
& 39.57 \\

\textbf{\lorekvE}
& 24.05 & 26.85 & \best{47.59}
& \best{42.19} & \second{26.39} & \second{19.20}
& \second{21.48} & 22.46 & \best{22.78}
& \second{69.50} & \best{86.64} & \second{41.08}
& 3.44 & \second{84.37}
& \best{54.90} & \best{51.56}
& \best{40.28} \\

\textbf{\lorekvD}
& 24.45 & \best{27.82} & 47.04
& \second{42.09} & \best{26.74} & \best{19.34}
& 21.44 & 22.39 & 21.58
& 69.00 & \second{86.51} & 41.00
& \best{3.83} & \best{84.62}
& 54.12 & 51.45
& \second{40.21} \\

\midrule
\multicolumn{18}{c}{\textit{Budget} $B=256$} \\
\midrule

H2O
& 21.54 & 22.92 & 42.56
& 31.07 & 22.53 & 13.76
& 22.52 & 22.40 & 23.09
& 40.50 & 84.20 & 40.77
& \best{3.41} & 86.10
& 50.98 & 48.17
& 36.03 \\

StreamingLLM
& 17.93 & 16.01 & 33.36
& 30.71 & 21.30 & 10.08
& 20.66 & 19.47 & 22.89
& 53.50 & 73.59 & 29.22
& 3.00 & 27.77
& 42.30 & 39.87
& 28.85 \\

SnapKV
& 22.37 & 23.74 & 48.13
& 38.56 & 22.43 & 15.66
& 21.91 & 23.13 & 23.15
& 61.50 & 85.45 & 41.42
& 3.09 & 84.54
& 53.22 & 50.24
& 38.66 \\

PyramidKV
& 20.09 & 24.00 & 47.33
& 38.24 & 22.48 & 16.02
& 21.40 & 22.45 & 22.63
& 63.00 & 84.93 & 40.98
& \second{3.40} & 82.48
& 52.78 & 49.36
& 38.22 \\

Ada-KV
& 23.02 & 25.70 & 49.23
& 40.18 & 24.76 & 17.43
& 22.17 & 23.21 & 23.48
& 67.50 & 85.58 & 41.67
& 2.99 & 84.65
& 55.07 & 52.17
& 39.93 \\

CriticalKV
& 23.70 & 26.66 & 49.00
& 40.05 & 25.12 & 17.86
& 21.99 & 23.35 & 23.45
& 68.00 & 86.19 & 42.07
& 2.92 & 85.81
& 53.81 & 50.42
& 40.03 \\

AnDPro
& 25.71 & 28.17 & \best{49.31}
& 40.04 & 25.81 & 18.68
& 21.66 & 23.64 & 23.63
& \best{70.50} & \best{86.59} & \best{42.60}
& 2.86 & 85.62
& 55.74 & 52.87
& 40.84 \\

LAQ++(8)
& 25.23 & \second{29.16} & \second{49.24}
& 41.70 & \second{26.85} & 18.62
& \best{23.73} & \second{23.69} & 23.38
& \best{70.50} & 86.24 & \second{42.54}
& 3.36 & 86.11
& 55.59 & 52.49
& 41.15 \\

\textbf{\lorekvE}
& \second{25.88} & \best{29.58} & 48.10
& \best{42.13} & 26.83 & \best{18.90}
& 23.23 & \best{23.79} & \best{24.07}
& \best{70.50} & \second{86.31} & 41.09
& 3.27 & \second{89.09}
& \second{56.15} & \second{53.12}
& \best{41.38} \\

\textbf{\lorekvD}
& \best{26.22} & 28.89 & 48.21
& \second{42.02} & \best{27.42} & \second{18.81}
& \second{23.72} & 23.04 & \second{24.00}
& \second{70.00} & 85.47 & 39.84
& 2.80 & \best{89.20}
& \best{56.58} & \best{53.44}
& \second{41.23} \\

\midrule
\multicolumn{18}{c}{\textit{Budget} $B=512$} \\
\midrule

H2O
& 21.72 & 26.03 & 44.81
& 32.33 & 23.16 & 14.86
& 23.65 & 22.84 & 24.70
& 42.00 & 85.22 & 41.57
& \best{3.40} & 86.45
& 53.04 & 49.68
& 37.22 \\

StreamingLLM
& 18.76 & 17.17 & 37.09
& 30.21 & 21.64 & 9.93
& 24.44 & 20.00 & \best{25.57}
& 62.00 & 72.36 & 29.95
& 2.48 & 18.17
& 43.70 & 40.13
& 29.60 \\

SnapKV
& 24.60 & 27.81 & 48.98
& 39.46 & 25.25 & 16.98
& 23.70 & 22.96 & 24.37
& 67.00 & 85.88 & 41.26
& 2.78 & 86.56
& 54.81 & 51.71
& 40.26 \\

PyramidKV
& 23.23 & 27.94 & 48.87
& 40.50 & 24.36 & 16.74
& 23.22 & 23.16 & 24.37
& 67.00 & 85.73 & 41.74
& \second{3.16} & 85.67
& 54.16 & 50.34
& 40.01 \\

Ada-KV
& 23.88 & 29.05 & 48.79
& 40.44 & 25.30 & 17.66
& 23.47 & 23.47 & 24.48
& 69.50 & 86.39 & 41.38
& 2.94 & 87.56
& 56.58 & 52.88
& 40.86 \\

CriticalKV
& 23.83 & 28.67 & 49.13
& 40.18 & 26.18 & 18.19
& 23.52 & 23.57 & 24.28
& \second{70.00} & 86.39 & 42.16
& 2.82 & 87.55
& 56.51 & 52.64
& 40.98 \\

AnDPro
& 24.87 & 30.99 & \second{49.58}
& 41.34 & \second{27.01} & 18.44
& 23.40 & 23.76 & 25.01
& \best{71.00} & \best{86.74} & \best{43.65}
& 2.83 & 86.23
& \best{57.46} & \second{53.61}
& 41.62 \\

LAQ++(8)
& 25.49 & 30.92 & \best{49.72}
& \second{41.50} & 26.84 & \second{19.20}
& \best{25.67} & \second{24.04} & \second{25.31}
& \best{71.00} & 86.43 & \second{43.14}
& 2.90 & 85.27
& 56.80 & 53.54
& \second{41.74} \\

\textbf{\lorekvE}
& \second{25.58} & \best{31.14} & 48.98
& \best{41.82} & \best{27.41} & 19.12
& 25.46 & 23.97 & 25.05
& \best{71.00} & \second{86.49} & 41.92
& 2.95 & \best{88.20}
& \second{56.85} & \best{53.82}
& \best{41.86} \\

\textbf{\lorekvD}
& \best{25.68} & \second{31.01} & 48.65
& 41.44 & 26.89 & \best{19.25}
& \second{25.48} & \best{24.67} & 24.82
& \best{71.00} & 86.12 & 42.16
& 2.98 & \second{88.00}
& 55.62 & 53.17
& 41.68 \\

\midrule
\multicolumn{18}{c}{\textit{Budget} $B=1024$} \\
\midrule

H2O
& 23.90 & 28.62 & 46.46
& 37.03 & 24.74 & 15.04
& 25.30 & 23.11 & 25.92
& 46.00 & 85.93 & 41.80
& \best{3.24} & 86.57
& 54.46 & 51.01
& 38.70 \\

StreamingLLM
& 19.42 & 21.69 & 41.75
& 32.40 & 22.18 & 11.18
& \best{27.13} & 21.09 & \best{26.59}
& 67.00 & 71.79 & 30.11
& 2.88 & 16.57
& 44.82 & 39.76
& 31.02 \\

SnapKV
& 25.47 & 29.57 & 49.33
& 40.90 & 25.53 & \best{19.01}
& 25.94 & 23.89 & 26.21
& 69.50 & 86.48 & 42.10
& 2.98 & \best{88.56}
& 55.57 & 51.92
& 41.44 \\

PyramidKV
& 24.21 & 29.86 & 48.93
& 40.75 & 25.05 & 18.77
& 25.73 & 24.06 & 25.65
& 68.50 & 86.31 & 42.25
& 2.97 & 87.17
& 54.75 & 52.10
& 41.07 \\

Ada-KV
& 25.60 & 31.15 & 49.02
& 41.47 & 27.07 & 18.77
& 25.40 & 23.84 & 25.92
& \second{70.50} & 86.55 & 43.07
& 2.54 & 87.14
& \second{57.27} & \best{53.85}
& 41.82 \\

CriticalKV
& \second{26.03} & 31.50 & 48.39
& 41.73 & 27.10 & 18.84
& 25.30 & \second{24.10} & 25.99
& \best{71.00} & 86.30 & 43.16
& 2.66 & \second{87.23}
& 56.69 & 53.20
& 41.83 \\

AnDPro
& 25.78 & \best{32.54} & \second{49.35}
& 42.20 & 26.94 & 18.63
& 25.33 & 24.01 & 26.29
& \best{71.00} & 86.30 & \second{43.46}
& 2.68 & 86.98
& 57.18 & \second{53.77}
& 42.03 \\

\textbf{\lorekvE}
& 25.98 & 31.81 & 49.26
& \second{43.96} & \second{27.17} & \second{18.91}
& 26.98 & \best{24.12} & \second{26.41}
& \best{71.00} & \second{86.60} & 42.30
& 2.68 & 86.49
& \best{57.83} & \second{53.77}
& \second{42.20} \\

\textbf{\lorekvD}
& \best{26.58} & \second{31.98} & \best{49.61}
& \best{44.57} & \best{28.19} & 18.50
& \second{27.04} & 23.57 & 25.71
& \second{70.50} & \best{87.00} & \best{44.07}
& \second{3.22} & 87.17
& 56.87 & 53.60
& \best{42.39} \\
\bottomrule
\end{tabular}
\end{adjustbox}
\end{table}

\subsection{Evaluation on LongBench} \label{subsec:longbench}
\begin{wrapfigure}{r}{0.49\textwidth}
    \centering
    \includegraphics[width=\linewidth]{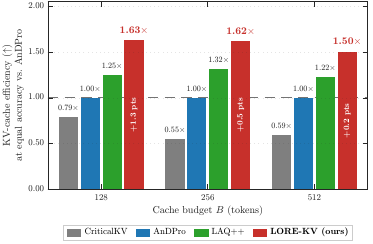}
    \caption{Equal-accuracy KV-cache efficiency on LongBench (Mistral-7B).}
    \label{fig:budget_efficiency}
\end{wrapfigure}

\textbf{Overall performance.} 
Tables~\ref{tab:longbench_mistral} and~\ref{tab:longbench_llama} show that LORE-KV consistently outperforms the matched baselines across both backbones and cache budgets, with the largest gains under aggressive compression and diminishing margins as the cache budget increases. At $B=128$ on Mistral, \lorekvE\ achieves $40.28$ versus $38.96$ for AnDPro. 

Across three independent seeds, \lorekvE\ obtains a mean LongBench score of $40.28$ (95\% CI $[39.97,40.59]$). The lower bound remains above AnDPro ($38.96$), indicating that the improvement is robust to lookahead sampling variation\footnote{LAQ++(8) is omitted at $B=1024$ because the original paper does not report results for this setting.}. We report published LAQ++(8) results only as a contextual comparison. For broader context, Figure~\ref{fig:external_comparison} shows a consistently favorable accuracy-memory trade-off against OBCache \citep{gu2025obcache}, CAOTE \citep{goel2025caote}, and Expected Attention \citep{devoto2025expected}: LORE-KV preserves near-full-cache quality at substantially lower cache retention. These comparisons are contextual rather than fully controlled because the evaluation protocols differ. Importantly, the improvement is not limited to pointwise score differences: Figure~\ref{fig:budget_efficiency} shows that LORE-KV reaches the same LongBench accuracy with $1.50$--$1.63\times$ greater cache efficiency than AnDPro across the aggressive-compression operating points.

\textbf{Where the gain comes from.} Table~\ref{tab:ablation_summary} shows that LORE-KV's main gain comes from correcting the query-reference mismatch. At $B=128$, replacing prompt-side queries with a single response-side rollout yields the largest improvement ($35.87\!\rightarrow\!39.81$), while multiple trajectories provide a further gain to $40.28$/$40.21$. Response-side queries drive the primary gain; multiple futures add further benefit, while deletion scoring and reliability weighting provide only minor refinements as discussed in Appendix~\ref{app:divergence}.

\textbf{Lookahead geometry.} Table~\ref{tab:ablation_geometry} shows that extending the lookahead horizon matters more than simply adding trajectories: for example, at $M=4$, increasing $T$ from $2$ to $16$ improves performance from $38.96$ to $40.34$. For a compute-controlled comparison, we combine LAQ-style response-side lookahead with AnDPro scoring, so that AnDPro scores the identical $M\times T$ lookahead queries under the same cache budgets and selector. LORE-KV remains above this hybrid baseline across the matched quality-compute frontier (Table~\ref{tab:compute_pareto}), showing that its gains are not explained by additional lookahead compute alone.

\textbf{Generalization across backbones.} The same operating-point trend holds on Qwen2.5-14B-Instruct and Ministral-3-8B-Instruct-2512 (Tables~\ref{tab:longbench_qwen2} and~\ref{tab:longbench_ministral3}). At $B=128$, LORE-KV improves over the strongest baseline by $2.75$ points on Qwen2.5 and $0.56$ points on Ministral-3, while the margins become very small by $B=1024$. The smaller gain on Ministral-3 likely reflects its long-context training and position-aware query scaling, which reduce the prompt-to-future query mismatch that LORE-KV is designed to correct (Appendix~\ref{app:new_backbones}).

\textbf{Hybrid-attention backbones.} LORE-KV also generalizes to Qwen3.5-9B and Gemma-4-12B-it, consistently outperforming the strongest baseline across cache budgets (Table~\ref{tab:longbench_hybrid_avg}), with the largest gains under aggressive compression. Detailed results are provided in Appendix~\ref{app:hybrid} and Tables~\ref{tab:longbench_qwen35}-\ref{tab:longbench_gemma4}.

\begin{wraptable}{r}{0.62\textwidth}
\vspace{-10pt}
\centering
\caption{\textbf{LongBench on hybrid-attention backbones.} Eviction is applied only to the KV-growing layers.}
\label{tab:longbench_hybrid_avg}

\vspace{-4pt}
\resizebox{\linewidth}{!}{%
\begin{tabular}{@{}lcccccccc@{}}
\toprule
& \multicolumn{4}{c}{Qwen3.5-9B}
& \multicolumn{4}{c}{Gemma-4-12B-it} \\
\cmidrule(lr){2-5}
\cmidrule(lr){6-9}

Method
& $B{=}128$ & $256$ & $512$ & $1024$
& $B{=}128$ & $256$ & $512$ & $1024$ \\
\midrule

Full Cache
& \multicolumn{4}{c}{52.97}
& \multicolumn{4}{c}{27.27} \\

\midrule

CriticalKV
& 46.30 & 48.30 & 49.58 & 50.31
& 19.74 & 21.53 & 22.45 & 23.22 \\

AnDPro
& 46.89 & 48.72 & 50.08 & 50.42
& 20.21 & 21.92 & 22.73 & 23.47 \\

\textbf{\lorekvE}
& \second{48.49}
& \second{50.05}
& \second{51.12}
& \second{51.79}
& \second{20.85}
& \second{22.63}
& \second{23.69}
& \second{24.44} \\

\textbf{\lorekvD}
& \best{48.57}
& \best{50.22}
& \best{51.16}
& \best{51.92}
& \best{20.93}
& \best{22.76}
& \best{23.81}
& \best{24.59} \\

\bottomrule
\end{tabular}%
}
\vspace{-8pt}
\end{wraptable}

\textbf{Cache efficiency.}
The accuracy gains also translate into lower cache requirements. Table~\ref{tab:budget_equivalence} shows that matching \lorekvE\ requires the strongest baseline to retain about $208$ tokens, corresponding to a $38.5\%$ cache saving, with a $31.1\%$ saving still observed at $B=512$. Figure~\ref{fig:budget_equivalence} confirms the same trend across the full budget range, showing that LORE-KV improves the overall quality-memory trade-off rather than only isolated operating points.
\subsection{Evaluation on Needle-in-a-Haystack (NIAH)}\label{subsec:niah}
LORE-KV also improves retrieval under aggressive compression. On Llama-3.1-8B-Instruct at $B=1024$, \lorekvE\ and \lorekvD\ tie for the best exact-match score ($0.977$), outperforming CriticalKV ($0.955$) and AnDPro ($0.932$), with the same trend under word recall.

As Figure~\ref{fig:niah-heatmap} shows (setup in Appendix~\ref{app:niah_extra}), the gains come from boundary-depth needles ($0\%$ and $100\%$) at 16K, which the baselines miss and LORE-KV retrieves. Figure~\ref{fig:retained_token_composition} further shows that LORE-KV retains answer-bearing tokens more often, linking the retrieval gains to its future-query-aware token selection.

\begin{figure}
\centering
\includegraphics[width=\textwidth]{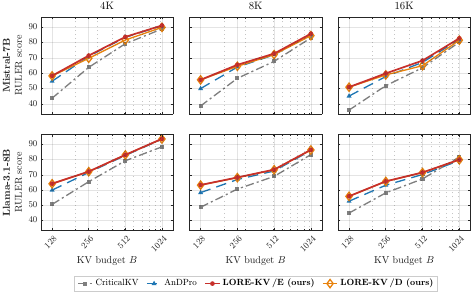}
\caption{RULER accuracy averaged over 13 tasks at 4K, 8K, and 16K context
lengths. Per-task results are given in Tables~\ref{tab:ruler_mistral_full}
and~\ref{tab:ruler_llama_full}. Results at 64K and 128K are given in Appendix~\ref{app:ruler_long}.}
\label{fig:ruler_main}
\end{figure}
\subsection{Evaluation on RULER: Sweeping Context Length Across Fixed Tasks} \label{subsec:ruler}
\begin{figure}
  \centering
  \includegraphics[width=\textwidth]{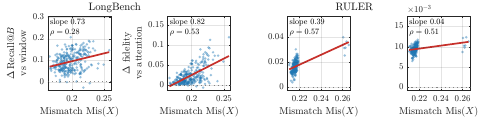}
  \caption{\textbf{Gain versus query mismatch.} Per-example gains in oracle retention and output fidelity with increasing prompt-to-future mismatch. Mistral-7B-Instruct-v0.2, $B=128$, $M=4$.}
  \label{fig:mismatch_vs_gain}
\end{figure}

\textbf{Context and budget scaling.}
RULER evaluates the same $13$ tasks across $4$K--$16$K contexts with fixed lookahead cost, where \lorekvE\ achieves the best overall average and wins $23$ of $24$ settings, with the largest gains on aggregation-style tasks (Figure~\ref{fig:ruler_main}; Tables~\ref{tab:ruler_mistral_full} and~\ref{tab:ruler_llama_full}). The advantage is largest under aggressive compression and longer contexts: on Mistral at $B=128$, the gain over AnDPro grows from $3.55$ points at $4$K to $5.85$ at $16$K, while becoming small at $B=1024$. With fixed lookahead cost, this shows that future-query estimation is most valuable when the cache is tight and prompt-side queries are less representative of later decoding needs (Figure~\ref{fig:mismatch_vs_gain}). Table~\ref{tab:ruler_ablation} further shows that multiple futures outperform a single rollout, while \lorekvD\ closely tracks \lorekvE.

\textbf{Scaling to 64K and 128K contexts.}
On Llama-3.1-8B-Instruct at $B{=}128$, with the same lookahead settings, \lorekvE\ obtains the best average at 64K ($57.75$) and 128K ($51.05$). Its margin over AnDPro grows from $3.25$ points at 16K to $8.13$ at 64K and $11.65$ at 128K (Figure~\ref{fig:ruler_long_avg}; Table~\ref{tab:ruler_llama_long}). Appendix~\ref{app:ruler_long} gives the per-task analysis.

\begin{figure}
\centering
\includegraphics[width=\textwidth]{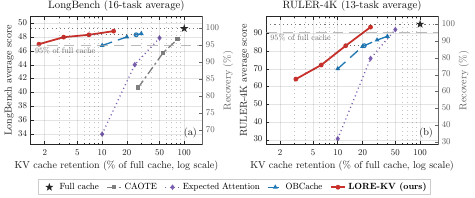}
\caption{Accuracy-memory trade-off on Llama-3.1-8B-Instruct. Higher and further left indicates
a better trade-off.}
\label{fig:external_comparison}
\vspace{-4pt}
\end{figure}

\textbf{Quality-memory trade-off.}
In the fixed $4$K setting, Figure~\ref{fig:external_comparison} shows that \lorekvE\ achieves higher quality at lower retention, outperforming OBCache \citep{gu2025obcache} at $12.5\%$ versus $20\%$ retention and Expected Attention \citep{devoto2025expected} at the same $1{,}024$-token budget. The Expected Attention comparison is contextual because it is query-agnostic by design, but follows the mismatch trend in Figure~\ref{fig:mismatch_vs_gain}. 

\subsection{Memory and Latency} \label{subsec:efficiency}
At a fixed cache budget $B$, all methods have the same steady-state decoding memory and per-token cost; LORE-KV adds a one-time lookahead and scoring overhead from $M$ trajectories of length $T$, while reliability weighting requires no additional decoding, giving \lorekvE\ and \lorekvD\ the same cost profile, as detailed in Algorithm~\ref{alg:lorekv}. Table~\ref{tab:compute_pareto} shows that LORE-KV remains above the strongest baseline at every matched lookahead-token budget, with computational overhead detailed in Appendix~\ref{app:overhead_details}. Figure~\ref{fig:latency_quality_mistral} reports the latency-quality trade-off; stage-wise LORE-KV latency is averaged over two seeds, and because the figure and Table~\ref{tab:runtime_comparison} use separate timing runs, their totals may differ slightly. With fixed $(M,T)$, lookahead and scoring costs are largely independent of $B$; the lower latency at $B{=}1024$ is consistent with runtime variation. Figure~\ref{fig:budget_efficiency} further shows that LORE-KV reaches the same accuracy with a smaller retained cache.

\begin{figure}
  \centering
  \includegraphics[width=\textwidth]{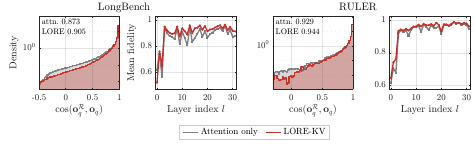}
  \caption{\textbf{Attention-output fidelity.} Similarity between retained- and full-cache outputs on future queries across layers. Mistral-7B-Instruct-v0.2, $B=128$.}
  \label{fig:cache_distribution}
\end{figure}

\subsection{Mechanism Analysis: Locating the Gain} \label{subsec:mechanism}
We analyze whether LORE-KV benefits from better alignment with future decoding queries, using retrospectively observed $\mathcal{Q}_{\mathrm{future}}$ (Appendices~\ref{app:kv_background} and~\ref{app:mechanism_protocol}). Figure~\ref{fig:montecarlo_queries} shows that response-side lookahead improves LongBench coverage of $\mathcal{Q}_{\mathrm{future}}$ from $0.785$ for the prompt window to $0.802$ at $M{=}2$ and $0.811$ at $M{=}8$, saturating near $M{=}4$; on this geometric diagnostic a single rollout ($0.786$) barely separates from the prompt window, so the coverage gain comes from spanning several plausible futures rather than from one surrogate continuation. The $M$-ablations in Tables~\ref{tab:ablation_summary} and~\ref{tab:ruler_ablation} measure the corresponding end-to-end benefit, which is already large at $M{=}1$; coverage is a nearest-neighbor statistic over query directions and therefore tracks the geometry of the estimated query set rather than the realized deletion cost (Appendix~\ref{app:query_coverage}). Correspondingly, output fidelity increases from $0.873$ to $0.905$ on LongBench and from $0.929$ to $0.944$ on RULER (Figure~\ref{fig:cache_distribution}), with gains distributed across layers and heads (Figure~\ref{fig:layer_head_analysis}).

LORE-KV's gains increase with prompt-to-future query mismatch. Defining $\operatorname{Mis}(X)=1-\operatorname{Cov}(\mathcal{Q}_{\mathrm{future}};\{\mathcal{Q}^{l,h}_{\mathcal{W}}\})$, Figure~\ref{fig:mismatch_vs_gain} shows positive Spearman correlations with oracle-retention recall ($\rho=0.28/0.57$) and output fidelity ($\rho=0.53/0.51$) on LongBench/RULER, consistent with the budget- and context-length trends in Section~\ref{subsec:ruler}, including the wider margins at 64K and 128K (Figure~\ref{fig:ruler_long_avg}; Appendix~\ref{app:ruler_long}). Figure~\ref{fig:criterion_comparison} further shows that, with prefill, particles, and selection fixed, LORE-KV better matches the future-query oracle and preserves full-cache output fidelity than CriticalKV, AnDPro, and the prompt-only control. Under identical trajectories, the two reliability variants have nearly identical averages, so entropy and disagreement weighting act as robustness corrections on top of the future-query reference rather than as a second source of gain (Appendix~\ref{app:divergence}). Additional diagnostics are provided in Appendix~\ref{app:experiments_extra}, Figures~\ref{fig:score_comparison}, \ref{fig:retained_token_composition}, \ref{fig:layer_head_analysis}, and~\ref{fig:budget_equivalence}.

\section{Conclusion} \label{sec:conclusion}
At a broader level, LORE-KV treats cache eviction as a future-use prediction problem: cached tokens should be retained based on how useful they are likely to be during generation, not only on how important they appeared during the prompt. \textbf{Limitations.} LORE-KV still incurs one-time lookahead overhead, makes irreversible eviction decisions, and does not explicitly model cross-head or cross-layer interactions. These limitations motivate adaptive lookahead, recoverable or dynamically updated caches, and globally coordinated retention. \textbf{Future work.} More broadly, this perspective points toward dynamic inference-time memory systems that adapt retention as the model's computational needs evolve. Since the benefits of future-aware eviction are strongest under tighter cache budgets and longer contexts, where the RULER margin over AnDPro reaches $11.65$ points at 128K, such systems could decide when lookahead is worth invoking, allocate extra lookahead only when future-query uncertainty is high, and revise cache decisions during generation. A practical extension is to use a cheap proxy for prompt-to-future mismatch to skip lookahead when prompt-side queries are already sufficiently informative. This would move beyond a fixed prompt-time eviction policy toward memory that tracks what information is likely to matter next.

\subsection*{AI Use Statement}

\texttt{ChatGPT} by \texttt{OpenAI} and \texttt{Claude} by \texttt{Anthropic} were used for minor editing, proofreading, and resolving \LaTeX~formatting issues in this document. \texttt{Codex} by \texttt{OpenAI} was used to assist with routine code writing, plotting, table generation, and code refactoring in the implementation of \texttt{LORE-KV}. The authors reviewed and verified all AI-assisted content and take full responsibility for the paper.

\subsection*{Ethics Statement}

This work uses publicly available models and benchmarks and involves no human subjects, personally identifiable information, or new data collection.

\subsection*{Reproducibility Statement}

LORE-KV is training-free and requires no modification to model weights. The methodology and algorithm are provided in Section~\ref{sec:methodology} and Algorithm~\ref{alg:lorekv}; hyperparameters, implementation details, evaluation settings, and ablation protocols are documented in Table~\ref{tab:hyperparams}, Appendices~\ref{app:implementation}, \ref{app:ruler_results}, and~\ref{app:mechanism_protocol}, and Tables~\ref{tab:longbench_datasets} and~\ref{tab:ablation_summary}-\ref{tab:ruler_ablation}. All methods are evaluated under a matched implementation, with backbone-specific verification described in Appendix~\ref{app:new_backbones}.

\bibliography{main}
\bibliographystyle{iclr2027_conference}

\appendix
\vspace{1.5em}
\startcontents[appendices]

\begin{center}
    {\Large\bfseries Appendix Contents}
\end{center}
\vspace{0.5em}

\begingroup
\small
\setcounter{tocdepth}{2} 
\printcontents[appendices]{}{1}{}
\endgroup

\clearpage

\section{Extended Related Work}
\label{app:extended_related_work}

\subsection{Prompt-Side Eviction and Budget Allocation}
Beyond the representative methods discussed in Section~\ref{subsec:rw_eviction}, Scissorhands uses accumulated attention statistics \citep{liu2023scissorhands}, TOVA uses current-query attention \citep{oren2024transformers}, FastGen applies head-specific retention patterns \citep{ge2024fastgen}, and CAKE improves layer-wise budget allocation \citep{qin2025cake}. HeadKV, DynamicKV, SAGE-KV, LAVa, and LU-KV further adapt retention across heads, layers, or token groups \citep{fu2024headkv,zhou2024dynamickv,wang2025sagekv, shen2025lava,tang2026lukv}. Although these approaches differ substantially in scoring and allocation, their eviction signals are primarily derived from queries or statistics available on the prompt side.

\subsection{Future-Aware and Output-Aware Eviction}
Several additional methods address robustness to unknown future queries. KVP and globally calibrated retention methods learn predictors of future token utility \citep{moschella2026learning,bui2026make}, whereas KVzip, FastKVzip, and ContrastKV seek query-agnostic or multi-query robustness \citep{kim2026kvzip,chen2026contrastkv, kim2026fastkvzip}. CriticalKV incorporates value information into token importance \citep{feng2026criticalkvoptimizingkvcache}, while CAOTE scores tokens according to attention-output error \citep{goel2025caote}. These methods motivate the two axes considered by LORE-KV: the query states against which importance is evaluated and the output distortion induced by deletion, but generally address one axis independently or replace explicit response-side sampling with learned, analytical, or query-agnostic proxies.

\subsection{Quantization and Representation Compression}
KV-cache quantization methods include QAQ, MiKV, MiniKV, RotateKV, KVTuner, MixKVQ, PM-KVQ, and DiffKV \citep{dong2024qaq,yang2024notokenleft,sharma2024minikv, su2025rotatekv,li2025kvtuner,zhang2025mixkvq,liu2025pmkvq, 10.1145/3731569.3764810}. GEAR, KVMerger, xKV, inter-layer compression, and transform-coding approaches instead exploit low-rank, mergeable, cross-layer, or coded cache representations \citep{kang2024gear,wang2024kvmerger,chang2025xkv, ma2024interlayer,staniszewski2026kv}. 

\subsection{Decode-Time and Serving-Oriented Memory Management} 
KVServe targets communication-efficient KV-cache serving \citep{liu2026kvserve}. DMS, ThinKV, RaaS, LazyEviction, R-KV, RocketKV, LongFlow, and CONF-KV instead control cache growth or retention during extended decoding \citep{lancucki2026inference,ramachandran2025thinkv,hu2025raas, zhang2025lazyevict,cai2025rkv,behnam2025rocketkv, su2026longflow,li2026confkv}. These methods address a different stage of the memory-management problem and can in principle be combined with a future-aware prompt-time eviction criterion such as LORE-KV. 

\section{Background: The KV Cache and the Eviction Problem} \label{app:kv_background}

\textbf{KV caching and memory cost.}
For $L$ layers, $H$ KV heads, head dimension $d$, sequence length $n$, batch size $b$, and $p$ bytes per element, the KV cache requires $\mathrm{Mem}_{\mathrm{KV}}=2bLHdnp$ bytes, where the factor $2$ accounts for keys and values; hence, its memory grows linearly with context length. For example, a 70B-class model with $L=80$, $H=8$, $d=128$, $n=131{,}072$ (128K tokens), $b=1$, and $p=2$ bytes (BF16) requires approximately $42.9$ GB of KV-cache memory, about $31\%$ of the roughly $140$ GB required by its BF16 model weights. Our 128K RULER evaluation (Table~\ref{tab:ruler_llama_long}) works at this context length, where a budget of $B{=}128$ keeps about $0.1\%$ of the prompt. Given a cache budget $B\ll n$, let $\mathcal{R}^{l,h}\subseteq\{1,\ldots,n\}$ denote the positions retained by layer $l$ and KV head $h$, with $|\mathcal{R}^{l,h}|\le B_{l,h}$. Defining $\delta(\mathbf{q},\mathcal{R})=\mathrm{dist}\!\left(\mathrm{output}(\mathbf{q};\text{full cache}),\mathrm{output}(\mathbf{q};\text{retained cache})\right)$ as the distortion induced by eviction for a future query $\mathbf{q}$, the ideal retained cache is $\mathcal{R}^{\star}=\arg\min_{\{\mathcal{R}^{l,h}:|\mathcal{R}^{l,h}|\le B_{l,h}\}}\mathbb{E}_{\mathbf{q}\sim\mathcal{Q}_{\mathrm{future}}}\!\left[\delta(\mathbf{q},\mathcal{R})\right]$. Thus, eviction seeks to retain the KV entries that best preserve full-cache behavior over future decoding queries; Section~\ref{subsec:mechanism} evaluates this approximation against the actual future queries observed after generation.

\section{Method Details} \label{app:method_details}
This section provides the implementation and computational details of LORE-KV, including protected-token handling, grouped-query attention, reliability weighting, and efficient perturbation scoring. Algorithm~\ref{alg:lorekv} summarizes the full LORE-KV pipeline. Starting from the prefilled KV cache, LORE-KV samples short lookahead continuations to obtain plausible future queries, reliability-weights these trajectories, scores candidate KV entries by their projected deletion effect, and retains the highest-scoring entries under the cache budget.

\subsection{Implementation: Protected Set, Grouped-Query Attention, Fused Kernels}
\label{app:implementation}
\label{app:setup_details}
\label{app:gqa}
\label{app:kernels}

\textbf{Protected positions.} The optional set $\mathcal{B}$ contains prompt-structure markers, chat-template delimiters (\verb!<|system|>!, \verb!<|user|>!, \verb!<|assistant|>!, \verb![INST]!, \verb![/INST]!), tool-call markers, and document separators identified directly from the tokenized prompt and always retained; if none are present, $\mathcal{B}=\emptyset$. Following prior work \citep{xiao2023streamingllm,li2024snapkv}, this prevents instructions and prompt boundaries from being removed under aggressive eviction. 

\textbf{Grouped-query attention.} Under GQA/MQA, multiple query heads share each KV head. In our implementation, we therefore use all query heads associated with KV head $h$, denoted by $\mathcal{G}(h)$, to score each cached pair $(\mathbf{k}^{l,h}_{i},\mathbf{v}^{l,h}_{i})$ by aggregating their deletion effects from Eq.~\ref{eq:deletion_effect}. Throughout, $H$ denotes the number of KV heads, and we assign cache budgets at the KV-head level.

\textbf{Fused attention kernels.} Since kernels such as FlashAttention do not materialize the full $n\times n$ attention matrix required by perturbation scoring, we explicitly compute attention only for the scoring-query set $\mathcal{Q}^{l,h}_{\mathrm{score}}$, whose size $|\mathcal{Q}^{l,h}_{\mathrm{score}}|=w+MT\ll n$, against the cached keys. The fused kernel therefore remains unchanged for the rest of prefill and final decoding, making eviction scoring a small side computation that never requires full-matrix attention on the main inference path.

\subsection{Relation to Prior Surrogate-Query and Output-Aware Methods}
\label{app:lookahead_details}

The $M{=}1$ setting already captures most of LORE-KV's gain, demonstrating that correcting the prompt-to-future query mismatch is the primary driver of performance. Increasing to $M{>}1$ further improves robustness by covering multiple plausible futures and reducing sensitivity to any single rollout through reliability-weighted aggregation, while $\lambda_L{=}0$ provides the prompt-only control. Together, these results show that LORE-KV is both effective with minimal lookahead and increasingly robust as future-query coverage expands.

\begin{algorithm}[t]
\caption{LORE-KV eviction. Lookahead trajectories are used only for scoring and discarded before decoding, leaving model weights and decoding unchanged. Scoring uses only $w+MT\ll n$ queries.}
\label{alg:lorekv}
\begin{algorithmic}[1]
\Require frozen LLM $f_\theta$; prompt $X=(x_1,\ldots,x_n)$; window $\mathcal{W}$ of size $w$; particles $M$, horizon $T$; layer budgets $\{B_l\}_{l=1}^{L}$; reliability coefficients $(\beta,\gamma,\lambda_D)$, mixing weight $\lambda_L$, guard $\epsilon$; plug-in switches $(\alpha,\kappa,\tau,\lambda_k,\lambda_v)$, all $0$ for the core method (Table~\ref{tab:hyperparams})
\Ensure compressed KV cache $(\mathbf{K}^{\star},\mathbf{V}^{\star})$
\Statex
\State $\big(\{\mathbf{K}^{l,h},\mathbf{V}^{l,h}\},\{\mathcal{Q}^{l,h}_{\mathcal{W}}\}\big)\gets\operatorname{Prefill}(f_\theta,X)$
\State $\mathcal{P}\gets\{1\}\cup\mathcal{W}\cup\mathcal{B}$;\quad $\mathcal{C}\gets\{1,\ldots,n\}\setminus\mathcal{P}$ \Comment{protected and candidate positions}

\algstage{Stage 1: Monte Carlo future-query estimation}{Sec.~\ref{subsec:lookahead}}
\For{$m=1,\ldots,M$ \textbf{in parallel}}
  \State $Y^{(m)}\sim f_\theta(\cdot\mid X)$ with $|Y^{(m)}|=T$ \Comment{one plausible continuation}
  \State cache $\mathcal{Q}^{(m),l,h}_{\mathrm{look}}$ for all $(l,h)$, and $\{c^{(m)}_u,H^{(m)}_u\}_{u=1}^{T}$ \Comment{kept for scoring and weighting}
\EndFor
\State $\mathcal{Q}^{l,h}_{\mathrm{score}}\gets\mathcal{Q}^{l,h}_{\mathcal{W}}\cup\bigcup_{m=1}^{M}\mathcal{Q}^{(m),l,h}_{\mathrm{look}}$ \Comment{$w+MT$ scoring queries in total}

\algstage{Stage 2: particle reliability weighting}{Sec.~\ref{subsec:confidence}}
\State $\hat D_m\gets\big(D_m-\bar D\big)/\operatorname{sd}(D)$ for $M\ge3$, else $\hat D_m\gets0$ \Comment{drift from the ensemble}
\State $\bar c_m\gets\tfrac1T\sum_{u}c^{(m)}_u$, \ \ $\bar H_m\gets\tfrac1T\sum_{u}H^{(m)}_u$ \Comment{per-step means, not sequence totals}
\State $R_m\gets\beta\,\bar c_m-\gamma\,\bar H_m-\lambda_D\,\hat D_m$; \ \ $\boldsymbol{\omega}\gets\operatorname{softmax}\big((R_m)_{m=1}^{M}\big)$ \Comment{soft trajectory weights}

\algstage{Stage 3: projected counterfactual scoring and selection}{Secs.~\ref{subsec:deletion}--\ref{subsec:core_score}}
\For{layer $l=1,\ldots,L$}
  \For{KV head $h=1,\ldots,H$}
    \Comment{under GQA, over the query group $\mathcal{G}(h)$}
    \For{candidate $i\in\mathcal{C}$ and scoring query $\mathbf{q}\in\mathcal{Q}^{l,h}_{\mathrm{score}}$}
      \State $(a^{l,h}_{q,i},\mathbf{o}^{l,h}_{q})\gets\operatorname{Attn}(\mathbf{q},\mathbf{K}^{l,h},\mathbf{V}^{l,h})$ \Comment{full-cache weight and output}
      \State $\Delta\mathbf{o}^{l,h}_{q,i}\gets\big[a^{l,h}_{q,i}/(1-a^{l,h}_{q,i}+\epsilon)\big]\,\big(\mathbf{v}^{l,h}_{i}-\mathbf{o}^{l,h}_{q}\big)$ \Comment{output shift if $i$ is evicted}
      \State $e^{l,h}_{q,i}\gets\big\|\mathbf{W}^{O}_{l,h}\,\Delta\mathbf{o}^{l,h}_{q,i}\big\|_2^2$ \Comment{that shift, in the residual stream}
    \EndFor
\State $S^{l,h}_{i}
\gets
\psi(\mathcal{E}^{l,h}_{i,\mathcal{W}})
+
\lambda_L
\sum_{m=1}^{M}
\omega_m\,
\psi(\mathcal{E}^{l,h}_{i,m})
\quad \forall i\in\mathcal{C}$
\Comment{combine prompt and weighted lookahead importance}
  \EndFor
  \State $E_{l,h}\gets\sum_{i\in\mathcal{C}}S^{l,h}_{i}$; \ \ $\{B_{l,h}\}_{h=1}^{H}\gets\operatorname{Alloc}\big(B_l,\{E_{l,h}\}_{h=1}^{H};\tau\big)$ \Comment{budget split across heads}
  \For{KV head $h=1,\ldots,H$}
    \State $\mathcal{R}^{l,h}_{\mathrm{final}}\gets\mathcal{P}\cup\operatorname{Select}\big(\mathbf{S}^{l,h},B_{l,h}\big)$ \Comment{keep the top-scoring candidates}
    \State $(\mathbf{K}^{l,h}_{\star},\mathbf{V}^{l,h}_{\star})\gets\operatorname{Gather}\big(\mathbf{K}^{l,h},\mathbf{V}^{l,h},\mathcal{R}^{l,h}_{\mathrm{final}}\big)$ \Comment{build the compressed cache}
  \EndFor
\EndFor
\State discard all lookahead states $\{\mathcal{Q}^{(m),l,h}_{\mathrm{look}}\}_{m=1}^{M}$
\State \Return $(\mathbf{K}^{\star},\mathbf{V}^{\star})$ \Comment{decoding uses the compressed cache only}
\end{algorithmic}
\end{algorithm}

\subsection{Reliability Weighting: Derivations, Calibration, and Edge Cases}
\label{app:reliability_details}

This section gives the statistical interpretation and implementation
details of the reliability weighting in Section~\ref{subsec:confidence}.

\textbf{Centered typicality and scale.}
For a sampled token $y_u\sim q(\cdot\mid h_u)$, the centered score
of Eq.~\ref{eq:centered_score} satisfies
\[
\mathbb{E}[c_u\mid h_u]
=
\sum_y q(y\mid h_u)\log q(y\mid h_u)+H(q(\cdot\mid h_u))
=0.
\]
Hence $\mathbb{E}[\bar c_m]=0$, so $\bar c_m$ measures how typical a sampled trajectory is relative to the model's local uncertainty. Positive values indicate that the sampled tokens are more likely than expected, while $\bar H_m$ separately captures how uncertain the model is along that trajectory.

The identity
\[
\frac{1}{T}\sum_{u=1}^{T}
\log q(y_u^{(m)}\mid h_u^{(m)})
=
\bar c_m-\bar H_m, 
\]
also makes the parameterization of Eq.~\ref{eq:confidence}
transparent. Uniform weighting corresponds to
$(\beta,\gamma,\lambda_D)=(0,0,0)$, centered-typicality-only
weighting to $(1,0,0)$, and raw log-probability weighting to
$(1,1,0)$. The reported reliability settings are given in
Table~\ref{tab:hyperparams}.

Because $\{c_u\}$ are martingale differences, for $u<v$,
\[
\mathbb{E}[c_uc_v]
=
\mathbb{E}\!\left[c_u\,\mathbb{E}[c_v\mid h_v]\right]
=0.
\]
For a fixed-length trajectory of length $T$,
\begin{equation}
\operatorname{Var}(\bar c_m)
=
\frac{1}{T^2}\sum_{u=1}^{T}\mathbb{E}[V_u],
\qquad
V_u=
\operatorname{Var}_{y\sim q(\cdot\mid h_u)}
[\log q(y\mid h_u)],
\label{eq:varentropy}
\end{equation}
where $V_u$ is the varentropy. For a trajectory that terminates after
$T_m<T$ steps, $T$ is replaced by $T_m$ in the corresponding
per-trajectory averages. The optional normalization
\[
\tilde c_m
=
\frac{\bar c_m}{\sqrt{\bar V_m/T_m}},
\qquad
\bar V_m=\frac{1}{T_m}\sum_{u=1}^{T_m}V_u^{(m)},
\]
normalizes $\bar c_m$ by its trajectory-level standard-error scale;
it is evaluated only as an ablation and is not part of the default
method.

\textbf{Query-space disagreement.}
Predictive entropy is defined in token-probability space, whereas
eviction ultimately depends on query directions. We therefore measure
whether trajectory $m$ follows an atypical query path relative to the
other sampled futures. Let $H_q$ denote the number of query heads and
$T_{mr}=\min(T_m,T_r)$ the common length of trajectories $m$ and $r$.
We define
\begin{equation}
D_m
=
\frac{1}{M-1}\sum_{r\neq m}
\frac{1}{T_{mr}LH_q}
\sum_{u=1}^{T_{mr}}
\sum_{l=1}^{L}
\sum_{g=1}^{H_q}
\left[
1-\cos\!\left(
\mathbf q_u^{(m),l,g},
\mathbf q_u^{(r),l,g}
\right)
\right].
\label{eq:disagreement}
\end{equation}
Queries compared at step $u$ occupy the same absolute position
$n+u$ and therefore share the same rotary phase. We standardize
$D_m$ across particles as
\[
\hat D_m=
\begin{cases}
(D_m-\bar D)/\operatorname{sd}(D),
& M\ge3 \text{ and } \operatorname{sd}(D)>0,\\[2mm]
0, & \text{otherwise},
\end{cases}
\qquad
\bar D=\frac1M\sum_{m=1}^{M}D_m.
\]
This definition covers both the $M\le2$ degeneracy and the rare case
in which all particles have identical disagreement scores. The
computation reuses the stored lookahead queries and therefore requires
no additional decoding. Since $\bar H_m$ is measured in nats whereas
$\hat D_m$ is standardized, their coefficients are calibrated on
their realized across-particle scales rather than compared by raw
magnitude.

\textbf{Estimator interpretation.}
No separate softmax temperature is needed: multiplying
Eq.~\ref{eq:confidence} by a concentration parameter $\eta$ is
equivalent to rescaling $(\beta,\gamma,\lambda_D)$. With uniform weights, the lookahead term is the ordinary Monte Carlo estimate of the expected deletion effect under $p_\theta(Y\mid X)$. When $\lambda_D=0$, the normalized weighted estimate is a self-normalized importance estimator whose large-$M$ target is
\[
\pi(Y)
\propto
p_\theta(Y\mid X)
\exp\!\left(
\beta\bar c(Y)-\gamma\bar H(Y)
\right).
\]
For any trajectory statistic $f$, a first-order expansion around
$(\beta,\gamma)=(0,0)$ gives
\[
\mathbb{E}_{\pi}[f]
=
\mathbb{E}_{p_\theta}[f]
+
\beta\,\operatorname{Cov}_{p_\theta}(f,\bar c)
-
\gamma\,\operatorname{Cov}_{p_\theta}(f,\bar H)
+
O\!\left(\|(\beta,\gamma)\|^2\right).
\]
Centering therefore makes the typicality channel zero-mean under the
sampling distribution, but does not make the finite-$M$ weighted
estimator unbiased. When $\lambda_D>0$, $\hat D_m$ depends on the
entire particle ensemble, so the resulting rule is an
interacting-particle estimator rather than finite-$M$ importance
weighting for a fixed single-trajectory target.

\textbf{Failure modes and safeguards.}
Reliability weighting is introduced as a robustness safeguard because future-query estimation introduces an additional source of uncertainty: sampled continuations can be plausible yet unrepresentative of the queries encountered during actual decoding. Existing KV-cache methods typically rely on prompt-side or single-surrogate query references without explicitly assessing their reliability. LORE-KV addresses this gap by aggregating multiple plausible futures and softly reweighting them using complementary entropy and query-space disagreement signals, while $\lambda_L$ limits the influence of imperfect lookahead. This reliability-aware formulation makes future-aware eviction robust to sampling variability rather than assuming that every predicted future is equally trustworthy. Appendix~\ref{app:divergence} quantifies how these reliability signals relate to oracle particle utility, and separates that effect from the coverage gain of Figure~\ref{fig:montecarlo_queries}.

\begin{table*}[t]
\centering
\caption{Average inference time per sample (seconds). Lower is better. LORE-KV uses fixed lookahead compute, so non-monotonic latency across budgets, including at $B{=}1024$, is consistent with runtime variation.}
\label{tab:runtime_comparison}
\small
\setlength{\tabcolsep}{5.0pt}
\renewcommand{\arraystretch}{1.08}
\begin{tabular}{lllcccc}
\toprule
\textbf{Backbone} & \textbf{Dataset} & \textbf{Method}
& \textbf{128} & \textbf{256} & \textbf{512} & \textbf{1024} \\
\midrule

\multirow{9}{*}{Mistral}
& \multirow{3}{*}{LongBench}
& AnDPro      & \second{7.8799} & \second{7.4885} & \second{7.4355} & \second{7.5338} \\
&& CriticalKV & \best{6.6867} & \best{6.3991} & \best{6.3982} & \best{6.5276} \\
&& \textbf{\lorekvE} & 17.5342 & 19.5496 & 19.6767 & 17.7930 \\
\cmidrule(lr){2-7}

& \multirow{3}{*}{RULER}
& AnDPro      & \second{6.2360} & \second{4.5775} & \second{5.9707} & \second{3.1746} \\
&& CriticalKV & \best{6.1907} & \best{4.1849} & \best{4.9899} & \best{2.4039} \\
&& \textbf{\lorekvE} & 13.6877 & 11.1720 & 14.5537 & 7.2315 \\
\cmidrule(lr){2-7}

& \multirow{3}{*}{NIAH}
& AnDPro      & \second{2.9842} & \second{2.8559} & \second{2.8779} & \second{2.9169} \\
&& CriticalKV & \best{2.3514} & \best{2.1441} & \best{2.1109} & \best{2.1961} \\
&& \textbf{\lorekvE} & 7.3873 & 7.4716 & 7.4072 & 7.3407 \\
\midrule

\multirow{9}{*}{Llama}
& \multirow{3}{*}{LongBench}
& AnDPro      & \second{8.3345} & \second{8.2447} & \second{8.1550} & \second{8.0648} \\
&& CriticalKV & \best{7.0724} & \best{7.0455} & \best{7.0165} & \best{6.9867} \\
&& \textbf{\lorekvE} & 12.1815 & 14.4038 & 15.8882 & 13.1342 \\
\cmidrule(lr){2-7}

& \multirow{3}{*}{RULER}
& AnDPro      & \second{4.3311} & \second{4.0226} & \second{3.7141} & \second{3.4060} \\
&& CriticalKV & \best{4.2994} & \best{3.6776} & \best{3.1042} & \best{2.5791} \\
&& \textbf{\lorekvE} & 6.6410 & 6.8587 & 6.3249 & 5.4202 \\
\cmidrule(lr){2-7}

& \multirow{3}{*}{NIAH}
& AnDPro      & \second{2.6046} & \second{2.5137} & \second{2.5269} & \second{2.6172} \\
&& CriticalKV & \best{1.9267} & \best{1.9222} & \best{1.9450} & \best{1.8561} \\
&& \textbf{\lorekvE} & 7.0298 & 6.9553 & 7.0076 & 6.9816 \\
\bottomrule
\end{tabular}
\end{table*}
\begin{figure}[t]
\centering
\includegraphics[width=\textwidth]{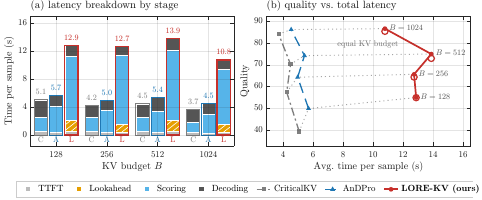}
\caption{\textbf{Latency-quality trade-off.} (a) Stage-wise latency averaged over two seeds and (b) quality versus total latency across cache budgets on Mistral-7B-Instruct-v0.2, averaged over LongBench, RULER, and NIAH.}
\label{fig:latency_quality_mistral}
\end{figure}

\subsection{Computational Overhead} \label{app:overhead_details}
LORE-KV restricts scoring to at most $w+MT\ll n$ prompt-window and lookahead queries, keeping the additional computation localized. The entropy statistics for \lorekvE\ are obtained directly from the sampling logits, while \lorekvD\ adds only pairwise cosine disagreement over already-materialized query states, requiring $O(M^2LTH_qd)$ arithmetic and no additional model decoding. Table~\ref{tab:runtime_comparison} shows that \lorekvE\ incurs approximately $1.46\times$--$2.77\times$ the per-sample wall-clock time of AnDPro across the evaluated settings; \lorekvD\ has the same lookahead decoding cost with only the additional query-space reduction. Since $(M,T)$ is fixed, this one-time overhead is independent of the retained cache budget $B$, explaining its non-monotonic variation with $B$. Importantly, the robustness of the resulting gains is examined through matched lookahead-token budget controls and analyses across cache budgets, context lengths, and prompt-to-future mismatch in Sections~\ref{subsec:longbench}-\ref{subsec:mechanism}, while the single-surrogate score-level control is reported in Appendix~\ref{app:score_shift}. Figure~\ref{fig:latency_quality_mistral} separates LORE-KV's one-time lookahead/scoring overhead from the shared decoding cost and shows that this extra compute consistently translates into higher quality at matched cache budgets.
\begin{table}[t]
\caption{LORE-KV hyperparameters and default settings.}
\label{tab:hyperparams}
\centering
\begin{tabular}{llc}
\toprule
Symbol & Meaning & Default \\
\midrule
\multicolumn{3}{@{}l}{\textit{(a) Core method}} \\
$w$ & prompt-end observation-window size & $32$ \\
$M$ & number of lookahead particles & $4$ \\
$T$ & lookahead particle length & $8$ \\
$\lambda_L$ & lookahead mixing coefficient in Eq.~\ref{eq:final_score} & $1.0$ \\
$\epsilon$ & renormalization guard in Eq.~\ref{eq:deletion_effect} & $10^{-6}$ \\
& chunk size (adjacent-token grouping) & $4$ \\
\midrule
\multicolumn{3}{@{}l}{\textit{(b) Plug-in switches (Appendix~\ref{sec:ablation_components})}} \\
$\kappa,\alpha$ & robust aggregation & $0,\,0$ \\
$\tau$ & adaptive head budgets & $0$ \ \ (ablation: $1$) \\
$\lambda_k,\lambda_v$ & diversity selection & $0$ \ \ (ablation: $0.1$) \\
\midrule
\multicolumn{3}{@{}l}{\textit{(c) Reliability settings in Eq.~\ref{eq:confidence}}} \\
\textbf{\lorekvE}
& log-probability $+$ entropy penalty
& $(1,\,1.3,\,0)$ \\
\textbf{\lorekvD}
& log-probability $+$ disagreement penalty
& $(1,\,1,\,0.3)$ \\
Log-probability only
& control
& $(1,\,1,\,0)$ \\
Entropy $+$ disagreement
& control
& $(1,\,1.3,\,0.3)$ \\
\bottomrule
\end{tabular}
\end{table}

\begin{table}[t]
\caption{LongBench English datasets, the abbreviations used in all LongBench results tables, metrics, average input lengths, and sample counts \citep{bai2024longbench}.}
\label{tab:longbench_datasets}
\centering
\small
\setlength{\tabcolsep}{4pt}
\begin{tabular}{@{}llllrlr@{}}
\toprule
Task & Abbr. & Task type & Metric & Avg.\ len & Language & \#Samples \\
\midrule
NarrativeQA & NQA & Single-Doc. QA & F1 & 18{,}409 & EN & 200 \\
Qasper & Qasp & Single-Doc. QA & F1 & 3{,}619 & EN & 200 \\
MultiFieldQA-en & MF & Single-Doc. QA & F1 & 4{,}559 & EN & 150 \\
HotpotQA & HQA & Multi-Doc. QA & F1 & 9{,}151 & EN & 200 \\
2WikiMultihopQA & 2WQA & Multi-Doc. QA & F1 & 4{,}887 & EN & 200 \\
MuSiQue & Musq & Multi-Doc. QA & F1 & 11{,}214 & EN & 200 \\
GovReport & GovR & Summarization & Rouge-L & 8{,}734 & EN & 200 \\
QMSum & QMS & Summarization & Rouge-L & 10{,}614 & EN & 200 \\
MultiNews & MN & Summarization & Rouge-L & 2{,}113 & EN & 200 \\
TREC & TREC & Few-shot Learning & Accuracy & 5{,}177 & EN & 200 \\
TriviaQA & TQA & Few-shot Learning & F1 & 8{,}209 & EN & 200 \\
SAMSum & SAMS & Few-shot Learning & Rouge-L & 6{,}258 & EN & 200 \\
PassageCount & PC & Synthetic & Accuracy & 11{,}141 & EN & 200 \\
PassageRetrieval-en & PR & Synthetic & Accuracy & 9{,}289 & EN & 200 \\
LCC & LCC & Code & Edit Sim & 1{,}235 & Python/C\#/Java & 500 \\
RepoBench-P & RB-P & Code & Edit Sim & 4{,}206 & Python/Java & 500 \\
\bottomrule
\end{tabular}
\end{table}

\begin{table}[t]
\caption{LongBench results on Llama-3.1-8B-Instruct. Dataset abbreviations follow Table~\ref{tab:longbench_datasets}.}
\label{tab:longbench_llama}
\centering
\footnotesize
\setlength{\tabcolsep}{2pt}
\begin{adjustbox}{max width=\textwidth}
\begin{tabular}{@{}l*{17}{c}@{}}
\toprule
& \multicolumn{3}{c}{Single-Doc QA} & \multicolumn{3}{c}{Multi-Doc QA}
& \multicolumn{3}{c}{Summ.} & \multicolumn{3}{c}{Few-shot}
& \multicolumn{2}{c}{Synth.} & \multicolumn{2}{c}{Code} & \\
\cmidrule(lr){2-4}\cmidrule(lr){5-7}\cmidrule(lr){8-10}
\cmidrule(lr){11-13}\cmidrule(lr){14-15}\cmidrule(lr){16-17}
Method & NQA & Qasp & MF & HQA & 2WQA & Musq
& GovR & QMS & MN & TREC & TQA & SAMS
& PC & PR & LCC & RB-P & Avg. \\
\midrule

Full Cache
& 30.12 & 46.60 & 56.41 & 58.10 & 49.01 & 32.52
& 34.04 & 25.37 & 27.06 & 73.00 & 91.90 & 43.46
& 7.06 & 100.00 & 62.13 & 51.80 & 49.29 \\

\midrule
\multicolumn{18}{c}{\textit{Budget} $B=128$} \\
\midrule

SnapKV
& 25.54 & 31.40 & 50.99 & 55.17 & 45.50 & 27.88
& 20.77 & 23.41 & 21.05 & 47.50 & 90.46 & 40.76
& 8.00 & \second{99.50} & 56.88 & 46.50 & 43.21 \\

PyramidKV
& \second{27.83} & 32.98 & 51.45 & 56.72 & 44.67 & 30.54
& 21.89 & 22.90 & 21.67 & 63.00 & 90.45 & 40.06
& 8.00 & \best{100.00} & 55.37 & 43.79 & 44.46 \\

Ada-KV
& 25.06 & 34.50 & 52.27 & 56.29 & 47.57 & 29.06
& 21.17 & 23.36 & 21.51 & 62.00 & \best{92.17} & 40.84
& 8.00 & \second{99.50} & 58.45 & 48.72 & 45.03 \\

CriticalKV
& 26.00 & 34.96 & 51.79 & 56.01 & 46.53 & 28.95
& 21.31 & 23.14 & 21.33 & 62.00 & 89.51 & 40.74
& 7.59 & 97.89 & 57.85 & 48.28 & 44.62 \\

AnDPro
& 25.25 & 35.85 & \second{52.85} & 54.35 & 44.41 & 27.51
& 22.82 & 23.60 & 22.20 & 69.00 & \second{92.00}
& 41.50 & 7.02 & \best{100.00} & \best{62.26}
& \best{56.34} & 46.06 \\

LaProx
& 25.78 & 31.38 & \best{52.97} & 55.17 & 44.95 & 29.46
& 21.66 & 23.66 & 21.47 & 59.00 & 89.94 & 41.09
& \best{8.89} & \second{99.50} & 61.91
& \second{56.16} & 45.19 \\

LAQ++(8)
& \best{28.65} & 10.65 & 26.04 & 24.23 & 21.56 & 15.67
& \second{23.50} & \second{23.74} & 22.75
& \best{72.00} & 91.95 & \best{42.19}
& \second{8.37} & 94.81 & \second{62.18}
& 50.95 & 38.70 \\

\textbf{\lorekvE}
& 27.38 & \best{37.66} & 51.14 & \second{57.37}
& \second{48.47} & \second{30.55} & \best{23.77}
& 23.60 & \second{23.14} & 68.00 & 90.46
& \second{41.77} & 7.34 & 97.90 & 58.99 & 50.15
& \second{46.11} \\

\textbf{\lorekvD}
& 27.33 & \second{37.50} & 51.81 & \best{57.73}
& \best{48.72} & \best{30.63} & 23.12
& \best{24.25} & \best{23.62}
& \second{69.50} & 90.11 & 41.69
& 7.98 & 97.02 & 59.09 & 50.17
& \best{46.27} \\

\midrule
\multicolumn{18}{c}{\textit{Budget} $B=256$} \\
\midrule

SnapKV
& 26.92 & 38.19 & 52.62 & 57.03 & 46.75 & 29.63
& 23.07 & 23.92 & 22.82 & 58.50 & 91.83 & 40.81
& 7.75 & 99.50 & 59.95 & 48.63 & 45.50 \\

PyramidKV
& \second{28.46} & 39.36 & 54.30 & 56.77 & 45.50
& 30.74 & 23.84 & 23.88 & 22.91 & 69.00 & 91.09 & 40.74
& \second{7.88} & 99.50 & 56.94 & 45.18 & 46.01 \\

Ada-KV
& 27.04 & 39.24 & \second{54.41} & 57.36
& 47.53 & 31.35 & 23.06 & 23.81 & 23.43
& 69.00 & \best{92.50} & 41.23 & 7.67
& \best{100.00} & 61.45 & 49.52 & 46.79 \\

CriticalKV
& 25.45 & 39.60 & 52.90 & 55.72 & 46.05 & 29.97
& 23.10 & 23.77 & 22.77 & 68.00 & 90.48 & 41.08
& 7.48 & 97.55 & 60.19 & 48.25 & 45.77 \\

AnDPro
& 27.23 & 42.01 & \best{54.62} & \best{57.87}
& 47.43 & 31.92 & 23.53 & 24.22 & 23.54
& \second{71.50} & \second{92.37} & 41.47
& 7.66 & \second{99.81} & 62.45 & 51.67 & 47.46 \\

LaProx
& 28.02 & 38.56 & 53.32 & 55.23 & 45.72 & 30.83
& 24.84 & 23.78 & 23.92 & 67.50 & 92.12
& \second{42.40} & \best{8.56} & 99.50
& \second{63.57} & \best{57.61} & 47.22 \\

LAQ++(8)
& \best{30.25} & 12.43 & 26.63 & 25.77 & 22.83 & 18.45
& \second{25.07} & 23.67 & 23.75
& \best{72.00} & 91.97 & \best{43.17}
& 6.93 & 94.59 & \best{64.07}
& \second{53.38} & 39.69 \\

\textbf{\lorekvE}
& 28.26 & \best{44.01} & 53.43 & 57.73
& \second{48.78} & \second{32.30} & \best{25.63}
& \best{24.93} & \second{24.10} & 71.00 & 91.03
& 39.98 & 7.39 & 98.52 & 61.70 & 51.28
& \best{47.50} \\

\textbf{\lorekvD}
& 28.33 & \second{42.69} & 52.66 & \second{57.75}
& \best{48.86} & \best{32.34} & 24.95
& \second{24.87} & \best{24.16}
& 70.00 & 91.09 & 40.95
& 7.77 & 99.73 & 61.61 & 51.84
& \second{47.48} \\

\midrule
\multicolumn{18}{c}{\textit{Budget} $B=512$} \\
\midrule

SnapKV
& 28.40 & 41.52 & 55.40 & 57.66 & 47.48 & 30.95
& 24.92 & 24.46 & 24.54 & 68.50 & \second{92.33}
& 41.98 & 7.33 & 99.50 & 61.25 & 50.39 & 47.29 \\

PyramidKV
& 29.15 & 43.04 & \second{55.55} & 57.71 & 47.67
& 31.27 & 26.05 & 24.08 & 24.54 & 70.50
& \best{92.67} & 41.52 & 7.33 & \best{100.00}
& 59.96 & 47.51 & 47.41 \\

Ada-KV
& 26.92 & 42.68 & 55.20 & 57.47 & 48.06 & 31.27
& 24.95 & 24.17 & 24.40 & \second{71.00} & 92.14
& 42.35 & 7.33 & \best{100.00}
& 62.26 & 50.62 & 47.55 \\

CriticalKV
& 26.00 & 42.53 & 53.90 & 56.62 & 46.76 & 31.13
& 24.26 & 23.76 & 23.81 & 70.00 & 89.56 & 40.75
& 6.92 & 97.20 & 60.51 & 49.74 & 46.47 \\

AnDPro
& 27.97 & \best{44.15} & 55.49 & \second{58.16}
& 48.19 & 31.92 & 24.80 & 24.50 & 24.77
& \best{72.50} & 91.76 & 41.99 & 7.31
& \second{99.78} & 62.54 & 52.00 & 47.99 \\

LaProx
& \second{29.82} & 43.46 & 54.33 & 55.44 & 46.16 & 30.83
& 27.08 & 24.10 & \best{25.30} & 70.50 & 91.89
& \second{42.48} & \second{8.40} & 99.50
& \second{64.10} & \best{58.16} & \second{48.23} \\

LAQ++(8)
& 29.64 & 13.22 & 26.79 & 27.58 & 23.49 & 18.63
& 26.94 & 24.04 & \second{25.21}
& \best{72.50} & 92.25 & \best{42.83}
& \best{8.43} & 96.25 & \best{65.00}
& \second{53.37} & 40.39 \\

\textbf{\lorekvE}
& 29.77 & 43.61 & 55.22 & \best{58.32}
& \second{48.68} & \best{32.73} & \second{27.40}
& \second{24.96} & 25.02 & \best{72.50} & 91.50
& 40.50 & 7.46 & 99.52 & 60.38 & 51.25 & 48.05 \\

\textbf{\lorekvD}
& \best{30.91} & \second{43.68} & \best{55.88} & 57.58
& \best{49.38} & \second{32.71} & \best{28.22}
& \best{25.34} & 24.94 & \best{72.50} & 91.22
& 40.87 & 7.18 & 99.46
& 61.02 & 51.29 & \best{48.26} \\

\midrule
\multicolumn{18}{c}{\textit{Budget} $B=1024$} \\
\midrule

SnapKV
& 28.32 & \second{45.37} & \best{56.81} & 58.16
& 48.28 & 31.89 & 26.91 & 24.58 & 25.82 & 70.50
& 92.22 & 43.08 & 6.96 & \best{100.00}
& 62.46 & 51.50 & 48.30 \\

PyramidKV
& 29.32 & 44.57 & 56.14 & 58.39 & 48.36
& 31.93 & 28.20 & 23.78 & 26.00 & 70.50
& \best{92.41} & 42.46 & 6.88 & \best{100.00}
& 61.33 & 49.12 & 48.09 \\

Ada-KV
& 28.52 & \best{46.01} & 55.72 & 57.85 & 48.10
& 32.02 & 26.98 & 24.72 & 25.96 & 72.00
& 91.97 & 42.55 & 6.96 & \best{100.00}
& 61.98 & 51.52 & 48.30 \\

CriticalKV
& 27.27 & 44.74 & 54.24 & 56.17 & 46.95
& 31.65 & 26.35 & 23.91 & 24.93 & 70.00
& 89.08 & 41.00 & 6.70 & 96.86 & 59.85 & 49.75
& 46.84 \\

AnDPro
& \best{31.36} & 43.42 & 55.10 & 55.05 & 45.93
& 29.79 & 28.46 & 24.70 & \second{26.56} & 72.50
& 91.42 & \best{43.57} & 7.39 & \second{99.50}
& \second{64.86} & \best{59.01} & 48.66 \\

LaProx
& 31.08 & 44.45 & 55.64 & 55.82 & 45.98
& 31.96 & \best{29.22} & 24.42 & 26.36 & 71.50
& 91.75 & \second{43.32} & \best{8.23} & \second{99.50}
& \best{65.01} & \second{58.69} & \best{48.94} \\

\textbf{\lorekvE}
& 29.99 & 44.83 & \second{56.19} & \second{58.60}
& \best{49.01} & \best{33.02} & 28.90
& \best{25.49} & 26.38 & \second{73.00}
& \second{92.35} & 42.24 & \second{7.50}
& \second{99.50} & 61.31 & 51.60 & 48.74 \\

\textbf{\lorekvD}
& \second{31.19} & 44.92 & 56.02 & \best{58.96}
& \second{48.91} & \second{32.79} & \second{28.99}
& \second{25.42} & \best{26.65} & \best{73.50}
& 92.23 & 42.40 & 7.20
& 98.75 & 60.55 & 52.03
& \second{48.78} \\
\bottomrule
\end{tabular}
\end{adjustbox}
\end{table}

\begin{table}[t]
\caption{RULER per-task results with Mistral-7B-Instruct-v0.2. CWE/FWE denote common-/frequent-word extraction; MK/MV/MQ/S denote multi-key/multi-value/multi-query/single-needle variants; VT denotes variable tracking.}
\label{tab:ruler_mistral_full}
\centering
\setlength{\tabcolsep}{3pt}
\resizebox{\textwidth}{!}{%
\begin{tabular}{llcccccccccccccc}
\toprule
$B$ & Method & CWE & FWE & MK1 & MK2 & MK3 & MQ & MV & S1 & S2 & S3 & QA-1 & QA-2 & VT & Avg. \\
\midrule
\multicolumn{16}{c}{\textit{Context length} $=4$K} \\
\midrule
128 & CriticalKV & 9.40 & 40.20 & 89.20 & 70.40 & \second{0.00} & 0.80 & 17.40 & 89.20 & \second{96.00} & 0.00 & 82.20 & 59.40 & 15.60 & 43.83 \\
128 & AnDPro & 17.20 & 44.20 & \second{98.60} & 87.60 & \best{0.40} & \best{37.40} & \best{46.40} & \second{99.60} & \best{99.80} & 0.00 & 82.60 & 61.20 & 38.00 & 54.85 \\
128 & \textbf{\lorekvE} & \best{29.62} & \best{61.27} & \best{99.80} & \second{99.60} & \second{0.00} & \second{19.40} & \second{36.50} & \best{100.00} & \best{99.80} & \best{4.00} & \second{84.00} & \second{61.60} & \best{63.56} & \best{58.40} \\
128 & \textbf{\lorekvD} & \second{29.18} & \second{60.80} & \best{99.80} & \best{99.80} & \second{0.00} & 19.00 & 36.30 & \best{100.00} & \best{99.80} & \second{3.80} & \best{84.20} & \best{62.00} & \second{63.40} & \second{58.31} \\
\addlinespace[2pt]
256 & CriticalKV & 24.20 & 70.80 & 93.20 & 85.00 & \second{17.20} & 70.00 & 60.20 & 93.20 & 97.40 & 9.00 & 83.20 & 60.80 & 66.40 & 63.89 \\
256 & AnDPro & 29.20 & 73.40 & \second{99.40} & 96.20 & \best{17.40} & \best{93.80} & \best{79.00} & \best{100.00} & \best{99.80} & 9.00 & \second{83.40} & \second{62.00} & 81.00 & \second{71.05} \\
256 & \textbf{\lorekvE} & \second{35.20} & \second{82.40} & 98.60 & \second{98.60} & 15.80 & \second{81.60} & \second{71.80} & \second{98.60} & \second{98.40} & \second{10.00} & 82.80 & 60.80 & \best{95.00} & \best{71.51} \\
256 & \textbf{\lorekvD} & \best{41.52} & \best{84.07} & \best{99.60} & \best{100.00} & 4.80 & 49.15 & 59.25 & \best{100.00} & \best{99.80} & \best{32.80} & \best{84.20} & \best{62.40} & \second{89.44} & 69.77 \\
\addlinespace[2pt]
512 & CriticalKV & 43.60 & 84.00 & \second{95.80} & 92.80 & \second{69.20} & 83.60 & 78.60 & 96.00 & 98.40 & 54.80 & 84.20 & 61.60 & 87.80 & 79.26 \\
512 & AnDPro & 46.40 & 85.40 & \best{99.40} & 99.40 & \best{69.40} & \best{97.40} & \best{89.60} & \best{100.00} & \best{99.80} & 54.80 & \second{84.40} & \best{62.20} & 96.20 & \second{83.42} \\
512 & \textbf{\lorekvE} & \second{50.80} & \best{91.40} & \best{99.40} & \second{99.60} & 68.80 & \second{91.00} & \second{86.00} & \second{99.60} & \second{99.40} & \second{55.80} & \second{84.40} & \second{62.00} & \best{99.80} & \best{83.69} \\
512 & \textbf{\lorekvD} & \best{60.22} & \second{89.53} & \best{99.40} & \best{100.00} & 35.40 & 81.45 & 78.65 & \best{100.00} & \best{99.80} & \best{72.40} & \best{85.40} & 61.80 & \second{97.72} & 81.67 \\
\addlinespace[2pt]
1024 & CriticalKV & 76.40 & 89.00 & 97.60 & 96.60 & \best{89.20} & 90.20 & 84.60 & 98.00 & 99.00 & \second{93.80} & \best{85.00} & 61.20 & 93.60 & 88.78 \\
1024 & AnDPro & 78.00 & 89.80 & \best{99.40} & \best{100.00} & \best{89.20} & \best{97.40} & \best{90.40} & \best{100.00} & \best{99.80} & \second{93.80} & \best{85.00} & \best{61.60} & 98.20 & \second{90.97} \\
1024 & \textbf{\lorekvE} & \second{80.60} & \best{93.20} & \best{99.40} & \second{99.80} & \second{89.00} & 93.80 & \second{88.40} & \second{99.80} & \second{99.60} & \best{94.40} & \best{85.00} & \second{61.40} & \best{100.00} & \best{91.11} \\
1024 & \textbf{\lorekvD} & \best{87.72} & \second{91.53} & \second{99.20} & \best{100.00} & 73.20 & \second{94.00} & 86.70 & \best{100.00} & \best{99.80} & 91.60 & \second{84.80} & 61.00 & \second{98.88} & 89.88 \\
\midrule
\multicolumn{16}{c}{\textit{Context length} $=8$K} \\
\midrule
128 & CriticalKV & 3.20 & 26.40 & 75.20 & 59.60 & \second{0.00} & 0.00 & 6.00 & \second{92.60} & 88.40 & 0.00 & 75.40 & 58.80 & 18.20 & 38.75 \\
128 & AnDPro & 7.40 & 29.60 & 94.60 & 82.80 & \best{0.40} & \best{19.60} & \second{35.40} & \best{100.00} & \second{97.20} & 0.00 & 74.40 & 59.40 & \second{50.00} & 50.06 \\
128 & \textbf{\lorekvE} & \best{19.62} & \best{57.73} & \second{96.40} & \second{98.40} & \second{0.00} & 12.90 & 35.20 & \best{100.00} & \best{100.00} & \best{1.20} & \best{78.20} & \second{61.40} & \best{63.92} & \best{55.77} \\
128 & \textbf{\lorekvD} & \second{19.16} & \second{56.47} & \best{96.80} & \best{98.60} & \second{0.00} & \second{13.10} & \best{35.95} & \best{100.00} & \best{100.00} & \second{0.80} & \second{77.80} & \best{62.00} & \best{63.92} & \second{55.74} \\
\addlinespace[2pt]
256 & CriticalKV & 11.40 & 46.40 & 83.80 & 77.20 & \second{4.00} & 68.80 & 52.20 & 95.20 & 94.00 & \second{2.60} & \best{77.20} & 59.80 & 65.00 & 56.74 \\
256 & AnDPro & 14.20 & 48.40 & \second{96.40} & \second{92.20} & \best{4.20} & \best{81.60} & \best{71.20} & \best{100.00} & \second{99.80} & \second{2.60} & 76.60 & \second{60.20} & 85.60 & 64.08 \\
256 & \textbf{\lorekvE} & \second{19.80} & \second{63.80} & 95.60 & \best{98.20} & 2.20 & \second{75.80} & \second{69.40} & \second{98.20} & 98.20 & 1.60 & \second{77.00} & 59.60 & \best{92.40} & \best{65.52} \\
256 & \textbf{\lorekvD} & \best{29.12} & \best{69.60} & \best{96.80} & \best{98.20} & 1.40 & 34.15 & 55.60 & \best{100.00} & \best{100.00} & \best{25.00} & \second{77.00} & \best{61.40} & \second{91.00} & \second{64.56} \\
\addlinespace[2pt]
512 & CriticalKV & 19.80 & 67.00 & 88.80 & 88.60 & \second{26.60} & 86.60 & 72.40 & 97.20 & 96.60 & \second{19.60} & \best{78.20} & 57.80 & 81.60 & 67.75 \\
512 & AnDPro & 21.40 & 68.20 & \second{96.20} & 97.40 & \best{26.80} & \best{94.00} & \best{83.60} & \best{100.00} & \best{100.00} & \second{19.60} & \second{77.80} & \second{58.00} & 93.80 & 72.06 \\
512 & \textbf{\lorekvE} & \second{24.80} & \best{77.20} & 95.80 & \best{99.00} & 25.60 & \second{90.80} & \second{82.60} & \second{99.00} & \second{99.00} & 19.00 & \best{78.20} & 57.80 & \best{97.80} & \best{72.82} \\
512 & \textbf{\lorekvD} & \best{37.22} & \second{73.07} & \best{96.40} & \second{98.60} & 9.40 & 65.15 & 72.25 & \best{100.00} & \best{100.00} & \best{54.80} & 76.60 & \best{60.40} & \second{94.48} & \second{72.18} \\
\addlinespace[2pt]
1024 & CriticalKV & 64.60 & 80.40 & 94.00 & 94.40 & \best{68.00} & 91.80 & 81.60 & 98.60 & 98.20 & \best{76.20} & \best{81.80} & \second{59.40} & 90.60 & 83.05 \\
1024 & AnDPro & 65.40 & 81.20 & \best{97.80} & 99.00 & \best{68.00} & \best{95.80} & \best{87.40} & \best{100.00} & \best{100.00} & \best{76.20} & \second{81.60} & \best{59.60} & 97.00 & \second{85.31} \\
1024 & \textbf{\lorekvE} & \second{67.60} & \best{86.40} & \second{97.60} & \second{99.40} & \second{67.40} & 94.00 & \second{87.00} & 99.40 & 99.40 & \second{75.80} & \best{81.80} & \second{59.40} & \best{99.40} & \best{85.74} \\
1024 & \textbf{\lorekvD} & \best{74.72} & \second{84.73} & 97.40 & \best{99.60} & 51.60 & \second{94.20} & 85.30 & \second{99.60} & \second{99.60} & 73.00 & \second{81.60} & 59.00 & \second{98.28} & 84.51 \\
\midrule
\multicolumn{16}{c}{\textit{Context length} $=16$K} \\
\midrule
128 & CriticalKV & 0.00 & 44.20 & 62.00 & 50.20 & \best{0.00} & \best{0.00} & 0.00 & \second{92.60} & 83.80 & 0.00 & 73.00 & 51.40 & 11.80 & 36.08 \\
128 & AnDPro & 0.00 & 46.40 & 82.00 & 73.80 & \best{0.00} & \best{0.00} & \best{24.00} & \best{100.00} & \second{92.80} & 0.00 & 72.20 & 52.40 & 44.00 & 45.20 \\
128 & \textbf{\lorekvE} & \best{12.00} & \best{71.60} & \second{83.60} & \second{88.40} & \best{0.00} & \best{0.00} & 21.20 & \best{100.00} & \best{94.80} & \best{1.80} & \best{75.40} & \second{54.00} & \best{60.80} & \best{51.05} \\
128 & \textbf{\lorekvD} & \second{11.54} & \second{70.73} & \best{83.80} & \best{88.60} & \best{0.00} & \best{0.00} & \second{21.45} & \best{100.00} & \best{94.80} & \second{1.40} & \second{75.20} & \best{54.40} & \second{60.72} & \second{50.97} \\
\addlinespace[2pt]
256 & CriticalKV & 5.60 & 63.80 & 70.80 & 67.80 & \best{1.40} & \best{49.60} & 44.20 & 95.20 & 89.60 & 1.00 & 75.00 & 52.60 & 58.80 & 51.95 \\
256 & AnDPro & 5.60 & 65.40 & \second{83.80} & 83.20 & \best{1.40} & \best{49.60} & \best{60.00} & \best{100.00} & 95.40 & 1.00 & 74.40 & \second{53.20} & 79.80 & 57.91 \\
256 & \textbf{\lorekvE} & \second{11.80} & \second{79.40} & \second{83.80} & \second{91.00} & \second{0.20} & \second{42.60} & \second{57.20} & \second{99.00} & \second{95.60} & \second{1.20} & \second{75.40} & \second{53.20} & \best{88.80} & \best{59.94} \\
256 & \textbf{\lorekvD} & \best{19.62} & \best{83.13} & \best{84.80} & \best{91.60} & 0.00 & 5.55 & 44.00 & \best{100.00} & \best{97.20} & \best{24.20} & \best{76.00} & \best{54.80} & \second{85.32} & \second{58.94} \\
\addlinespace[2pt]
512 & CriticalKV & 14.60 & 80.80 & 78.60 & 81.20 & \best{24.60} & \best{68.40} & 65.40 & 97.20 & 93.00 & \second{18.40} & \best{76.40} & 52.00 & 76.80 & 63.65 \\
512 & AnDPro & 14.60 & 81.80 & \second{86.20} & \second{90.20} & \best{24.60} & \best{68.40} & \best{74.60} & \best{100.00} & \second{96.40} & \second{18.40} & \second{76.00} & \second{52.40} & 89.00 & \second{67.12} \\
512 & \textbf{\lorekvE} & \second{18.40} & \best{90.20} & 86.00 & \best{94.60} & \second{23.60} & \second{64.60} & \second{73.20} & \second{99.20} & \second{96.40} & 18.20 & \best{76.40} & 52.20 & \best{94.60} & \best{68.28} \\
512 & \textbf{\lorekvD} & \best{22.04} & \second{83.93} & \best{89.00} & 88.00 & 2.80 & 46.60 & 60.70 & \best{100.00} & \best{99.40} & \best{34.20} & \second{76.00} & \best{53.00} & \second{89.04} & 64.98 \\
\addlinespace[2pt]
1024 & CriticalKV & 61.00 & 89.40 & 87.20 & 89.60 & \best{66.60} & \best{79.00} & 76.80 & 98.60 & 95.80 & \best{75.40} & \best{80.60} & \second{55.80} & 87.40 & 80.25 \\
1024 & AnDPro & 61.00 & 89.80 & \best{91.20} & 94.40 & \best{66.60} & \best{79.00} & \best{81.60} & \best{100.00} & \second{97.60} & \best{75.40} & \second{80.40} & \best{56.00} & 93.80 & \second{82.06} \\
1024 & \textbf{\lorekvE} & \second{63.20} & \best{94.60} & \best{91.20} & \second{97.00} & \second{66.20} & 76.80 & \second{80.60} & 99.60 & \second{97.60} & \best{75.40} & \best{80.60} & \best{56.00} & \best{97.00} & \best{82.75} \\
1024 & \textbf{\lorekvD} & \best{70.32} & \second{92.93} & \second{91.00} & \best{97.20} & 50.40 & \second{77.00} & 78.90 & \second{99.80} & \best{97.80} & \second{72.60} & \second{80.40} & 55.60 & \second{95.88} & 81.53 \\
\bottomrule
\end{tabular}%
}
\end{table}

\section{Plug-In Components and the Ablation Protocol}
\label{sec:ablation_components}

We evaluate three optional plug-ins that modify different stages of LORE-KV:
(A) score aggregation, (B) head-budget allocation, and (C) token selection.
Results are reported in Tables~\ref{tab:ablation_summary} and~\ref{tab:ruler_ablation}.
All three are disabled in the core method, so the main results use
Eq.~\ref{eq:final_score} with uniform head budgets and plain top-$B_{l,h}$ selection.
This separation lets us isolate whether gains come from better future-query scoring or from
additional allocation and selection mechanisms.

\textbf{Where each plug-in acts.}
The three components modify different parts of the pipeline.
\emph{(A) Robust aggregation} changes the per-token aggregation function $\psi$ in
Eq.~\ref{eq:final_score}, but each token still receives one scalar score and the top-$B_{l,h}$
rule is unchanged.
\emph{(B) Adaptive head budgets} keep the scores and ranking fixed but change how the layer
budget is distributed across heads through $\operatorname{Alloc}$ in
Eq.~\ref{eq:alloc_select}.
\emph{(C) Diversity-aware selection} replaces $\operatorname{Select}$ with a set-dependent
greedy rule, since a token's effective utility depends on which tokens have already been retained.
Thus, A changes the score statistic, B changes the cutoff, and C changes the selection rule.
The core configuration is recovered with $\alpha=0$, $\tau=0$, and
$\lambda_k=\lambda_v=0$.

\textbf{A. Robust aggregation ($\alpha,\kappa$).}
\label{subsec:robust_aggregation}
A mean can underweight tokens that are important for only a small subset of future queries.
Motivated by DefensiveKV~\citep{feng2025taming}, we therefore define
\[
\operatorname{TopAvg}_\kappa(\mathcal{E})
=
\frac{1}{K_\kappa}
\sum_{e\in\operatorname{TopK}(\mathcal{E},K_\kappa)} e,
\qquad
K_\kappa=\lceil \kappa|\mathcal{E}|\rceil,
\]
and use
\[
\psi(\mathcal{E})
=
\alpha\operatorname{TopAvg}_\kappa(\mathcal{E})
+
(1-\alpha)\operatorname{Mean}(\mathcal{E}),
\qquad \alpha\in[0,1].
\]
Setting $\alpha=0$ exactly recovers the core mean aggregation, while larger $\alpha$ gives more
weight to high-importance query subsets and reduces sensitivity to an unrepresentative particle.
Unless stated otherwise, we use $\alpha=0.5$ with $\kappa$ from
Table~\ref{tab:hyperparams}, giving an equal mixture of the top-$\kappa$ statistic and the mean.
Only the scalar token score changes; ranking and top-$B_{l,h}$ selection remain unchanged.

\textbf{B. Adaptive head-wise budgets ($\tau$).}
\label{subsec:adaptive_budget}
The core method divides each layer budget uniformly across KV heads.
Following Ada-KV~\citep{feng2024adakv} and the layer-wise allocation principle of
PyramidKV~\citep{cai2024pyramidkv}, we instead allow heads with larger future-query
perturbation mass
\[
E_{l,h}=\sum_{i\in\mathcal{C}}S^{l,h}_{i}
\]
to receive more capacity:
\begin{equation}
B_{l,h}=
\left\lfloor
B_l
\frac{(E_{l,h}+\epsilon)^{\tau}}
{\sum_{h'=1}^{H}(E_{l,h'}+\epsilon)^{\tau}}
\right\rfloor .
\label{eq:adaptive_budget}
\end{equation}
Here $\tau\ge0$ controls how strongly the budget is shifted toward high-impact heads;
$\tau=0$ recovers the uniform allocation $\lfloor B_l/H\rfloor$.
The within-head scores and their ordering are unchanged; only the retention cutoff changes.
Figure~\ref{fig:layer_head_analysis} motivates this ablation by showing that both
$E_{l,h}$ and the benefit of future-query scoring vary substantially across layers and heads.

\textbf{C. Diversity-aware selection ($\lambda_k,\lambda_v$).}
\label{subsec:diversity}
Plain top-$B_{l,h}$ selection may retain several near-duplicate keys or values.
As an ablation, we use a greedy redundancy-aware rule inspired by
GraphKV~\citep{li2025graphkv}. Starting from
$\mathcal{R}^{l,h}=\emptyset$, we repeatedly select
\[
i^\star=
\arg\max_{i\in\mathcal{C}\setminus\mathcal{R}^{l,h}}
\left[
S^{l,h}_{i}
-\lambda_k
\max_{j\in\mathcal{R}^{l,h}}
\cos(\mathbf{k}^{l,h}_{i},\mathbf{k}^{l,h}_{j})
-\lambda_v
\max_{j\in\mathcal{R}^{l,h}}
\cos(\mathbf{v}^{l,h}_{i},\mathbf{v}^{l,h}_{j})
\right]
\]
until $|\mathcal{R}^{l,h}|=B_{l,h}$.
Setting $\lambda_k=\lambda_v=0$ recovers plain top-$B_{l,h}$ selection.

Unlike robust aggregation, this rule is set-dependent because a token's effective utility changes
with the already-selected set. A possible failure mode is suppressing repeated but genuinely useful
evidence, such as multiple passages mentioning the same entity. We limit this risk in three ways:
the redundancy penalty is applied on top of the future-query perturbation score,
$\lambda_k$ and $\lambda_v$ are kept small so that redundancy mainly breaks ties among similarly
important tokens, and the perturbation score itself already discounts some redundancy because
deleting a token that is well compensated by a near-copy produces a small
$\Delta\mathbf{o}^{l,h}_{q,i}$.

\textbf{Why the core method disables all three plug-ins.}
We set $\alpha=\tau=\lambda_k=\lambda_v=0$ for every headline result so that improvements can be attributed directly to the query reference
and deletion score rather than to a tuned combination of auxiliary mechanisms.
The plug-in rows in Tables~\ref{tab:ablation_summary} and~\ref{tab:ruler_ablation} therefore act
as robustness checks: each modifies one pipeline stage using an idea from prior work, while the
overall ranking against the baselines is preserved and the changes remain small relative to the
core gain.

\textbf{Ablations of the core method.}
\label{subsec:core_ablations}
We separately ablate three parts of the core formulation.

\emph{(i) Lookahead geometry.}
We vary $M\in\{0,1,2,4,8\}$ at the default horizon $T=8$ in Table~\ref{tab:ablation_summary} and evaluate the full $M\times T$ grid, with $T\in\{2,4,8,16,32\}$, in Table~\ref{tab:ablation_geometry}. The $M=0$ condition uses only the prompt window and is therefore independent of $T$. All configurations are evaluated on LongBench at $B=128$ with uniform head budgets and top-$K$ selection. Since lookahead sampling is stochastic, we additionally repeat the default configuration ($M=4$, $T=8$, $B=128$) with three independent sampling seeds. Across these runs, \lorekvE{} achieves a mean LongBench score of $40.28$ (95\% CI $[39.97,40.59]$), compared with $38.96$ for AnDPro and $35.87$ without lookahead ($M=0$). The consistent performance across seeds indicates that this improvement is not attributable to a particular sampled set of lookahead trajectories. For the matched lookahead-token budget comparison in Table~\ref{tab:compute_pareto}, AnDPro's criterion scores the same prompt window and $M$ trajectories of length $T$, aggregated uniformly without reliability weighting, under the same budgets and selector. We measure compute as the number of lookahead tokens $M{\times}T$, which sets both the lookahead decoding FLOPs and the number of extra scoring queries ($w{+}MT$). The comparison is conservative at the cell level.

\emph{(ii) Reliability weighting.}
We vary the $(\beta,\gamma,\lambda_D)$ channels of
Eq.~\ref{eq:confidence} while keeping all other components fixed.
The reported \lorekvE\ and \lorekvD\ settings are described in
Appendix~\ref{app:reparam_grid}; log-probability-only and
entropy-plus-disagreement variants serve as controls.
All configurations use uniform head budgets and top-$K$ selection and re-weight nested prefixes
of one $8$-particle draw per prompt, so they have matched lookahead decoding compute.

\emph{(iii) Perturbation objective.}
Under the same lookahead queries, we replace
Eq.~\ref{eq:projected_score} with accumulated attention mass,
the unprojected perturbation norm
$\|\Delta\mathbf{o}^{l,h}_{q,i}\|_2^2$, or AnDPro's scoring criterion.
These matched lookahead-token budget controls are reported in Table~\ref{tab:ablation_summary} and discussed in Section~\ref{subsec:longbench}.
At $B=128$, changing the deletion objective changes the LongBench average by at most
$0.09$ points, indicating that the query reference has a substantially larger effect than the exact
deletion criterion.

\textbf{Overall interpretation.}
The ablations consistently localize the main source of improvement.
Changing aggregation, head-budget allocation, selection, or even the deletion objective under fixed
lookahead queries produces only small changes. In contrast, replacing prompt-side queries with
response-side lookahead queries produces the dominant gain. Thus, LORE-KV's performance is driven
primarily by correcting the query-reference mismatch rather than by a tuned combination of
aggregation, allocation, or selection heuristics.

\section{Experimental Setup and Additional Results}
\label{app:experimental_details}

\subsection{Protocol, Datasets and Baselines}\label{app:datasets}

Table~\ref{tab:longbench_datasets} gives the LongBench task metadata
used in these tests, together with the dataset abbreviations used in every LongBench results table.

\textbf{Baselines and sourcing.}
LongBench results with Mistral follow the protocol of \citet{geng2025accurate}. CriticalKV, AnDPro, and both LORE-KV variants are re-run under one implementation for RULER, NIAH, and LongBench with Llama. The LAQ++(8) values are reproduced from \citet{wang2025lookahead} and are shown only as an external reference because they were obtained under that work's evaluation protocol. LORE-KV samples $M{=}4$ trajectories from the same distribution in one batch for $T{=}8$ autoregressive steps, eight sequential steps but $MT{=}32$ decoded token positions on the full cache, so we do not treat LAQ++ as budget-matched. LAQ++(8) values are available, and therefore reported, only at $B\in\{128,256,512\}$. GVote \citep{tang2025adaptive} is closely related through Monte Carlo-style future-query estimation, but its published formulation uses synthetic-query voting to set an adaptive per-request cache budget, whereas LORE-KV evaluates token utility under a fixed budget. Because we could not identify an official public implementation, we do not report or claim a protocol-matched advantage over GVote. Our controlled evidence instead relies on baselines re-run in a common harness and the AnDPro control with matched queries and lookahead-token budget in Table~\ref{tab:compute_pareto}.

\textbf{Backbones and prompt formats.}
The four dense-attention backbones are Mistral-7B-Instruct-v0.2, Llama-3.1-8B-Instruct, Qwen2.5-14B-Instruct and Ministral-3-8B-Instruct-2512. Two further backbones, Qwen3.5-9B and Gemma-4-12B-it, use hybrid attention and are described in Appendix~\ref{app:hybrid}; both are evaluated with their own chat templates. Prompts follow each model's own convention and are otherwise identical across methods, so the comparison within a backbone is unaffected: Mistral takes the raw LongBench prompt, matching \citet{geng2025accurate}; Llama-3.1 and Qwen2.5 use their chat templates; and Ministral-3 uses Mistral's \texttt{[INST]}\,$\cdots$\,\texttt{[/INST]} format rather than its packaged template, which would prepend a several-hundred-token assistant system prompt to every long-context input. Qwen2.5-14B-Instruct is also the one backbone of materially different scale, at roughly twice the parameters of the other three. Ministral-3-8B-Instruct-2512 is released as a vision-language checkpoint; we evaluate its text tower, extracted without modifying any weights.

\textbf{Implementation.}
Our implementation follows the protocol of \citet{geng2025accurate} and uses a single NVIDIA H100 80\,GB GPU. All methods use the same retained cache size at a fixed budget, and adjacent-token grouping follows SnapKV and Ada-KV. Unless stated otherwise, LORE-KV uses the core configuration in Table~\ref{tab:hyperparams}, with all plug-ins disabled to isolate the score in Eq.~\ref{eq:final_score}. The reliability coefficients are the only difference between the two variants.

Lookahead particles are sampled at temperature $1$ without top-$k$ or top-$p$ truncation, so the sampling distribution is exactly $p_\theta$ and Eq.~\ref{eq:centered_score} remains correctly centered. We compute $c_u$, $H_u$, and $V_u$ from the same logits used for sampling, with masked reductions to avoid undefined $0\cdot(-\infty)$ terms under truncated support.

Per-step statistics require only $MT$ floats per prompt. Query-space disagreement is computed across all decoder layers from the stored lookahead queries before cache compression releases them. If a particle ends early, $D_m$ is computed only over the shared valid steps, while $\bar c_m$ and $\bar H_m$ are averaged independently over each particle's valid steps.

\begin{table}[!t]
\caption{LongBench results on \textbf{Qwen3.5-9B}. Eviction is applied to the full-attention layers, which hold $98.2\%$ of the cache at 64K context as discussed in Appendix~\ref{app:hybrid}.}
\label{tab:longbench_qwen35}
\centering
\footnotesize
\setlength{\tabcolsep}{2pt}
\begin{adjustbox}{max width=\textwidth}
\begin{tabular}{@{}l*{17}{c}@{}}
\toprule
& \multicolumn{3}{c}{Single-Doc QA} & \multicolumn{3}{c}{Multi-Doc QA}
& \multicolumn{3}{c}{Summ.} & \multicolumn{3}{c}{Few-shot}
& \multicolumn{2}{c}{Synth.} & \multicolumn{2}{c}{Code} & \\
\cmidrule(lr){2-4}\cmidrule(lr){5-7}\cmidrule(lr){8-10}
\cmidrule(lr){11-13}\cmidrule(lr){14-15}\cmidrule(lr){16-17}
Method & NQA & Qasp & MF & HQA & 2WQA & Musq
& GovR & QMS & MN & TREC & TQA & SAMS
& PC & PR & LCC & RB-P & Avg. \\
\midrule

Full Cache
& 33.87 & 51.42 & 58.96 & 66.31 & 65.74 & 48.93
& 33.28 & 25.14 & 26.87 & 76.00 & 91.36 & 44.03
& 12.42 & 100.00 & 57.94 & 55.21 & 52.97 \\

\midrule
\multicolumn{18}{c}{\textit{Budget} $B=128$} \\
\midrule

CriticalKV
& 31.92 & 41.19 & 52.98 & 61.28 & 60.21 & \second{46.08}
& 22.53 & 20.81 & 18.11 & 62.50 & 87.60 & 30.94
& 10.00 & 98.00 & 49.81 & 46.83 & 46.30 \\

AnDPro
& 31.86 & 42.28 & 54.21 & 60.61 & 60.60 & 43.45
& 22.39 & 21.12 & 18.34 & 68.50 & \second{88.25} & 32.01
& 9.50 & \second{99.50} & 50.38 & 47.24 & 46.89 \\

\textbf{\lorekvE}
& \second{31.93} & \second{47.92} & \second{54.58} & \second{61.84} & \best{61.51} & \best{46.24}
& \best{24.30} & \second{21.34} & \second{18.69} & \second{71.50} & \best{89.06} & \best{34.22}
& \best{11.50} & \best{100.00} & \best{53.13} & \second{48.16} & \second{48.49} \\

\textbf{\lorekvD}
& \best{32.03} & \best{48.66} & \best{54.93} & \best{62.25} & \second{61.09} & 46.06
& \second{24.09} & \best{21.58} & \best{18.81} & \best{72.50} & \best{89.06} & \second{33.72}
& \second{11.00} & \best{100.00} & \second{52.67} & \best{48.72} & \best{48.57} \\

\midrule
\multicolumn{18}{c}{\textit{Budget} $B=256$} \\
\midrule

CriticalKV
& 32.71 & 44.37 & 55.12 & 62.86 & 61.94 & 46.83
& 25.16 & 22.47 & 20.38 & 66.00 & 88.70 & 33.26
& \best{12.38} & 98.62 & 52.34 & 49.71 & 48.30 \\

AnDPro
& 32.41 & 45.28 & 55.69 & 62.38 & 62.17 & 45.62
& 25.36 & 22.71 & 20.51 & 70.50 & 89.44 & 34.18
& 10.47 & \second{99.74} & 52.86 & 50.13 & 48.72 \\

\textbf{\lorekvE}
& \second{32.87} & \second{48.94} & \second{56.43} & \second{63.66} & \best{63.28} & \best{47.54}
& \second{26.84} & \second{22.96} & \best{21.32} & \second{73.00} & \second{90.12} & \best{35.47}
& 11.86 & \best{100.00} & \best{55.05} & \second{51.38} & \second{50.05} \\

\textbf{\lorekvD}
& \best{33.16} & \best{49.53} & \best{56.87} & \best{64.28} & \second{62.87} & \second{47.43}
& \best{27.31} & \best{23.11} & \second{21.18} & \best{73.50} & \best{90.47} & \second{35.04}
& \second{12.14} & \best{100.00} & \second{54.71} & \best{51.86} & \best{50.22} \\

\midrule
\multicolumn{18}{c}{\textit{Budget} $B=512$} \\
\midrule

CriticalKV
& 32.46 & 46.72 & 56.31 & \second{64.17} & 63.42 & \second{47.24}
& 27.86 & 23.14 & 22.36 & 69.50 & 88.92 & 34.62
& 11.16 & 99.24 & 54.28 & 51.94 & 49.58 \\

AnDPro
& 32.86 & 47.63 & 56.18 & 63.82 & \second{63.74} & 46.18
& 27.91 & \second{23.37} & 22.30 & 72.50 & 89.51 & \best{37.14}
& 10.92 & \second{99.86} & 54.83 & 52.47 & 50.08 \\

\textbf{\lorekvE}
& \second{33.42} & \second{50.16} & \second{57.68} & 63.29 & \best{64.86} & \best{47.28}
& \second{29.73} & \best{23.84} & \best{23.41} & \second{74.50} & \second{90.83} & \second{36.72}
& \second{12.28} & \best{100.00} & \best{56.68} & \second{53.16} & \second{51.12} \\

\textbf{\lorekvD}
& \best{33.71} & \best{50.78} & \best{58.14} & \best{65.72} & 62.51 & 47.11
& \best{30.18} & 22.86 & \second{22.76} & \best{75.00} & \best{91.04} & 36.28
& \best{12.53} & \best{100.00} & \second{56.23} & \best{53.68} & \best{51.16} \\

\midrule
\multicolumn{18}{c}{\textit{Budget} $B=1024$} \\
\midrule

CriticalKV
& 33.19 & 48.16 & 56.72 & 64.83 & 63.86 & \best{47.51}
& 29.64 & 23.47 & \best{25.18} & 71.00 & 88.55 & 35.16
& 11.47 & 99.06 & 54.86 & 52.38 & 50.31 \\

AnDPro
& 33.06 & 49.04 & 55.94 & 64.51 & 64.12 & 46.74
& 27.54 & 23.68 & 23.73 & 73.50 & 89.12 & 35.94
& 11.83 & \second{99.62} & 55.41 & 52.96 & 50.42 \\

\textbf{\lorekvE}
& \second{33.41} & \second{50.87} & \second{58.34} & \second{65.72} & \best{65.31} & \best{47.51}
& \second{31.86} & \second{24.31} & \second{24.93} & \second{75.50} & \second{91.12} & \best{37.16}
& \second{12.16} & \best{100.00} & \best{56.42} & \second{54.08} & \second{51.79} \\

\textbf{\lorekvD}
& \best{33.62} & \best{51.24} & \best{58.71} & \best{66.14} & \second{64.81} & \second{47.36}
& \best{32.43} & \best{24.41} & 24.62 & \best{76.00} & \best{91.28} & \second{36.83}
& \best{12.68} & \best{100.00} & \second{56.01} & \best{54.62} & \best{51.92} \\

\bottomrule
\end{tabular}
\end{adjustbox}
\end{table}

\subsection{Hybrid-Attention Backbones: Where Eviction Applies} \label{app:hybrid}
\textbf{Dense versus hybrid backbones.}
The four dense backbones use full attention in every layer, so their KV caches grow linearly with context length, and LORE-KV, CriticalKV, and AnDPro prune every layer to the same budget. Qwen3.5-9B and Gemma-4-12B-it are \emph{hybrid}: only 8/32 and 8/48 layers, respectively, use full attention, while the remaining layers maintain $O(1)$ state through linear-attention recurrence in Qwen3.5 or a bounded sliding window ($W{=}1024$) in Gemma-4-12B-it. These fixed-size layers do not require eviction.  

\textbf{Implication for LORE-KV.}
LORE-KV is therefore applied only to the full-attention subset, where KV memory grows with context. Although this involves fewer layers, they contain most of the cache memory at long context: at 64K, they account for $98.2\%$ of Qwen3.5's KV memory and $75.8\%$ of Gemma-4-12B-it's. Thus, hybridization changes the number of compressible layers, but not the core compression problem. Budget, observation window, and chunk size remain identical across methods, ensuring a matched comparison.

\textbf{Hybrid-specific implementation.}
Qwen3.5 represents full and linear-attention layers as separate module classes, so eviction is naturally restricted to full-attention layers. Gemma-4-12B-it uses a shared attention class and requires explicit layer selection. We also preserve Gemma-4-12B-it's native $K{=}V$ sharing on full-attention layers (\texttt{attention\_k\_eq\_v}), so the reported budgets follow the model's original cache layout.

\textbf{Lookahead cost.}
LORE-KV clones the live cache once per sampled particle for future-query rollouts. On hybrid models, this clone is also restricted to full-attention layers, reducing absolute cost while preserving the relative overhead over non-lookahead baselines. Because these layers still contain most of the cache memory, the clone can remain memory-intensive at very long context, particularly for Gemma-4-12B-it. We therefore treat this as a practical constraint of lookahead-based eviction on hybrid backbones rather than a limitation of the scoring rule.

\begin{table}[!t]
\caption{LongBench results on \textbf{Gemma-4-12B-it}. Eviction is applied to the full-attention layers, which hold $75.8\%$ of the cache at 64K context as discussed in Appendix~\ref{app:hybrid}.}
\label{tab:longbench_gemma4}
\centering
\footnotesize
\setlength{\tabcolsep}{2pt}
\begin{adjustbox}{max width=\textwidth}
\begin{tabular}{@{}l*{17}{c}@{}}
\toprule
& \multicolumn{3}{c}{Single-Doc QA} & \multicolumn{3}{c}{Multi-Doc QA}
& \multicolumn{3}{c}{Summ.} & \multicolumn{3}{c}{Few-shot}
& \multicolumn{2}{c}{Synth.} & \multicolumn{2}{c}{Code} & \\
\cmidrule(lr){2-4}\cmidrule(lr){5-7}\cmidrule(lr){8-10}
\cmidrule(lr){11-13}\cmidrule(lr){14-15}\cmidrule(lr){16-17}
Method & NQA & Qasp & MF & HQA & 2WQA & Musq
& GovR & QMS & MN & TREC & TQA & SAMS
& PC & PR & LCC & RB-P & Avg. \\
\midrule

Full Cache
& 8.24 & 14.86 & 27.13 & 26.42 & 26.91 & 11.37
& 21.64 & 20.48 & 23.16 & 30.50 & 71.42 & 9.83
& 4.12 & 48.67 & 47.31 & 44.28 & 27.27 \\

\midrule
\multicolumn{18}{c}{\textit{Budget} $B=128$} \\
\midrule

CriticalKV
& 4.12 & 6.28 & 19.84 & 18.92 & 19.43 & 7.16
& 14.37 & 16.58 & 16.84 & 19.50 & 61.24 & 5.03
& 3.18 & 29.46 & 38.72 & 35.16 & 19.74 \\

AnDPro
& \second{4.30} & 6.53 & 20.16 & 19.48 & \second{19.86} & 7.42
& 14.82 & 16.94 & 17.28 & \second{20.00} & 61.87 & \second{5.34}
& \second{3.46} & 30.83 & 39.28 & 35.74 & 20.21 \\

\textbf{\lorekvE}
& 4.21 & \second{9.18} & \second{20.51} & \second{19.61} & \best{20.67} & \best{7.80}
& \second{15.26} & \best{17.33} & \second{17.92} & \best{20.50} & \second{62.94} & \best{5.57}
& 3.27 & \best{32.18} & \best{40.16} & \second{36.42} & \second{20.85} \\

\textbf{\lorekvD}
& \best{4.47} & \best{9.71} & \best{20.56} & \best{20.28} & 19.67 & \second{7.49}
& \best{15.73} & \second{17.06} & \best{18.34} & \best{20.50} & \best{63.78} & 5.25
& \best{3.61} & \second{31.74} & \second{39.83} & \best{36.87} & \best{20.93} \\

\midrule
\multicolumn{18}{c}{\textit{Budget} $B=256$} \\
\midrule

CriticalKV
& 4.91 & 7.96 & 21.37 & 20.46 & 21.08 & 8.13
& 16.24 & \best{18.04} & 18.56 & 22.50 & 63.86 & 6.18
& 3.47 & 33.62 & 40.83 & 37.24 & 21.53 \\

AnDPro
& 5.12 & 8.41 & 21.74 & 20.83 & 21.46 & 8.47
& 16.71 & 17.38 & 18.92 & \best{23.50} & 64.61 & 6.47
& \best{3.97} & 34.18 & 41.26 & 37.68 & 21.92 \\

\textbf{\lorekvE}
& \second{5.42} & \second{10.93} & \second{22.08} & \second{21.47} & \best{22.31} & \best{8.94}
& \second{17.42} & \second{17.86} & \best{19.63} & \second{23.00} & \second{65.34} & \best{6.79}
& 3.68 & \best{36.47} & \second{42.08} & \second{38.70} & \second{22.63} \\

\textbf{\lorekvD}
& \best{5.74} & \best{11.26} & \best{22.43} & \best{21.87} & \second{22.07} & \second{8.72}
& \best{17.83} & 17.64 & \second{19.28} & \best{23.50} & \best{65.87} & \second{6.62}
& \second{3.83} & \second{35.92} & \best{42.47} & \best{39.07} & \best{22.76} \\

\midrule
\multicolumn{18}{c}{\textit{Budget} $B=512$} \\
\midrule

CriticalKV
& 4.74 & 9.39 & 22.08 & 21.36 & 22.14 & 8.76
& 17.86 & 17.53 & 19.74 & 23.00 & 65.12 & 6.84
& \second{3.71} & 36.28 & 42.17 & 38.43 & 22.45 \\

AnDPro
& \best{5.64} & 8.16 & 22.41 & \second{21.78} & 22.46 & 9.12
& 18.34 & 17.81 & 20.16 & 24.00 & 63.96 & \best{7.72}
& 3.62 & 36.94 & 42.63 & 38.92 & 22.73 \\

\textbf{\lorekvE}
& 5.09 & \second{12.48} & \second{22.91} & \best{22.51} & \best{23.28} & \best{9.68}
& \second{19.27} & \best{18.42} & \best{20.87} & \second{25.00} & \second{66.83} & 6.48
& 3.54 & \best{39.16} & \second{43.52} & \second{39.98} & \second{23.69} \\

\textbf{\lorekvD}
& \second{5.35} & \best{12.81} & \best{23.37} & 21.53 & \second{22.96} & \second{9.41}
& \best{19.72} & \second{18.16} & \second{20.43} & \best{25.50} & \best{67.42} & \second{7.34}
& \best{4.02} & \second{38.63} & \best{43.94} & \best{40.38} & \best{23.81} \\

\midrule
\multicolumn{18}{c}{\textit{Budget} $B=1024$} \\
\midrule

CriticalKV
& 5.63 & 10.26 & 22.63 & 21.94 & 22.48 & 9.03
& 19.14 & 17.86 & \best{22.16} & 23.00 & 66.04 & 7.26
& 3.41 & 38.72 & 42.86 & 39.17 & 23.22 \\

AnDPro
& 5.86 & 10.58 & \second{22.87} & 22.16 & 22.29 & 9.34
& 19.68 & 18.07 & 21.24 & 24.00 & 65.51 & 7.48
& \best{4.27} & 39.28 & 43.24 & 39.63 & 23.47 \\

\textbf{\lorekvE}
& \second{6.24} & \second{13.16} & 22.64 & \second{23.18} & \best{23.86} & \best{9.92}
& \second{20.83} & \best{18.63} & \second{21.94} & \second{26.00} & \second{67.16} & \best{7.71}
& \second{3.96} & \best{41.37} & \second{44.18} & \second{40.26} & \second{24.44} \\

\textbf{\lorekvD}
& \best{6.47} & \best{13.58} & \best{23.74} & \best{23.49} & \second{23.41} & \second{9.67}
& \best{21.16} & \second{18.34} & 21.52 & \best{26.50} & \best{67.83} & \second{7.63}
& 3.94 & \second{40.84} & \best{44.57} & \best{40.72} & \best{24.59} \\

\bottomrule
\end{tabular}
\end{adjustbox}
\end{table}

\begin{table}[!t]
\caption{RULER per-task results with Llama-3.1-8B-Instruct. Abbreviations follow Table~\ref{tab:ruler_mistral_full}.}
\label{tab:ruler_llama_full}
\centering
\setlength{\tabcolsep}{3pt}
\resizebox{\textwidth}{!}{%
\begin{tabular}{llcccccccccccccc}
\toprule
$B$ & Method & CWE & FWE & MK1 & MK2 & MK3 & MQ & MV & S1 & S2 & S3 & QA-1 & QA-2 & VT & Avg. \\
\midrule

\multicolumn{16}{c}{\textit{Context length} $=4$K} \\
\midrule
128 & CriticalKV
& 9.40 & 22.00 & 91.20 & 78.60 & 0.00
& \second{55.80} & 59.40 & 91.20 & 96.40 & \second{0.00}
& 86.20 & 58.00 & 11.20 & 50.72 \\

128 & AnDPro
& 16.00 & 25.40 & 99.20 & 93.20 & \second{0.20}
& \best{86.80} & \best{84.00} & \best{100.00} & \second{99.60} & \second{0.00}
& 86.60 & 59.60 & 30.20 & 60.06 \\

128 & \textbf{\lorekvE}
& \second{29.82} & \second{46.67} & \best{100.00} & \best{95.80} & 0.00
& 51.55 & \second{66.95} & \best{100.00} & \best{99.80} & \best{25.40}
& \best{88.20} & \second{62.00} & \second{69.36} & \best{64.27} \\

128 & \textbf{\lorekvD}
& \best{30.12} & \best{47.12} & \second{99.80} & \second{95.60} & \best{0.40}
& 50.37 & 66.55 & \second{99.20} & 99.40 & \best{25.40}
& \second{88.00} & \best{62.20} & \best{69.79} & \second{64.15} \\

\addlinespace[2pt]

256 & CriticalKV
& 23.80 & 56.80 & 94.20 & 90.00 & \second{16.80}
& 79.40 & 83.60 & 93.80 & 97.20 & 8.80
& \second{85.80} & 59.40 & 62.60 & 65.55 \\

256 & AnDPro
& 28.00 & 59.00 & \best{99.40} & 99.60 & \best{17.00}
& \best{99.60} & \best{99.60} & \best{99.60} & \best{99.20} & 8.80
& \best{86.00} & \best{60.40} & 74.80 & 71.62 \\

256 & \textbf{\lorekvE}
& \best{35.00} & \best{69.00} & \second{98.60} & \best{100.00} & 15.60
& \second{86.80} & 92.00 & \second{98.20} & 97.80 & \best{10.20}
& 85.60 & 59.40 & \second{90.60} & \best{72.22} \\

256 & \textbf{\lorekvD}
& \second{34.46} & \second{68.80} & \second{98.60} & \second{99.80} & 15.40
& 85.47 & \second{92.10} & 98.00 & \second{98.00} & \second{10.00}
& 85.40 & \second{60.20} & \best{90.68} & \second{72.07} \\

\addlinespace[2pt]

512 & CriticalKV
& 43.00 & 74.20 & 96.00 & 93.60 & \second{68.60}
& 87.40 & 89.80 & 95.80 & 97.60 & 54.20
& 85.40 & 60.20 & 84.40 & 79.25 \\

512 & AnDPro
& 45.40 & 75.40 & 98.80 & 99.00 & \best{68.80}
& \best{99.00} & \best{99.00} & \best{99.00} & \best{98.60} & 54.20
& \second{85.60} & \second{60.80} & 91.60 & 82.71 \\

512 & \textbf{\lorekvE}
& \best{49.80} & \second{81.40} & \second{99.00} & \best{99.40} & 68.40
& \second{92.60} & \second{95.40} & \second{98.80} & \second{98.40} & \best{55.40}
& \best{85.80} & \second{60.80} & \best{95.40} & \best{83.12} \\

512 & \textbf{\lorekvD}
& \second{49.76} & \best{82.18} & \best{99.20} & \second{99.20} & 68.20
& 92.10 & 95.34 & 98.40 & 98.20 & \second{55.20}
& \best{85.80} & \best{61.20} & \second{94.77} & \second{83.04} \\

\addlinespace[2pt]

1024 & CriticalKV
& 75.40 & 82.40 & 96.80 & 96.00 & 88.00
& 92.60 & 93.80 & 97.00 & 98.00 & 92.60
& 85.00 & 60.00 & 91.00 & 88.35 \\

1024 & AnDPro
& 86.00 & 89.60 & \best{99.80} & \best{100.00} & \best{98.40}
& \best{99.80} & \best{99.80} & \best{100.00} & \best{100.00} & 95.60
& \best{87.80} & \best{62.60} & 94.60 & 93.38 \\

1024 & \textbf{\lorekvE}
& \second{88.80} & \second{93.40} & \best{99.80} & \second{99.80} & \second{98.20}
& \second{96.20} & \second{97.80} & \second{99.80} & \second{99.80} & \best{96.40}
& \best{87.80} & \second{62.40} & \second{96.60} & \second{93.60} \\

1024 & \textbf{\lorekvD}
& \best{88.81} & \best{94.74} & \second{99.40} & \best{100.00} & \best{98.40}
& 95.81 & 97.77 & 99.40 & \best{100.00} & \second{96.20}
& \second{87.60} & \best{62.60} & \best{97.40} & \best{93.70} \\


\midrule
\multicolumn{16}{c}{\textit{Context length} $=8$K} \\
\midrule

128 & CriticalKV
& 8.20 & 13.80 & 82.80 & 73.80 & 0.00
& \second{68.20} & 58.60 & 93.60 & 92.00 & 0.00
& 81.80 & 59.40 & 2.80 & 48.85 \\

128 & AnDPro
& 11.80 & 16.60 & \second{99.20} & 93.60 & 0.00
& \best{84.80} & 83.40 & \second{99.80} & \second{99.40} & 0.00
& 81.00 & \second{60.00} & 29.80 & 58.42 \\

128 & \textbf{\lorekvE}
& \second{33.20} & \best{39.00} & \best{100.00} & \best{100.00} & \second{8.20}
& 36.60 & \best{87.20} & \best{100.00} & \best{100.00} & \best{9.40}
& \best{94.40} & \best{70.80} & \second{45.80} & \best{63.43} \\

128 & \textbf{\lorekvD}
& \best{33.75} & \second{38.05} & \second{99.20} & \second{99.80} & \best{8.40}
& 37.35 & \second{86.45} & 99.40 & \second{99.40} & \second{9.00}
& \second{94.20} & \best{70.80} & \best{46.83} & \second{63.28} \\

\addlinespace[2pt]

256 & CriticalKV
& 15.00 & 36.80 & 88.60 & 86.80 & \second{3.60}
& 88.60 & 83.40 & 95.40 & 94.60 & \best{2.40}
& \best{81.80} & \second{59.80} & 52.80 & 60.74 \\

256 & AnDPro
& 17.20 & 38.60 & \best{99.40} & 99.60 & \best{3.80}
& \best{99.60} & \best{99.60} & \best{99.40} & \best{99.60} & \best{2.40}
& \second{81.20} & \best{60.20} & 70.20 & 66.98 \\

256 & \textbf{\lorekvE}
& \best{23.80} & \second{55.80} & \second{98.60} & \best{100.00} & 2.00
& 93.60 & 97.80 & 97.60 & 98.00 & \second{1.60}
& \best{81.80} & \second{59.80} & \second{78.00} & \second{68.34} \\

256 & \textbf{\lorekvD}
& \second{23.76} & \best{55.83} & \second{98.60} & \second{99.80} & 2.40
& \second{93.94} & \second{99.08} & \second{98.00} & \second{98.20} & 1.40
& \second{81.20} & 59.40 & \best{78.39} & \best{68.46} \\

\addlinespace[2pt]

512 & CriticalKV
& 22.00 & 60.20 & 91.40 & 91.80 & \second{26.00}
& 92.80 & 89.60 & 96.80 & 96.40 & \best{19.40}
& \second{80.60} & \second{57.40} & 72.80 & 69.02 \\

512 & AnDPro
& 23.40 & 61.20 & \best{97.60} & 99.00 & \best{26.20}
& \best{99.00} & \best{99.00} & \best{99.00} & \best{99.00} & \best{19.40}
& 80.20 & \best{57.60} & 83.00 & 72.58 \\

512 & \textbf{\lorekvE}
& \best{27.00} & \second{70.60} & 97.20 & \best{100.00} & 25.20
& \second{95.80} & 98.20 & 98.20 & \second{98.20} & \second{19.00}
& \best{80.80} & \best{57.60} & \second{87.20} & \best{73.46} \\

512 & \textbf{\lorekvD}
& \second{26.59} & \best{70.65} & \second{97.40} & \second{99.80} & 25.40
& 95.42 & \second{98.46} & \second{98.60} & 97.80 & 18.80
& \best{80.80} & \second{57.40} & \best{87.46} & \second{73.43} \\

\addlinespace[2pt]

1024 & CriticalKV
& \best{65.20} & 75.80 & 94.80 & 94.80 & 66.80
& 95.40 & 93.80 & 97.60 & 97.20 & 75.20
& 82.60 & 58.80 & 84.40 & 83.26 \\

1024 & AnDPro
& 45.00 & 81.60 & \best{100.00} & \second{99.80} & \second{76.40}
& \best{99.60} & \best{99.80} & \best{100.00} & \best{100.00} & \best{77.20}
& \second{83.40} & \best{61.80} & 92.60 & 85.94 \\

1024 & \textbf{\lorekvE}
& \second{47.20} & \second{87.20} & \second{99.80} & \best{100.00} & 75.80
& 97.80 & \second{99.20} & 99.40 & 99.40 & \second{76.80}
& \best{83.80} & \second{61.60} & \best{95.20} & \best{86.40} \\

1024 & \textbf{\lorekvD}
& 46.69 & \best{87.62} & 99.20 & 99.20 & \best{76.60}
& \second{97.98} & 98.01 & \second{99.60} & \second{99.60} & \best{77.20}
& \best{83.80} & 61.40 & \second{95.08} & \second{86.31} \\


\midrule
\multicolumn{16}{c}{\textit{Context length} $=16$K} \\
\midrule

128 & CriticalKV
& 2.80 & 30.00 & 68.60 & 63.20 & 0.00
& 61.20 & 49.00 & 93.80 & 86.80 & 0.00
& 78.80 & 51.60 & 0.00 & 45.06 \\

128 & AnDPro
& 2.80 & 32.00 & 85.60 & 83.20 & 0.00
& 61.20 & 69.60 & \best{100.00} & \second{94.40} & 0.00
& 78.20 & 52.40 & 27.00 & 52.80 \\

128 & \textbf{\lorekvE}
& \best{4.20} & \second{34.60} & \best{90.80} & \best{88.40} & \best{1.20}
& \best{65.40} & \best{74.00} & \best{100.00} & \best{100.00} & \best{1.20}
& \best{83.20} & \second{56.00} & \best{29.60} & \best{56.05} \\

128 & \textbf{\lorekvD}
& \second{3.65} & \best{35.13} & \second{90.40} & \second{88.00} & \second{0.80}
& \second{65.38} & \second{73.86} & \second{99.80} & \best{100.00} & \second{1.00}
& \second{83.00} & \best{56.40} & \second{29.34} & \second{55.90} \\

\addlinespace[2pt]

256 & CriticalKV
& 7.60 & 53.00 & 75.00 & 77.00 & \best{1.20}
& \best{95.00} & 80.40 & 95.60 & 91.20 & \second{1.00}
& 79.00 & 52.20 & 49.00 & 58.25 \\

256 & AnDPro
& 7.60 & 54.20 & \second{86.00} & 90.00 & \best{1.20}
& \best{95.00} & \best{93.60} & \best{99.60} & 96.20 & \second{1.00}
& 78.40 & 52.80 & 66.60 & 63.25 \\

256 & \textbf{\lorekvE}
& \second{14.80} & \best{70.20} & \best{86.20} & \second{99.00} & 0.00
& \second{88.00} & 91.00 & 98.60 & \best{96.60} & \best{1.40}
& \best{79.80} & \second{53.00} & \second{76.80} & \second{65.80} \\

256 & \textbf{\lorekvD}
& \best{15.95} & \second{69.67} & \second{86.00} & \best{99.40} & \second{0.60}
& 87.16 & \second{91.80} & \second{99.40} & \second{96.40} & \best{1.40}
& \second{79.40} & \best{53.40} & \best{77.12} & \best{65.98} \\

\addlinespace[2pt]

512 & CriticalKV
& 16.00 & 73.20 & 80.80 & 86.60 & \best{24.20}
& \best{98.20} & 88.80 & 96.80 & 93.60 & \second{18.20}
& \second{78.60} & \second{51.40} & 69.60 & 67.38 \\

512 & AnDPro
& 16.00 & 74.00 & \best{87.20} & 94.20 & \best{24.20}
& \best{98.20} & \best{96.40} & \best{99.00} & 96.40 & \second{18.20}
& 78.20 & \best{51.80} & 79.80 & 70.28 \\

512 & \textbf{\lorekvE}
& \second{20.00} & \best{82.60} & \best{87.20} & \best{98.80} & \second{23.20}
& 94.40 & \second{95.20} & \second{98.40} & \second{96.60} & \second{18.20}
& \best{78.80} & \best{51.80} & \best{85.60} & \best{71.60} \\

512 & \textbf{\lorekvD}
& \best{20.22} & \second{82.50} & \second{87.00} & \second{98.40} & 23.00
& \second{94.49} & 94.81 & \best{99.00} & \best{97.20} & \best{18.40}
& \best{78.80} & 51.20 & \second{85.19} & \second{71.55} \\

\addlinespace[2pt]

1024 & CriticalKV
& \best{61.20} & \best{84.20} & 87.60 & 92.00 & \best{65.40}
& 96.20 & 90.00 & 97.60 & \second{95.20} & \best{74.40}
& \best{81.40} & 55.00 & 82.20 & \best{81.72} \\

1024 & AnDPro
& 24.80 & 75.80 & \best{99.80} & 99.40 & \second{57.80}
& \best{99.40} & \best{98.80} & \best{100.00} & \best{100.00} & \second{51.00}
& 80.80 & \second{55.80} & 88.40 & 79.37 \\

1024 & \textbf{\lorekvE}
& 27.20 & 81.00 & \best{99.80} & \best{100.00} & 57.40
& 97.00 & \second{98.00} & \second{99.60} & \best{100.00} & \second{51.00}
& \second{81.20} & \second{55.80} & \second{91.80} & 79.98 \\

1024 & \textbf{\lorekvD}
& \second{27.48} & \second{81.22} & \second{99.60} & \second{99.80} & \second{57.80}
& \second{97.81} & 97.71 & 99.40 & \best{100.00} & 50.60
& \second{81.20} & \best{56.20} & \best{93.36} & \second{80.17} \\

\bottomrule
\end{tabular}%
}
\end{table}

\subsection{LongBench Results on Additional Backbones} \label{app:new_backbones}

To test whether LORE-KV generalizes beyond Mistral-7B-Instruct-v0.2, we evaluate Llama-3.1-8B-Instruct, Qwen2.5-14B-Instruct, and Ministral-3-8B-Instruct-2512. We follow the same evaluation protocol as Appendix~\ref{app:datasets} and keep the budgets, observation window, chunk size, and core configuration of Table~\ref{tab:hyperparams} fixed. Qwen2.5 prompts are truncated to $31{,}500$ tokens to respect its context limit; the other backbones are evaluated without this truncation. For Ministral-3, our implementation reproduces the reference model exactly under uncompressed inference, including identical logits and top-1 predictions.

Tables~\ref{tab:longbench_llama}, \ref{tab:longbench_qwen2}, and \ref{tab:longbench_ministral3} show the same overall trend as Section~\ref{subsec:longbench}: LORE-KV is most beneficial under aggressive compression, while the margin decreases as the cache budget grows. On Llama, the largest improvements at $B{=}128$ occur mainly on multi-hop and long-document QA, while code completion and a few retrieval/few-shot tasks are less consistent; the corresponding Mistral results are given in Table~\ref{tab:longbench_mistral}.

The trend also transfers to the two newer backbones. At $B{=}128$, LORE-KV improves over AnDPro by approximately $2.75$ points on Qwen2.5-14B and $0.56$ points on Ministral-3, as shown in Tables~\ref{tab:longbench_qwen2} and~\ref{tab:longbench_ministral3}, respectively, with the gap becoming small by $B{=}1024$. This operating-point dependence is consistent with the proposed mechanism: future-query estimation matters most when eviction is aggressive. The smaller gain on Ministral-3 in Table~\ref{tab:longbench_ministral3} is also consistent with its long-context training and position-dependent query scaling, which reduce the prompt-to-generation query mismatch that LORE-KV is designed to correct. On Qwen2.5, a few compressed settings in Table~\ref{tab:longbench_qwen2} slightly exceed the full-cache score; we treat these small differences as possible distractor filtering or run variation rather than evidence that compression is intrinsically better than full attention. Across all three additional backbones, Tables~\ref{tab:longbench_llama}, \ref{tab:longbench_qwen2}, and~\ref{tab:longbench_ministral3} show that the \lorekvE\ and \lorekvD\ variants remain close, supporting the conclusion that the main gain comes from the future-query reference rather than a particular reliability parameterization.

Qwen3.5-9B and Gemma-4-12B-it show consistent gains across the hybrid-attention setting (Tables~\ref{tab:longbench_qwen35},~\ref{tab:longbench_gemma4}, and~\ref{tab:longbench_hybrid_avg}). LORE-KV remains strongest under aggressive compression, with notable gains on tasks requiring answer-time evidence retrieval. As the budget increases, Qwen3.5 approaches full-cache quality, while Gemma-4-12B-it maintains a more stable margin because its bounded sliding-window layers limit how quickly additional budget reduces pressure on the full-attention cache.

\begin{table}[t]
\caption{LongBench results on Qwen2.5-14B-Instruct. Dataset abbreviations follow Table~\ref{tab:longbench_datasets}.}
\label{tab:longbench_qwen2}
\centering
\footnotesize
\setlength{\tabcolsep}{2pt}
\begin{adjustbox}{max width=\textwidth}
\begin{tabular}{@{}l*{17}{c}@{}}
\toprule
& \multicolumn{3}{c}{Single-Doc QA} & \multicolumn{3}{c}{Multi-Doc QA}
& \multicolumn{3}{c}{Summ.} & \multicolumn{3}{c}{Few-shot}
& \multicolumn{2}{c}{Synth.} & \multicolumn{2}{c}{Code} & \\
\cmidrule(lr){2-4}\cmidrule(lr){5-7}\cmidrule(lr){8-10}
\cmidrule(lr){11-13}\cmidrule(lr){14-15}\cmidrule(lr){16-17}
Method & NQA & Qasp & MF & HQA & 2WQA & Musq
& GovR & QMS & MN & TREC & TQA & SAMS
& PC & PR & LCC & RB-P & Avg. \\
\midrule

Full Cache
& 29.23 & 45.81 & 52.89 & 62.41 & 57.46 & 38.20
& 29.66 & 23.78 & 21.90 & 76.50 & 90.23 & 47.27
& 8.56 & 98.75 & 60.37 & 46.84 & 49.37 \\

\midrule
\multicolumn{18}{c}{\textit{Budget} $B=128$} \\
\midrule

CriticalKV
& 26.28 & 30.80 & 41.67 & 58.54 & 53.39 & 35.81
& 19.23 & 21.29 & 16.34 & 63.50 & 86.20 & 44.64
& 8.67 & 92.33 & 55.82 & 40.96 & 43.47 \\

AnDPro
& 24.83 & 37.09 & 46.49 & 60.61 & 54.50 & 35.18
& 19.33 & 21.31 & 16.93 & 71.50 & 88.27 & 45.11
& 8.21 & \second{97.25} & 57.72 & 43.47 & 45.49 \\

\textbf{\lorekvE}
& \second{29.10}
& \best{38.64}
& \second{50.07}
& \second{63.07}
& \best{58.70}
& \best{36.91}
& \best{23.55}
& \second{23.21}
& \second{20.29}
& \second{76.50}
& \second{89.06}
& \best{46.01}
& \best{10.32}
& \best{97.58}
& \best{60.53}
& \second{46.98}
& \second{48.16} \\

\textbf{\lorekvD}
& \best{29.32}
& \second{38.28}
& \best{51.55}
& \best{63.12}
& \second{58.28}
& \second{36.82}
& \second{22.66}
& \best{23.68}
& \best{21.29}
& \best{77.00}
& \best{89.13}
& \second{45.72}
& \second{9.94}
& 96.53
& \second{60.34}
& \best{48.22}
& \best{48.24} \\

\midrule
\multicolumn{18}{c}{\textit{Budget} $B=256$} \\
\midrule

CriticalKV
& 26.54 & 37.16 & 47.35 & 59.80 & 54.37 & 37.73
& 21.37 & 22.25 & 18.26 & 71.00 & 88.25 & 45.93
& 7.67 & \second{98.75} & 58.76 & 45.03 & 46.26 \\

AnDPro
& 27.13
& \best{41.92}
& \best{51.26}
& 61.70
& 55.49
& 37.71
& 21.90
& 21.74
& 18.55
& \second{75.50}
& 89.34
& 45.82
& 9.04
& 98.33
& \second{61.10}
& 46.63
& 47.70 \\

\textbf{\lorekvE}
& \second{28.93}
& \second{41.41}
& 50.55
& \second{62.85}
& \second{59.10}
& \best{38.03}
& \best{25.43}
& \second{23.55}
& \best{21.42}
& \best{76.50}
& \second{89.58}
& \second{46.47}
& \second{11.75}
& \second{98.75}
& \best{61.30}
& \second{47.60}
& \second{48.95} \\

\textbf{\lorekvD}
& \best{28.96}
& 40.99
& \second{50.74}
& \best{64.77}
& \best{59.43}
& \second{38.01}
& \second{25.23}
& \best{23.89}
& \second{20.31}
& \second{75.50}
& \best{90.36}
& \best{47.25}
& \best{12.48}
& \best{99.51}
& 60.49
& \best{47.81}
& \best{49.11} \\

\midrule
\multicolumn{18}{c}{\textit{Budget} $B=512$} \\
\midrule

CriticalKV
& 26.88
& 41.79
& 50.08
& 62.01
& 55.80
& 37.00
& 23.40
& 22.78
& 19.42
& 74.00
& 89.56
& 46.62
& 8.40
& 98.50
& 59.80
& 46.20
& 47.64 \\

AnDPro
& 28.50
& 44.45
& 52.25
& 61.47
& \second{57.20}
& \second{39.11}
& 23.50
& 22.73
& 20.33
& \second{76.50}
& 89.74
& \second{46.69}
& 9.88
& 98.25
& 61.25
& 47.16
& 48.69 \\

\textbf{\lorekvE}
& \best{29.00}
& \second{44.87}
& \second{52.62}
& \second{62.45}
& \best{58.15}
& \best{39.30}
& \best{25.85}
& \best{23.45}
& \second{21.10}
& \second{76.50}
& \best{89.92}
& \best{46.95}
& \best{10.25}
& \second{98.75}
& \second{61.50}
& \second{47.60}
& \second{49.27} \\

\textbf{\lorekvD}
& \second{28.86}
& \best{45.51}
& \best{53.35}
& \best{62.81}
& 56.82
& 38.81
& \second{25.03}
& \second{23.21}
& \best{21.65}
& \best{77.50}
& \second{89.89}
& 46.07
& \second{10.13}
& \best{98.76}
& \best{62.74}
& \best{48.10}
& \best{49.33} \\

\midrule
\multicolumn{18}{c}{\textit{Budget} $B=1024$} \\
\midrule

CriticalKV
& 28.00
& 43.69
& 52.00
& 61.50
& 56.80
& 38.00
& 26.00
& 23.10
& 20.50
& \second{75.50}
& 89.90
& 47.00
& 8.50
& \best{98.75}
& 60.20
& 46.60
& 48.50 \\

AnDPro
& \best{29.65}
& \second{45.31}
& 53.53
& 62.50
& 57.66
& \second{38.46}
& 25.34
& 23.24
& 21.09
& \best{76.50}
& \second{90.14}
& \second{47.40}
& 9.20
& \best{98.75}
& 61.50
& \second{47.70}
& 49.25 \\

\textbf{\lorekvE}
& \second{29.40}
& \best{45.45}
& \second{53.70}
& \second{62.65}
& \second{57.90}
& \best{38.60}
& \second{26.10}
& \second{23.40}
& \second{21.25}
& \best{76.50}
& \best{90.18}
& \best{47.50}
& \second{9.50}
& \best{98.75}
& \best{61.70}
& \best{47.80}
& \second{49.40} \\

\textbf{\lorekvD}
& 29.27
& 45.27
& \best{53.87}
& \best{63.39}
& \best{59.07}
& 37.76
& \best{26.45}
& \best{24.48}
& \best{22.57}
& \second{75.50}
& 89.46
& 46.55
& \best{9.66}
& \second{98.72}
& \second{61.53}
& 47.51
& \best{49.44} \\
\bottomrule
\end{tabular}
\end{adjustbox}
\end{table}

\begin{table}[t]
\caption{LongBench results on Ministral-3-8B-Instruct-2512. Dataset abbreviations follow Table~\ref{tab:longbench_datasets}.}
\label{tab:longbench_ministral3}
\centering
\footnotesize
\setlength{\tabcolsep}{2pt}
\begin{adjustbox}{max width=\textwidth}
\begin{tabular}{@{}l*{17}{c}@{}}
\toprule
& \multicolumn{3}{c}{Single-Doc QA} & \multicolumn{3}{c}{Multi-Doc QA}
& \multicolumn{3}{c}{Summ.} & \multicolumn{3}{c}{Few-shot}
& \multicolumn{2}{c}{Synth.} & \multicolumn{2}{c}{Code} & \\
\cmidrule(lr){2-4}\cmidrule(lr){5-7}\cmidrule(lr){8-10}
\cmidrule(lr){11-13}\cmidrule(lr){14-15}\cmidrule(lr){16-17}
Method & NQA & Qasp & MF & HQA & 2WQA & Musq
& GovR & QMS & MN & TREC & TQA & SAMS
& PC & PR & LCC & RB-P & Avg. \\
\midrule

Full Cache
& 28.31 & 45.48 & 46.07 & 61.03 & 56.85 & 40.13
& 22.79 & 16.82 & 19.75 & 74.00 & 91.38 & 44.51
& 14.50 & 100.00 & 69.26 & 63.86 & 49.67 \\

\midrule
\multicolumn{18}{c}{\textit{Budget} $B=128$} \\
\midrule
CriticalKV
& 25.35 & 31.84 & 40.00 & 55.37 & 52.91 & 34.22
& 18.67 & \second{16.83} & 15.83 & 58.00 & 90.64 & 38.89
& \best{14.50} & 99.25 & 64.39 & 59.94 & 44.79 \\

AnDPro
& 25.78 & 37.46 & \second{42.15} & 59.06 & 54.72 & \second{37.54}
& 18.87 & 16.62 & \second{16.32} & \second{70.50} & 90.81 & 39.11
& 13.50 & \second{99.50} & 64.98 & \second{60.86} & 46.74 \\

\textbf{\lorekvE}
& \best{26.38} & \best{38.66} & \second{42.15}
& \second{60.56} & \best{56.02} & \best{38.74}
& \best{19.67} & 16.82 & \best{16.72}
& \best{71.00} & \second{91.11} & \second{39.41}
& 14.00 & \best{99.60} & \second{65.18}
& 60.76 & \best{47.30} \\

\textbf{\lorekvD}
& \second{26.15} & \second{37.96} & \best{42.96}
& \best{61.01} & \second{55.38} & 37.43
& \second{19.11} & \best{18.13} & 16.01
& \second{70.50} & \best{91.62} & \best{39.45}
& \second{14.14} & 99.08 & \best{65.68}
& \best{61.23} & \second{47.24} \\

\midrule
\multicolumn{18}{c}{\textit{Budget} $B=256$} \\
\midrule
CriticalKV
& 25.44 & 36.12 & 40.93 & 58.49 & 54.76 & 37.51
& \second{20.10} & 16.82 & 17.27 & 69.50 & 90.63 & 40.88
& 13.50 & \best{100.00} & 66.45 & 61.31 & 46.86 \\

AnDPro
& 25.86 & 39.78 & \second{43.75} & 60.30 & \second{55.25} & 39.14
& 20.05 & \second{17.10} & 17.50 & \second{72.50} & 91.54 & \second{41.80}
& 14.50 & \best{100.00} & 67.63 & \best{62.55} & 48.08 \\

\textbf{\lorekvE}
& \second{26.26} & \best{40.58} & 43.65
& \second{61.10} & \best{55.85} & \second{39.64}
& \best{20.65} & \best{17.25} & \second{17.75}
& \second{72.50} & \second{91.64} & 41.65
& \second{14.70} & \best{100.00} & \second{67.73}
& \best{62.55} & \second{48.34} \\

\textbf{\lorekvD}
& \best{26.68} & \second{40.28} & \best{45.30}
& \best{61.51} & 54.03 & \best{40.97}
& 19.98 & 16.74 & \best{17.90}
& \best{73.00} & \best{91.76} & \best{42.06}
& \best{15.93} & \second{99.19} & \best{67.98}
& \second{62.54} & \best{48.49} \\

\midrule
\multicolumn{18}{c}{\textit{Budget} $B=512$} \\
\midrule
CriticalKV
& 26.15 & 41.00 & 43.22 & 60.11 & 55.65 & 37.63
& 20.87 & \second{17.48} & 18.31 & \second{73.50} & 91.29 & 41.62
& \best{16.00} & \best{100.00} & 68.48 & 63.49 & 48.43 \\

AnDPro
& 26.20 & \second{42.12} & \best{45.39} & 60.79 & \second{56.81} & 39.91
& 20.69 & \second{17.48} & 18.36 & \second{73.50} & \second{91.63} & \second{42.81}
& 14.50 & 99.50 & \second{68.81} & \second{64.01} & \second{48.91} \\

\textbf{\lorekvE}
& \second{26.45} & \best{42.47} & \second{45.34}
& \second{61.14} & \best{57.06} & \second{40.21}
& \second{20.99} & \best{17.56} & \second{18.46}
& \second{73.50} & \best{91.68} & 42.71
& \second{14.60} & 99.60 & \best{68.89}
& \second{64.01} & \best{49.04} \\

\textbf{\lorekvD}
& \best{26.97} & 41.36 & 44.92
& \best{61.96} & 55.14 & \best{40.51}
& \best{21.41} & 16.98 & \best{18.63}
& \best{74.00} & 91.41 & \best{43.47}
& 13.11 & \second{99.90} & 68.36
& \best{64.13} & 48.89 \\

\midrule
\multicolumn{18}{c}{\textit{Budget} $B=1024$} \\
\midrule
CriticalKV
& 27.78 & 43.72 & 45.60 & \best{61.47} & 56.30 & 39.21
& \second{21.57} & \best{17.31} & 19.34
& \second{73.50} & \best{91.38} & 42.45
& \best{14.50} & \best{100.00} & 69.85 & 63.57 & 49.22 \\

AnDPro
& 28.11 & \second{45.07} & \second{46.82} & 60.53 & 58.02 & 40.32
& 21.39 & 16.88 & 19.33
& \best{74.00} & \best{91.38} & \second{42.94}
& \best{14.50} & \best{100.00} & \second{69.88}
& \best{64.82} & 49.62 \\

\textbf{\lorekvE}
& \second{28.19} & \best{45.17} & 46.79
& \second{60.65} & \second{58.10} & \second{40.42}
& 21.47 & 16.90 & \second{19.35}
& \best{74.00} & \best{91.38} & 42.86
& \best{14.50} & \best{100.00} & \best{69.90}
& \second{64.77} & \second{49.65} \\

\textbf{\lorekvD}
& \best{28.58} & 44.95 & \best{48.04}
& 59.93 & \best{58.34} & \best{40.80}
& \best{21.76} & \second{17.05} & \best{19.63}
& \second{73.50} & \second{91.37} & \best{43.51}
& \second{14.45} & \second{99.75} & 69.43
& 64.45 & \best{49.72} \\
\bottomrule
\end{tabular}
\end{adjustbox}
\end{table}

\subsection{RULER: Protocol and Per-Task Results}
\label{app:ruler_results}

Tables~\ref{tab:ruler_mistral_full} and~\ref{tab:ruler_llama_full} report the full RULER results across 13 tasks, context lengths of 4K--16K, and $B\in\{128,256,512,1024\}$; Table~\ref{tab:ruler_llama_long} extends Llama-3.1-8B-Instruct to 64K and 128K at $B{=}128$, as discussed in Appendix~\ref{app:ruler_long}. Following the official RULER protocol \citep{hsieh2024ruler}, we evaluate 500 examples per task. Multi-reference synthetic tasks use \texttt{string\_match\_all}, which assigns fractional credit across reference answers, while QA uses \texttt{string\_match\_part}; scores are averaged over examples and rounded to two decimals, allowing finer granularity than the $0.2$-point step of binary accuracy. The overall trend matches the LongBench results: LORE-KV provides its largest gains under tight cache budgets, particularly on aggregation-heavy tasks such as variable tracking (VT), frequent-word extraction (FWE), and common-word extraction (CWE), while multi-query retrieval (MQ) and multi-key retrieval (MK3) remain more challenging. The matched RULER ablations fix $T=8$ and vary the remaining components. As one anchor point, at 4K and $B=128$, increasing the number of lookahead particles from $M=1$ to $M=4$ improves the average from $56.90$ to $58.40$. The same overall trend persists at larger budgets, whereas reliability weighting, adaptive head budgets, and diversity-aware selection produce comparatively small changes. Together, these results reinforce that the primary improvement comes from estimating the future-query reference rather than from the optional plug-ins. The complementary LongBench $(M,T)$ sweep is reported in Table~\ref{tab:ablation_geometry}.

\subsection{RULER at 64K and 128K Context Lengths}
\label{app:ruler_long}

\begin{figure}[t]
\centering
\includegraphics[width=\textwidth]{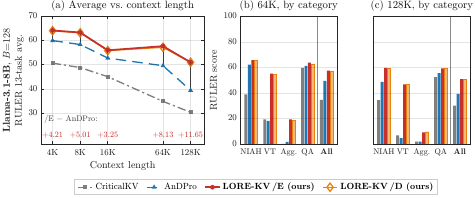}
\caption{\textbf{RULER at long context.} Llama-3.1-8B-Instruct, $B{=}128$. (a) 13-task average from 4K to 128K; red numbers give the \lorekvE{} margin over AnDPro. (b, c) Averages by task category at 64K and 128K, where Agg.\ averages CWE and FWE. Per-task results are given in Tables~\ref{tab:ruler_llama_full} and~\ref{tab:ruler_llama_long}.}
\label{fig:ruler_long_avg}
\end{figure}

\textbf{Setup.}
Table~\ref{tab:ruler_llama_long} extends the Llama-3.1-8B-Instruct results of Table~\ref{tab:ruler_llama_full} to 64K ($65{,}536$) and 128K ($131{,}072$) tokens at $B{=}128$, which keeps under $0.2\%$ of the context. We compare CriticalKV, AnDPro, and both LORE-KV variants on the same 13 tasks with 500 examples per task, so each method and length has $6{,}500$ scored examples, and we use the same string-match metrics as in Appendix~\ref{app:ruler_results}. LORE-KV keeps the core configuration of Table~\ref{tab:hyperparams} ($M{=}4$, $T{=}8$, $w{=}32$, uniform head budgets, top-$K$ selection), and all methods share the observation window, chunk size, and protected set used at 4K--16K. Only the context length changes.

\textbf{Average results.}
Figure~\ref{fig:ruler_long_avg} summarizes the averages. \lorekvE\ obtains the best 13-task average at both lengths, with $57.75$ at 64K and $51.05$ at 128K, against $49.62$ and $39.40$ for AnDPro and $34.82$ and $30.28$ for CriticalKV. The margin over AnDPro is $8.13$ points at 64K and $11.65$ points at 128K, and the margin over CriticalKV is $22.93$ and $20.77$ points. Together with the 4K--16K gaps of $4.21$, $5.01$, and $3.25$ points, this supports the trend in Section~\ref{subsec:ruler}: future-query estimation helps more when the retained cache is a smaller fraction of the context. Between 64K and 128K, the average of \lorekvE\ drops by $6.70$ points and that of AnDPro by $10.22$. CriticalKV drops by only $4.54$ points, but from a much lower 64K average. The two LORE-KV variants differ by at most $0.46$ points on average.

\textbf{Per-task results.}
The largest gains are on VT ($55.36$ versus $18.32$ for AnDPro at 64K; $47.00$ versus $4.80$ at 128K), on FWE ($36.80$ versus $3.13$ at 64K), on S3, where both baselines score zero, and on MV and MQ at 128K ($62.30$ versus $30.65$ and $46.25$ versus $26.50$). There are also regressions and saturated tasks. AnDPro is better on MQ at 64K ($74.35$ versus $50.90$), CriticalKV is better on S1 at 128K ($100.00$ versus $99.80$), and all methods score close to zero on MK3 at both lengths and on CWE at 128K.

\begin{table}[t]
\caption{RULER per-task results with Llama-3.1-8B-Instruct at 64K and 128K context lengths and $B{=}128$ (500 examples per task). Abbreviations follow Table~\ref{tab:ruler_mistral_full}. Columns in which all methods tie are not highlighted.}
\label{tab:ruler_llama_long}
\centering
\setlength{\tabcolsep}{3pt}
\resizebox{\textwidth}{!}{%
\begin{tabular}{llcccccccccccccc}
\toprule
$B$ & Method & CWE & FWE & MK1 & MK2 & MK3 & MQ & MV & S1 & S2 & S3 & QA-1 & QA-2 & VT & Avg. \\
\midrule

\multicolumn{16}{c}{\textit{Context length} $=64$K} \\
\midrule
128 & CriticalKV
& 0.16 & 0.47 & 72.40 & 34.80 & 0.00
& 15.90 & 12.30 & 100.00 & 77.40 & 0.00
& 73.00 & 46.60 & 19.60 & 34.82 \\

128 & AnDPro
& 0.82 & 3.13 & 97.20 & 67.60 & 0.00
& \best{74.35} & 66.30 & 100.00 & \second{94.60} & 0.00
& 76.00 & 46.80 & 18.32 & 49.62 \\

128 & \textbf{\lorekvE}
& \best{2.32} & \best{36.80} & \second{99.20} & \second{78.80} & 0.00
& 50.90 & \best{68.15} & 100.00 & \best{100.00} & \best{31.20}
& \best{78.60} & \best{49.40} & \best{55.36} & \best{57.75} \\

128 & \textbf{\lorekvD}
& \second{2.14} & \second{35.40} & \best{99.40} & \best{79.80} & 0.00
& \second{51.10} & \second{68.05} & 100.00 & \best{100.00} & \second{27.80}
& \second{77.00} & \second{49.00} & \second{55.04} & \second{57.29} \\

\midrule
\multicolumn{16}{c}{\textit{Context length} $=128$K} \\
\midrule
128 & CriticalKV
& 0.06 & 3.93 & 71.80 & 21.40 & 0.00
& 1.40 & 3.25 & \best{100.00} & 79.60 & 0.00
& 67.40 & 37.80 & \second{7.04} & 30.28 \\

128 & AnDPro
& 0.06 & 4.00 & 91.40 & 54.80 & 0.00
& 26.50 & 30.65 & 95.80 & \second{92.80} & 0.00
& 72.80 & 38.60 & 4.80 & 39.40 \\

128 & \textbf{\lorekvE}
& 0.06 & \second{18.47} & \second{98.40} & \best{60.20} & 0.00
& \best{46.25} & \second{62.30} & \second{99.80} & \best{100.00} & \best{12.40}
& \best{76.60} & \second{42.20} & \best{47.00} & \best{51.05} \\

128 & \textbf{\lorekvD}
& 0.06 & \best{19.60} & \best{98.60} & \second{59.80} & 0.00
& \second{45.10} & \best{63.15} & \second{99.80} & \best{100.00} & \second{11.60}
& \second{76.00} & \best{42.80} & \best{47.00} & \second{51.04} \\

\bottomrule
\end{tabular}%
}
\end{table}

\begin{table}
\centering
\captionsetup[subtable]{font=small, labelfont=bf, skip=2pt}
\caption{\textbf{Lookahead geometry and compute scaling.} LongBench on
Mistral-7B-Instruct-v0.2 at $B{=}128$: \textbf{(a)} \lorekvE{} across
trajectory count $M$ and horizon $T$; \textbf{(b)} quality frontier under a
matched lookahead-token budget versus AnDPro.}
\label{tab:ablation_geometry}

\makebox[\textwidth][c]{%
\begin{minipage}[c]{0.50\textwidth}
\vspace{-20pt}
\centering
\small
\setlength{\tabcolsep}{3.5pt}
\renewcommand{\arraystretch}{1.2}
\begin{tabular}{lccccc}
\toprule
$M$ & $T{=}2$ & $T{=}4$ & $T{=}8$ & $T{=}16$ & $T{=}32$ \\
\midrule
$0$ & \multicolumn{5}{c}{35.87 (independent of $T$)} \\
$1$ & 38.78 & 38.99 & 39.81 & 39.62 & 39.89 \\
$2$ & 38.83 & 39.49 & 40.17 & 40.21 & 40.14 \\
$4$ & 38.96 & 39.66 & 40.28 & \second{40.34} & 40.28 \\
$8$ & 39.01 & 39.72 & \best{40.69} & 40.39 & 40.30 \\
\bottomrule
\end{tabular}
\end{minipage}
\hfill
\begin{minipage}[c]{0.47\textwidth}
\centering
\includegraphics[width=\linewidth]
{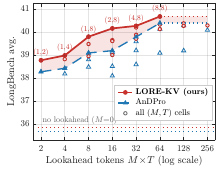}
\end{minipage}%
}

\vspace{2pt}

\makebox[\textwidth][c]{%
\begin{minipage}[c]{0.50\textwidth}
\centering
\subcaption{\lorekvE{} over $M$ and $T$}
\label{tab:ablation_geometry_grid}
\end{minipage}
\hfill
\begin{minipage}[c]{0.47\textwidth}
\centering
\subcaption{Pareto frontier at matched lookahead-token budget}
\label{tab:compute_pareto}
\end{minipage}%
}
\end{table}

\begin{table*}[t]
\caption{LongBench ablations (16-dataset average) with
Mistral-7B-Instruct-v0.2.}
\label{tab:ablation_summary}
\centering
\setlength{\tabcolsep}{3.5pt}
\renewcommand{\arraystretch}{1.12}
\resizebox{\columnwidth}{!}{%
\begin{tabular}{@{}llcccc@{}}
\toprule
\textbf{Axis} & \textbf{Variant}
& $B{=}128$ & $B{=}256$ & $B{=}512$ & $B{=}1024$ \\
\midrule

\multirow{2}{*}{Core}
& \textbf{\lorekvE\ core}
& 40.28 & 41.38 & 41.86 & 42.20 \\
& \textbf{\lorekvD\ core}
& 40.21 & 41.23 & 41.68 & 42.39 \\
\midrule

\multirow{4}{*}{Lookahead budget}
& $M{=}0$ (prompt window only; no lookahead term)
& 35.87 & 37.50 & 38.70 & 39.50 \\
& $M{=}1$ (single surrogate response)
& 39.81 & 40.85 & 41.56 & 42.03 \\
& $M{=}2$
& 40.17 & 40.18 & 41.10 & 41.22 \\
& $M{=}8$
& 40.69 & 40.97 & 41.52 & 41.65 \\
\midrule

\multirow{2}{*}{Reliability weighting}
& Log-probability only
& 40.29 & 41.39 & 41.30 & 41.40 \\
& Entropy $+$ disagreement
& 40.33 & 41.40 & 41.84 & 42.17 \\
\midrule

\multirow{4}{*}{Perturbation objective}
& Projected norm,
  $\|\mathbf{W}^{O}\Delta\mathbf{o}\|_2^2$
& 40.28 & 41.38 & 41.86 & 42.20 \\
& Accumulated attention mass
& 40.19 & 41.27 & 41.65 & 42.13 \\
& Unprojected norm, $\|\Delta\mathbf{o}\|_2^2$
& 40.22 & 41.22 & 41.65 & 42.13 \\
& AnDPro scoring with the same continuations
& \textit{40.24} & \textit{41.30} & \textit{41.74} & \textit{42.15} \\
\midrule

\multirow{3}{*}{Plug-ins}
& $+$ Adaptive head budgets
& 40.28 & 41.35 & 41.72 & 41.72 \\
& $+$ Adaptive budgets and diversity
& 40.30 & 40.33 & 41.50 & 41.80 \\
& $+$ Robust Aggregation
& 40.34 & 41.41 & 41.85 & 42.17 \\
\bottomrule
\end{tabular}%
}
\end{table*}

\begin{table}[t]
\caption{RULER ablations with Mistral-7B-Instruct-v0.2
(13-task averages).}
\label{tab:ruler_ablation}
\centering
\setlength{\tabcolsep}{4.5pt}
\resizebox{\textwidth}{!}{%
\begin{tabular}{@{}llllcccccc@{}}
\toprule
& & & &
\multicolumn{3}{c}{$B{=}128$}
& \multicolumn{3}{c}{$B{=}256$} \\
\cmidrule(lr){5-7}
\cmidrule(lr){8-10}

$M$ & Reliability weighting
& Head budgets & Selection
& 4K & 8K & 16K
& 4K & 8K & 16K \\
\midrule

\multicolumn{4}{@{}l}{AnDPro (strongest baseline)}
& 54.85 & 50.06 & 45.20
& 71.05 & 64.08 & 57.91 \\

\midrule

\multirow{3}{*}{1}
& \multirow{3}{*}{unit weight}
& uniform & top-$K$
& 56.90 & 54.29 & 52.15
& 68.45 & 62.77 & 57.70 \\

&
& adaptive & top-$K$
& 56.92 & 54.31 & \second{52.18}
& 68.50 & 62.81 & 57.75 \\

&
& adaptive & diversity
& 56.95 & 54.34 & \best{52.20}
& 68.55 & 62.85 & 57.78 \\

\midrule

\multirow{3}{*}{4}
& \multirow{3}{*}{Log-probability only}
& uniform & top-$K$
& \second{58.39} & 55.72 & 50.98
& 70.08 & 64.45 & 58.85 \\

&
& adaptive & top-$K$
& 58.28 & 55.68 & 50.95
& 70.04 & 64.42 & 58.82 \\

&
& adaptive & diversity
& 58.31 & 55.70 & 50.97
& 70.07 & 64.45 & 58.85 \\

\midrule

\multirow{2}{*}{4}
& \multirow{2}{*}{\textbf{\lorekvE}}
& adaptive & top-$K$
& 58.37 & \second{55.74} & 51.02
& \second{71.45} & \second{65.46} & \second{59.88} \\

&
& \textbf{uniform} & \textbf{top-$K$}
& \best{58.40} & \best{55.77} & 51.05
& \best{71.51} & \best{65.52} & \best{59.94} \\

\midrule

\multirow{3}{*}{4}
& \multirow{3}{*}{\textbf{\lorekvD}}
& adaptive & top-$K$
& 58.28 & 55.70 & 50.96
& 69.72 & 64.50 & 58.92 \\

&
& adaptive & diversity
& 58.31 & 55.72 & 50.98
& 69.75 & 64.53 & 58.95 \\

&
& \textbf{uniform} & \textbf{top-$K$}
& 58.31 & \second{55.74} & 50.97
& 69.77 & 64.56 & 58.94 \\

\bottomrule
\end{tabular}%
}
\end{table}

\subsection{NIAH Grid}
\label{app:niah_extra}
\label{app:large_models}
\begin{figure}[tb]
  \centering
  \includegraphics[width=\textwidth]
  {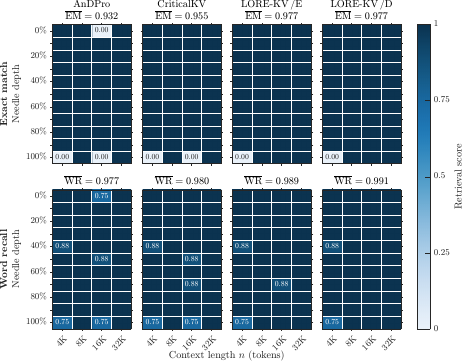}
\caption{\textbf{NIAH retrieval.} Exact-match and word-recall scores across needle depths and 4K--32K contexts on Llama-3.1-8B-Instruct at $B{=}1024$. Panel means are shown above each plot.}
  \label{fig:niah-heatmap}
\end{figure}

Figure~\ref{fig:niah-heatmap} reports exact match and word recall for
Llama-3.1-8B-Instruct at $B=1024$ over context lengths of 4K--32K and needle
depths of $0\%$--$100\%$ in $10\%$ steps, with one prompt per cell. All methods
use the same budget, observation window, and protected token set. Results are discussed in
Section~\ref{subsec:niah}.

\section{Mechanism Analysis}
\label{app:experiments_extra}

\subsection{Measurement Protocol}
\label{app:mechanism_protocol}

All mechanism analyses use the same states and scoring code as the deployed eviction pipeline, with
additional instrumentation used only for analysis. For each prompt, we capture the post-RoPE
prompt-window queries, the full prefill $\mathbf{K}^{l,h},\mathbf{V}^{l,h}$, and the $M$
lookahead trajectories together with $c^{(m)}_u$, $H^{(m)}_u$, $V^{(m)}_u$,
$D_m,\hat D_m$, and $\omega_m$. We also greedy-decode once with the \emph{full} cache to obtain
the future queries $\mathcal{Q}_{\mathrm{future}}$, which serve only as an analysis-time oracle.
These full-cache states are not retained during normal LORE-KV inference and are excluded from the
reported cache budget. The reliability channels of Eq.~\ref{eq:confidence} are stored separately,
including disagreement when $\lambda_D=0$, and per-step trajectory statistics require only $MT$
floats per prompt.

\begin{figure}[t]
  \centering
  \includegraphics[width=\textwidth]{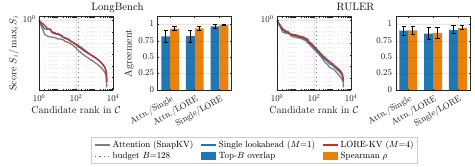}
  \caption{\textbf{Retention-order shift.} Candidate-score profiles, top-$B$ overlap, and Spearman correlation across eviction criteria. Mistral-7B-Instruct-v0.2, $B=128$, $M=4$.}
  \label{fig:score_comparison}
\end{figure}

For token-level analysis, we use the same output-perturbation criterion as LORE-KV:
\begin{equation}
s^{l,h}_{q,i}
=
\left|
\frac{a^{l,h}_{q,i}}
{1-a^{l,h}_{q,i}+\epsilon}
\right|
\left\|
\mathbf{W}^{O}_{l,h}\mathbf{v}^{l,h}_{i}
-
\mathbf{W}^{O}_{l,h}\mathbf{o}^{l,h}_{q}
\right\|_2 .
\label{eq:analysis_score}
\end{equation}
Scores are aggregated using Eq.~\ref{eq:final_score}, with all criteria sharing the same
captured states and selector. \textbf{Attention} uses SnapKV-style prompt-window attention;
\textbf{AnDPro window} uses prompt-side projected perturbation;
\textbf{LORE-KV} uses the projected perturbation score with \lorekvE\ weighting; and
\textbf{Single-lookahead} uses one unweighted trajectory as a score-level proxy for Lookahead
Q-Cache. These controlled comparisons measure how the query reference changes token ranking
(Figure~\ref{fig:score_comparison}), agreement with the future-query oracle
(Figure~\ref{fig:criterion_comparison}), retained evidence
(Figure~\ref{fig:retained_token_composition}), and behavioral fidelity across cache budgets
(Figure~\ref{fig:budget_equivalence}). Protected positions $\mathcal P$ are always retained,
and all remaining tokens use the same uniform per-head top-$\min(B,n)$ selector, so differences
are attributable to the scoring/query criterion rather than downstream selection.

Equation~\ref{eq:analysis_score} uses magnitude rather than the squared magnitude of
Eq.~\ref{eq:projected_score}; this preserves single-query rankings but changes the energy scale in
Figure~\ref{fig:layer_head_analysis}. For these analyses, chunking, adaptive head budgets, and
diversity-aware selection are disabled, and $B$ includes the ${\sim}33$ protected positions.
Figure~\ref{fig:layer_head_analysis} additionally sums energy over all positions and reports
GQA-expanded attention heads, while Figure~\ref{fig:retained_token_composition} identifies answer
tokens by exact token-id match to the gold answer; this may underestimate absolute evidence
retention but affects all criteria equally.

We use three complementary diagnostics. Query coverage measures how closely an estimated query set
matches the true future queries:
\begin{equation}
\operatorname{Cov}\big(\mathcal{Q}_{\mathrm{future}};
\{\hat{\mathcal{Q}}_s\},\boldsymbol{\pi}\big)
=
\frac{1}{H}\sum_h
\frac{1}{T_{\mathrm f}}\sum_t
\sum_s \pi_s
\max_{\hat{\mathbf q}\in\hat{\mathcal Q}_s}
\cos\!\big(
\mathbf q^{\mathrm{future}}_{t,h},
\hat{\mathbf q}
\big).
\label{eq:coverage}
\end{equation}
The oracle score $S^\star$ applies Eq.~\ref{eq:final_score} directly to
$\mathcal{Q}_{\mathrm{future}}$, and retention agreement is measured by
\begin{equation}
\operatorname{Recall}@B(S,S^\star)
=
\frac{
\left|\mathrm{top}_B(S)\cap\mathrm{top}_B(S^\star)\right|
}{B}.
\label{eq:recall_at_b}
\end{equation}
Finally, behavioral fidelity measures how closely the compressed cache reproduces the full-cache
attention output:
\begin{equation}
F(B)
=
\mathbb{E}_{\mathbf q\sim\mathcal Q_{\mathrm{future}}}
\cos\!\big(
\mathbf o^{\mathcal R(B)}_q,
\mathbf o_q
\big).
\label{eq:fidelity}
\end{equation}
Queries are unit-normalized for the cosine-based metrics, and the prompt-to-future mismatch
$\operatorname{Mis}(X)$ is defined in Section~\ref{subsec:mechanism}. Unless stated otherwise,
all analyses use Mistral-7B-Instruct-v0.2 on LongBench and 4K RULER with the core
\lorekvE\ configuration.

\subsection{Future-Query Coverage}
\label{app:query_coverage}

Figure~\ref{fig:montecarlo_queries} examines whether sampled lookahead queries better represent the
queries encountered during generation. We compare the prompt-window queries
$\mathcal{Q}^{l,h}_{\mathcal{W}}$, the sampled lookahead queries
$\mathcal{Q}^{(m),l,h}_{\mathrm{look}}$, and the true future queries
$\mathcal{Q}_{\mathrm{future}}$, where $\mathcal{Q}_{\mathrm{future}}$ is obtained only
retrospectively from full-cache decoding and is never available to the eviction rule.

Panel~(a) visualizes these three query sets for one representative layer and KV head
(layer $16$, head $18$) after projection onto the first two principal components.
The sampled trajectories move toward the region occupied by the true future queries,
illustrating the prompt-to-future query mismatch that motivates LORE-KV.

Panel~(b) quantifies this effect using the coverage metric in Eq.~\ref{eq:coverage}.
The prompt-window reference achieves $0.785$ coverage, while adding sampled trajectories
progressively increases coverage of $\mathcal{Q}_{\mathrm{future}}$, with most of the gain
obtained by approximately $M{=}4$. The sweep uses nested prefixes of the same eight sampled
trajectories for each prompt, so changes reflect increased future-query coverage rather than
different random draws.

Coverage measures geometric support in query space rather than downstream eviction quality.
Consequently, the small change from $M{=}0$ to $M{=}1$ ($0.785\!\to\!0.786$) does not
contradict the substantial end-to-end improvement from a single response-side rollout in
Tables~\ref{tab:ablation_summary} and~\ref{tab:ruler_ablation}: one trajectory can already
provide a more relevant query reference, while additional trajectories improve coverage of
alternative plausible futures. Together, these results support the central design of LORE-KV:
estimating token utility over multiple plausible response-side query trajectories provides a
better reference for eviction than relying on prompt-side queries alone.


\subsection{Retention-Order Shift}
\label{app:score_shift}

Figure~\ref{fig:score_comparison} tests whether lookahead changes token selection rather than simply
rescaling prompt-side scores. At $B{=}128$, attention and \lorekvE\ retain only $0.82$ of the same
tokens on LongBench and $0.86$ on RULER, despite high overall Spearman rank correlation
($\rho{=}0.94$ and $0.87$). Thus, lookahead preserves much of the global ranking while changing
the tokens near the retention boundary. In contrast, single-lookahead and \lorekvE\ are much more
similar ($0.97$/$0.92$ top-$B$ overlap on LongBench/RULER), showing that most of the ranking
shift already comes from replacing prompt-side queries with response-side queries.

\begin{figure}[t]
  \centering
  \includegraphics[width=\textwidth]{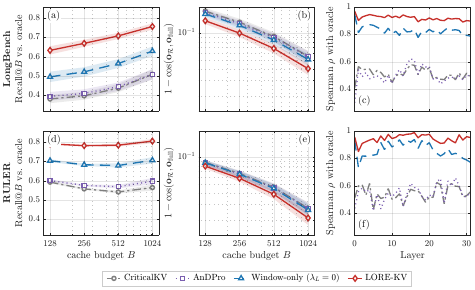}
  \caption{\textbf{Criterion comparison.}
  \textbf{(a,d)} Recall@$B$ against the future-query oracle.
  \textbf{(b,e)} Attention-output error (lower is better).
  \textbf{(c,f)} Layer-wise Spearman correlation with the oracle.}
  \label{fig:criterion_comparison}
\end{figure}

Figure~\ref{fig:criterion_comparison} compares each criterion with the future-query oracle under the same selection rule. At $B{=}128$ on LongBench, \lorekvE\ increases oracle-retention agreement from $0.50$ for the window-only control to $0.63$, while also reducing full-cache output error from $0.177$ to $0.138$. The same advantage persists across budgets and on RULER, showing that response-side lookahead yields a token ranking that better matches future decoding needs.

These controlled comparisons isolate the source of the gain: changing the query reference toward
plausible future decoding states produces substantially better agreement with the future-query oracle,
whereas prompt-side attention- and value-based criteria remain much closer to one another. This
supports the central hypothesis of LORE-KV that \emph{which queries define token importance} is
more important than further refining the prompt-side scoring rule.

\subsection{Reliability Weighting: Robustness Beyond the Query Reference}
\label{app:divergence}

Reliability weighting acts only on the already-sampled trajectories and therefore adds no extra test-time decoding. Trajectory disagreement $\hat D_m$ provides an additional reliability signal across sampled futures. Its role is deliberately conservative: the reliability signals are only weakly correlated with oracle retention utility $\operatorname{Recall}@B$ (Eq.~\ref{eq:recall_at_b}), particularly on RULER, and hard-filtering low-reliability trajectories does not reduce the residual query-coverage error $1-\operatorname{Cov}$. We use reliability as a soft correction through $\omega_m$ in Eq.~\ref{eq:confidence}, reducing the influence of uncertain or atypical trajectories without discarding plausible futures.

Importantly, the main robustness gain comes from \emph{covering multiple plausible futures}. Figure~\ref{fig:montecarlo_queries}(b) shows that future-query coverage increases from $0.785$ for the prompt-window reference to $0.802$ at $M{=}2$ and $0.811$ at $M{=}8$, with gains saturating near $M{=}4$. This shows that sampling multiple trajectories provides broader coverage of the future-query space than relying on a single prompt-side reference. Reliability weighting then acts as a complementary soft correction, reducing the influence of uncertain or atypical trajectories without discarding plausible futures.

The oracle diagnostics in Figure~\ref{fig:criterion_comparison} reinforce this interpretation. At $B{=}128$ on LongBench, $\operatorname{Recall}@B$ against the future-query oracle increases from $0.39$ for the prompt-side criteria (CriticalKV and AnDPro) and $0.50$ for the $\lambda_L{=}0$ window-only control to $0.63$ for \lorekvE. The same ordering persists across budgets and on RULER. Similarly, the attention-output error $1-\cos(\mathbf{o}_{\mathcal R},\mathbf{o}_{\mathrm{full}})$ decreases from $0.177$ to $0.138$ at $B{=}128$, while \lorekvE\ maintains substantially higher layer-wise agreement with the oracle (Spearman $\rho\approx0.90$-$0.95$) than either the window-only control ($\approx0.80$-$0.85$) or prompt-side criteria ($\approx0.45$-$0.60$). Finally, \lorekvE\ and \lorekvD\ remain within $0.19$ LongBench points at every budget (Table~\ref{tab:ablation_summary}). Together, these results show that LORE-KV is robust primarily because it estimates utility over multiple plausible future queries; entropy and disagreement weighting provide a secondary safeguard against noisy or atypical samples rather than a separate source of performance.

\begin{figure}[t]
  \centering
  \includegraphics[width=\textwidth]{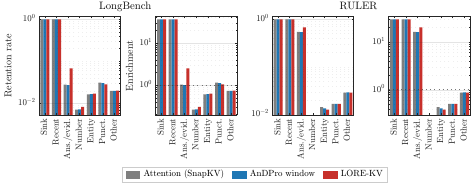}
  \caption{\textbf{Retention by token class.} Mean retained fraction (left) and class enrichment $P(c\mid\mathcal R)/P(c\mid X)$ (right).}
  \label{fig:retained_token_composition}
\end{figure}
\begin{figure}[t]
  \centering
  \includegraphics[width=\textwidth]{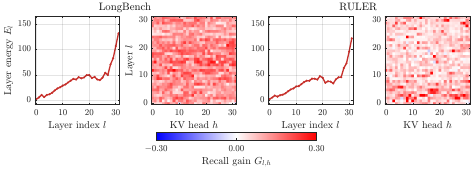}
  \caption{\textbf{Future-query signal across layers and heads.} Layer perturbation energy (left) and retention-recall gain from multi-lookahead queries (right). Mistral-7B-Instruct-v0.2, $B=128$, $M=4$.}
  \label{fig:layer_head_analysis}
\end{figure}
\begin{figure}[t]
  \centering
  \includegraphics[width=\textwidth]{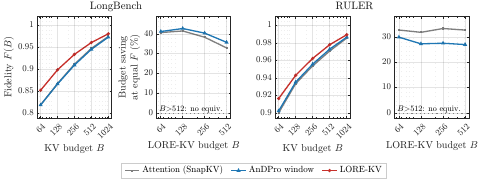}
  \caption{\textbf{Budget equivalence under behavioral fidelity.} Fidelity across cache budgets (left) and baseline cache savings required to match \lorekvE\ (right).}
  \label{fig:budget_equivalence}
\end{figure}
\subsection{Retained-Token Composition}
\label{app:composition}

Figure~\ref{fig:retained_token_composition} examines which token classes are preserved after compression. The protected sink and recent-token classes are identical across methods, while the main difference is in answer-bearing evidence. On LongBench, \lorekvE\ increases answer/evidence enrichment from approximately $1\times$ for the prompt-side baselines to $2.51\times$, with the corresponding retention rate rising from about $0.028$ to $0.069$. The same trend holds on RULER, where answer/evidence enrichment increases to $20.3$ compared with $16.5$-$16.3$ for the baselines. Other token classes change only modestly. This consistent shift across benchmarks shows that future-query lookahead does not indiscriminately retain more tokens; it preferentially preserves evidence that is relevant to subsequent decoding, providing a direct mechanism for the downstream gains reported in Section~\ref{subsec:niah}.

\subsection{Layer- and Head-Level Structure}
\label{app:layer_head}

Figure~\ref{fig:layer_head_analysis} shows that the future-query signal is distributed across the network rather than concentrated in a single component. Perturbation energy generally increases with depth, while replacing window-only queries with multi-lookahead queries improves retention recall for most heads, with the largest gains concentrated in a subset of middle-depth heads. The same qualitative pattern appears on both LongBench and RULER, supporting the robustness of the lookahead signal across tasks. This non-uniformity is consistent with head specialization \citep{feng2024adakv,fu2024headkv} and motivates the adaptive allocation in Eq.~\ref{eq:adaptive_budget}; however, its modest additional gain indicates that the core \lorekvE\ score already captures most of the benefit.

\subsection{Budget Equivalence in Behavioral Fidelity}
\label{app:budget_curve}
\begin{wraptable}{r}{0.49\textwidth}
\vspace{-11pt}
\centering
\caption{Baseline budgets matching \lorekvE\ on LongBench
(Mistral-7B-Instruct-v0.2).}
\label{tab:budget_equivalence}
\scriptsize
\setlength{\tabcolsep}{3.5pt}
\resizebox{0.49\textwidth}{!}{%
\begin{tabular}{lcccc}
\toprule
\textbf{\lorekvE\ budget} & 128 & 256 & 512 & 1024 \\
\midrule
Ada-KV      & $\sim$332 & $\sim$745 & $>$1024 & $>$1024 \\
CriticalKV  & \second{$\sim$307} & \second{$\sim$709} & $>$1024 & $>$1024 \\
AnDPro      & \best{$\sim$208} & \best{$\sim$414} &
              \best{$\sim$743} & $>$1024 \\
\midrule
Saving (\%) & 38.5 & 38.1 & 31.1 & $>0$ \\
\bottomrule
\end{tabular}
}
\vspace{-8pt}
\end{wraptable}
Figure~\ref{fig:budget_equivalence} complements Table~\ref{tab:budget_equivalence} by measuring cache quality continuously with the behavioral fidelity $F(B)$ of Eq.~\ref{eq:fidelity}. Across the full budget range, \lorekvE\ consistently preserves full-cache behavior better than the prompt-side baselines on both LongBench and RULER, with the largest advantage at tighter budgets and the same ordering maintained as $B$ increases.

Matching the fidelity of \lorekvE\ requires approximately $33$-$43\%$ more retained budget on LongBench and $27$-$34\%$ on RULER. This shows that the advantage is not tied to a single operating point, but persists across compression levels. These fidelity results are consistent with the task-level performance trends reported for LongBench in Table~\ref{tab:longbench_mistral} and for RULER in Figure~\ref{fig:ruler_main} and Tables~\ref{tab:ruler_mistral_full} and~\ref{tab:ruler_llama_full}. If a baseline does not reach the target fidelity within the tested budgets, we report only a lower bound on the required budget saving.

\section{The Two Reliability Variants in the
$(\beta,\gamma,\lambda_D)$ Plane}
\label{app:reparam_grid}

The \lorekvE\ and \lorekvD\ variants differ only in the reliability coefficients defined in
Table~\ref{tab:hyperparams}, with end-to-end comparisons reported in
Tables~\ref{tab:longbench_mistral}, \ref{tab:longbench_llama},
\ref{tab:ablation_summary}, and \ref{tab:ruler_ablation}, and in
Figure~\ref{fig:ruler_main}. At $B{=}128$, both variants re-weight the same nested prefixes from a
single eight-particle draw per prompt, keeping the sampled trajectories and test-time compute fixed.

\end{document}